\documentclass[10pt,twocolumn,letterpaper]{article}

\usepackage{wacv}              

\usepackage{booktabs}
\usepackage[table]{xcolor}
\usepackage{graphicx}
\usepackage{subcaption}
\usepackage{caption}
\definecolor{Gray}{gray}{0.9}
\definecolor{gaincolor}{RGB}{232,244,255}
\usepackage{float}
\usepackage{xcolor}
\usepackage{array}
\usepackage{multirow}
\usepackage{xcolor}
\usepackage{algpseudocode}
\usepackage{booktabs}
\usepackage{multirow}
\usepackage{caption}
\usepackage{tocloft}
\usepackage{dblfloatfix}
\usepackage{float}

\usepackage{color}
\usepackage{colortbl}
\usepackage{xcolor}
\usepackage{tikz}
\usepackage{comment}
\usepackage{algorithm}
\usepackage{bibunits}
\usepackage[percent]{overpic}
\usepackage{tikz}
\usetikzlibrary{calc}

\definecolor{gold}{HTML}{BD820B}
\definecolor{silver}{HTML}{909090}
\definecolor{bronze}{HTML}{9A5F26}
\definecolor{lgray}{gray}{0.95}

\definecolor{gaincolor}{RGB}{230, 245, 255}

\definecolor{Gray}{gray}{0.91}
\definecolor{LightCyan}{rgb}{0.82,0.82,1}
\newcolumntype{a}{>{\columncolor{Gray}}c}
\newcolumntype{B}{>{\columncolor{LightCyan}}c}

\definecolor{dullgreen}{RGB}{40,120,60}
\definecolor{dullyellow}{RGB}{180,150,40}
\definecolor{dullred}{RGB}{170,60,60}
\definecolor{dullblue}{RGB}{60,90,160}
\definecolor{dullcyan}{RGB}{80,140,170}
\definecolor{wacvblue}{rgb}{0.21,0.49,0.74}
\usepackage[pagebackref,breaklinks,colorlinks,allcolors=wacvblue]{hyperref}

\usepackage{enumitem}
\usepackage[toc,page]{appendix}
\usepackage{titletoc}

\def\wacvPaperID{596} 
\def\confName{WACV}
\def\confYear{2027}

\title{Distractors are the Real Enemy: Visual Distraction Adversarial Perturbations on SAM2-Based Video Tracking and Segmentation}

\author{
\mbox{
Adonay Demewez Gebremedhin\textsuperscript{1} \quad
Mohamad Alansari\textsuperscript{2} \quad
Wessam Shehieb\textsuperscript{1} \quad
Muzammal Naseer\textsuperscript{2,3}
}\\[0.6em]
\mbox{
\textsuperscript{1}Ajman University, Ajman, UAE
\qquad
\textsuperscript{2}Khalifa University, Abu Dhabi, UAE
}\\
\textsuperscript{3}University of Western Australia, Australia\\[0.5em]
{\scriptsize\ttfamily
202311465@ajmanuni.ac.ae \quad
100061914@ku.ac.ae \quad
w.shehieb@ajman.ac.ae \quad
muhammadmuzammal.naseer@ku.ac.ae
}
}

\begin{document}
\maketitle

\begin{abstract}
Promptable video trackers such as SAM2 achieve accurate and robust performance through strong temporal consistency, making them important foundations for downstream video understanding and tracking applications. As their use continues to expand, understanding their robustness to adversarial perturbations becomes increasingly important. Existing attacks on SAM2 largely focus on local mask degradation and short-term temporal feature disruption, leaving its longer-term dynamics and competing object interferences (Distractors) insufficiently explored. We observe that SAM2 itself exposes useful distractor cues when tracking becomes
uncertain. Based on these observations, we propose \textbf{DistractAttack}, a
white-box attack that uses background distractors to disrupt target
tracking persistently. Its core procedure,
\textbf{Counterfactual Distractor Search} (CDS), propagates clean and attacked
trajectories in parallel to discover plausible competing regions and identify
those that become increasingly effective under perturbation. The discovered
evidence is accumulated over time to weaken target support and promote
recurrent alternatives or disappearance predictions. Extensive experiments under video-level and one-shot perturbations across Visual Object Tracking and Video Object Segmentation demonstrate strong attack effectiveness,  raising DRE to $4.4\times$ and NRE to $7.3\times$ their nearest baselines under video perturbation, while transferring effectively across model scales and tracker variants in black-box settings. Project page: \url{https://adonaydem.github.io/sam2-da-page/}.
\end{abstract}

\section{Introduction}
\label{sec:introduction}

Recent advances in promptable video segmentation have demonstrated strong and generalizable performance across downstream tasks such as Visual Object Tracking (VOT) and Video Object Segmentation (VOS). SAM2 \cite{sam2}, in particular, extends promptable segmentation to video through a recurrent memory mechanism that propagates object information across frames. Building on this foundational model, recent methods have further improved long-term tracking through stronger motion modeling, memory management, and distractor handling \cite{samurai,sam2long,dam4sam,samite,him2sam}. While promptable trackers have become increasingly robust, it remains unclear whether their specialized designs resist adversarial perturbations or share transferable vulnerabilities.

\begin{figure}[t]
    \centering
    \includegraphics[
        width=0.92\columnwidth
    ]{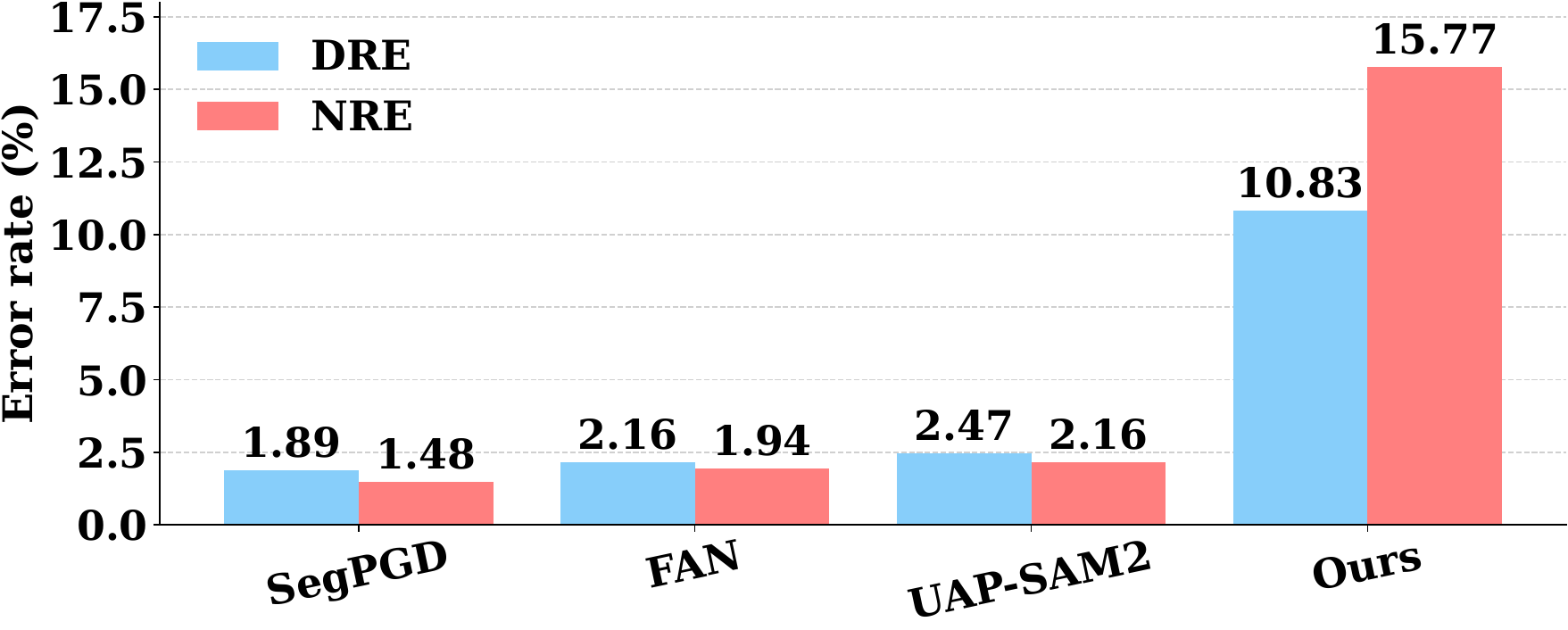}
    \caption{Drift-Rate Error (DRE) vs Not-Reported Error (NRE) performance across video perturbation methods. Higher DRE and NRE represent a balanced and comprehensive attack success.}
    \label{fig:hook}
\end{figure}

Attacks against single-object tracking and VOS manipulate classification,
bounding-box regression, temporal consistency, trajectories, and motion
cues~\cite{csa,fan,iouattack,spark,aba,chen2020one}, or exploit initialization
errors and hard regions to induce persistent segmentation
failure~\cite{jiang2023exploring,li2023adversarial}. Although these studies show
that temporal recurrence and stored representations can amplify brief
perturbations, their objectives were developed largely for earlier
architectures such as Siamese networks~\cite{fan} and do not readily transfer
to foundation models such as SAM2. In parallel, attacks on the Segment Anything
Model (SAM)~\cite{sam} consider mask manipulation, universal and prompt-agnostic
perturbations, and feature-specific objectives
~\cite{attacksam,samuap,pata,huang2023robustness,jiang2024cross,liu2024cross,9614158,uad,sra,darksam,xiao2026diffusion,umigrat,qin2025sam},
but remain image-based and overlook the memory-driven dynamics of promptable
video trackers. For SAM2, Robust-SAM showed that adversarial perturbations can
disrupt temporal consistency~\cite{long2025robust}, while UAP-SAM2 jointly
perturbs semantic features and consecutive-frame
consistency~\cite{uapsam2}. BadVSFM further considers training-time
backdoors~\cite{zhang2025backdoor}. Yet existing SAM2 attacks still largely
target segmentation degradation or mask disappearance, while their temporal
objectives remain centered on relatively local feature interactions rather
than the long-term object trajectory. This leaves open how an attack can
exploit the longer-term mechanisms that modern promptable trackers use to
preserve and recover target identity under adverse challenges.

We introduce \textbf{DistractAttack}, a distractor-centric white-box attack for recurrent promptable segmentation that uses model-supported distractors to destabilize target tracking rather than merely erase the current mask. Its core, \textbf{Counterfactual Distractor Search} (CDS), propagates clean and attacked trajectories in parallel: the clean trajectory identifies plausible non-target competitors, while the attacked trajectory reveals which become increasingly influential under a bounded $\ell_\infty$ perturbation. Accumulating this evidence over time, CDS turns local ambiguity into persistent tracking failures, including wrong-object capture, spatial displacement, unstable localization, and disappearance. We evaluate video-level and one-shot perturbations across VOT and VOS, with transfer studies across model scales and state-of-the-art tracker variants. SAM2's memory mechanism is particularly vulnerable to one-shot initialization perturbations, whose effects persist throughout the video. Under video-perturbation and matched protocol, DistractAttack achieves the highest Drift-Rate Error (DRE) and Not-Reported Error (NRE)~\cite{kristan2023first} by a substantial margin (see Fig.~\ref{fig:hook}).

Our core contributions are summarized as follows:
\begin{itemize}
   \item We propose the first distractor-centric adversarial attack against SAM2, exposing its vulnerability to distractors in the surrounding scene. \textbf{Counterfactual Distractor Search} couples clean and attacked trajectories to exploit evolving distractor evidence and induce persistent tracking failure.

    \item We extensively study DistractAttack under two complementary perturbation settings: \textbf{Video Perturbation (VP)} and \textbf{One-shot Perturbation (OP)} across both VOT and VOS tasks.

    \item We provide a broad evaluation against existing attacks and assess real-world practicality of our method via black-box transferability across model sizes and SAM2 tracker variants.
\end{itemize}

\section{Related Work}
\label{sec:related}

\subsection{Promptable Video Segmentation and Tracking}

SAM introduced promptable class-agnostic image segmentation \cite{sam}, and
SAM2 extended this paradigm to video through a streaming memory that propagates
an initialized object across frames \cite{sam2}. SAM2 further broadened the
family toward concept-prompted detection, segmentation, and tracking
\cite{sam3}. Recent SAM2 extensions strengthen long-term tracking through
motion-aware memory selection in SAMURAI \cite{samurai}, multiple temporal
hypotheses in SAM2Long \cite{sam2long}, distractor-aware memory in DAM4SAM
\cite{dam4sam}, position prompting in SAMITE \cite{samite}, and combined
long- and short-term modeling in HiM2SAM \cite{him2sam}. Their shared focus on
distractors and memory reliability identifies \emph{identity confusion} as a central
challenge in promptable tracking. We study the complementary adversarial question of whether a bounded perturbation can exploit such distractors to progressively confuse the tracker’s target representation and disrupt its temporal consistency.

\subsection{Adversarial Video Tracking and Segmentation}

Adversarial video tracking has shown that recurrence, association, and stored
representations can turn brief perturbations into persistent errors. In visual object tracking,
FAN, IoU Attack, SPARK, CSA, and ABA attack temporal box consistency,
sparse online updates, classification responses, and motion blur
\cite{fan,iouattack,spark,csa,aba}, while adversarial VOS exploits
initialization errors that aim to induce mask degradation
\cite{jiang2023exploring,li2023adversarial}. However, these frameworks remain closely
tied to their victim architectures and have limited compatibility with
recent SAM-based trackers. Attacks on SAM have covered mask removal, targeted
generation, prompt-agnostic feature manipulation,
region suppression and spatial--frequency attacks
\cite{attacksam,samuap,pata,huang2023robustness,jiang2024cross,liu2024cross,
uad,sra,darksam,xiao2026diffusion,umigrat,qin2025sam}. Direct studies of SAM2
include Robust~SAM (CPA), UAP-SAM2, BadVSFM, and DarkLLM
\cite{long2025robust,uapsam2,zhang2025backdoor,sun2026darkllm}. Among
these methods, UAP-SAM2 alone introduces an explicit temporal
objective, yet it merely relates consecutive encoder features rather than the history
accumulated in recurrent memory. We aim to exploit accumulated target history and recurring distractor evidence to induce persistent confusion between the target and plausible competing regions.

\section{Preliminary}
\label{sec:threat}

Let $x_{1:T}$ denote a video, $M_0$ the target annotation provided only at
the first frame, and $h_t$ the recurrent tracker state at $t$. We
follow standard single-object VOT and VOS protocols, where the target is initialized by a
box and mask, respectively. SAM2 then performs:
\begin{equation}
(M^s_t,h_t,\mathcal{A}_t)
=
\Phi(x_t,h_{t-1},M_0),
\label{eq:tracker}
\end{equation}
where $M^s_t$ is the selected mask and $\mathcal{A}_t$ collects auxiliary
decoder outputs, including confidence and predicted-IoU scores. Here, $\Phi$ may be either a clean or attacked tracker, $\Phi^c$ or $\Phi^a$, with encoder features $F^c_t$ or $F^a_t$, respectively. The SAM2 decoder
also produces multiple candidate masks before selecting $M^s_t$. We refer to
these candidates as \textbf{Mask Hypotheses} throughout the paper.

We assume a white-box attack with access to the model code, parameters, auxillary ouputs
and gradients, and can collect publicly available datasets. The attack acts exclusively
through bounded perturbations to the RGB input,
\begin{equation}
\widetilde{x}_t
=
\operatorname{clip}(x_t+\delta_t),
\qquad
\|\delta_t\|_{\infty}\leq\varepsilon ,
\label{eq:budget}
\end{equation}
where $\varepsilon$ denotes the perturbation budget.

\section{Methodology}
\label{sec:method}

\subsection{Observations and Overview}
\label{sec:observations}

\begin{figure}[!t]
    \centering
    \includegraphics[width=\columnwidth]{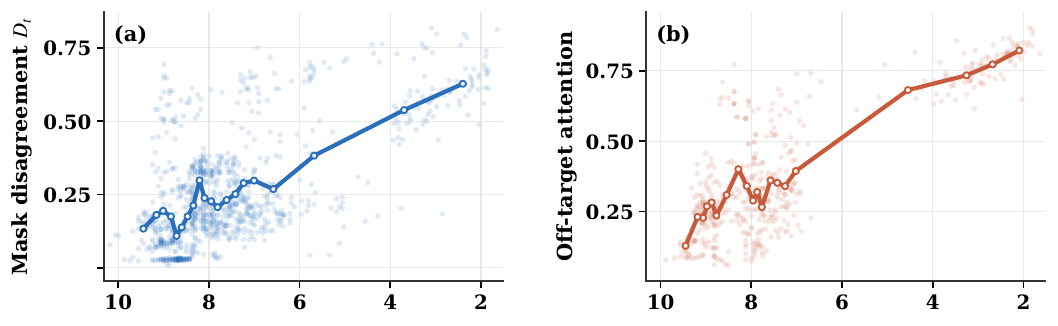}
    \caption{\textbf{Decoder ambiguity provides useful attack cues.}
    Lower objectness confidence (x-axis) is associated with (a) greater mask disagreement
    $D_t$ and (b) stronger off-target attention mass. The lines show averages across bins.}
    \label{fig:observations}
\end{figure}

\noindent\textbf{Observation I: Mask hypotheses diverge as confidence
falls.}
At each frame, the SAM2 decoder produces several mask hypotheses and selects
one using their predicted-IoU scores. We measure their disagreement as
\begin{equation}
D_t = 1-
\frac{1}{N_c}
\sum_{(i,j)\in\mathcal{C}_t}
\operatorname{IoU}\!\left(M_t^{(i)},M_t^{(j)}\right),
\qquad
N_c = |\mathcal{C}_t| ,
\label{eq:mask_disagreement}
\end{equation}
where $\mathcal{C}_t$ contains all pairs of predicted masks. A larger $D_t$
therefore indicates greater disagreement. For this analysis, we run clean SAM2.1-L on ten randomly selected videos, collect frame-wise decoder outputs. As shown in
Fig.~\ref{fig:observations}(a), disagreement increases as confidence in the
selected output falls. Further analysis shows that SAM~2 often shifts to distractors prior to tracking failure. The discarded
hypotheses therefore provide model-generated proposals for regions that may
compete with the target.

\noindent\textbf{Observation II: Decoder attention moves away from
the target as confidence falls.}
The selected mask may remain stable even as attention shifts toward competing
regions. We measure the fraction of mask-token attention outside the target object. Fig.~\ref{fig:observations}(b) shows that this fraction
rises as confidence falls, indicating growing support outside the target. The
attention map therefore complements the alternative masks in Obs. I by identifying
non-target regions that already influence the decoder.

Together, these observations motivate \emph{Counterfactual Distractor Search}
(CDS), which separates evidence collection from perturbation optimization. A
clean trajectory provides target features and competing regions, while an
attacked trajectory reveals how distractor evidence evolves under the bounded
perturbation. Optimization then suppresses target support, biases foreground responses toward
discovered distractors, and corrupts target distinctiveness across multiple
granularities. We consider two settings.
\textbf{Video Perturbation (VP)} optimizes one perturbation per video, while
\textbf{One-shot Perturbation (OP)} perturbs only the prompted frame
\cite{jiang2023exploring,chen2020one}. DistractAttack is primarily classified as untargeted because no distractor identity is specified in
advance. Its distractor proposals guide the optimization but may instead exhibit wrong-object capture, spatial
displacement, unstable switching, or disappearance.
\begin{figure*}[t]
    \centering

    \begin{tikzpicture}
        \node[inner sep=0] (fig) {
            \includegraphics[
                height=0.35\textheight,
                keepaspectratio
            ]{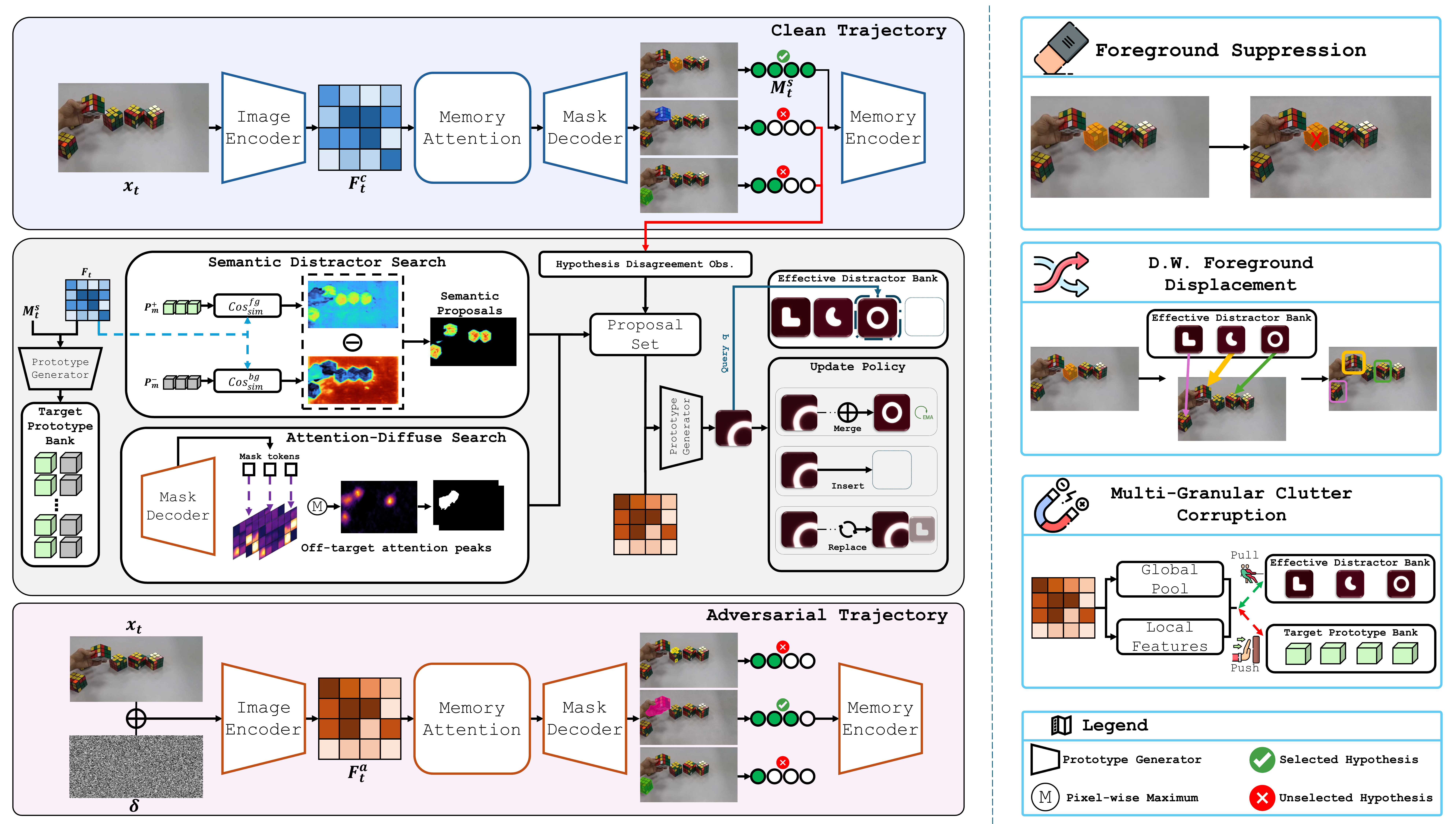}
        };

        \node[anchor=north] at
            ($(fig.south west)!0.333!(fig.south east)+(0,-0.3mm)$)
            {\scriptsize\textbf{(a)}};

        \node[anchor=north] at
            ($(fig.south west)!0.833!(fig.south east)+(0,-0.3mm)$)
            {\scriptsize\textbf{(b)}};
    \end{tikzpicture}

    \vspace{-3mm}

    \caption{\textbf{Overview of DistractAttack.}
(a) Counterfactual Distractor Search: Clean and attacked trajectories provide counterfactual target references
for proposing, validating, and accumulating distractors.
(b) CDS evidence guides foreground suppression, multi-granular clutter corruption, and
distractor-weighted foreground displacement to induce persistent tracking failure.}

    \label{fig:framework}
\end{figure*}

\subsection{Counterfactual Distractor Search (CDS)}
\label{sec:evidence}

\subsubsection{Target Prototype Bank}

For an arbitrary feature map $F$ and region mask $R$, we define a
\emph{Prototype Generator}:
\begin{equation}
\psi(F,R)=
\frac{1}{|R|}
\sum_{u\in R}F(u).
\label{eq:regional-prototype}
\end{equation}
where $u$ indexes spatial locations within region $R$.

We propagate a clean trajectory and store a reliable clean history of the target in a lightweight
prototype bank $\mathcal{T}$. Each reliable clean encoder features $F_t^c$ with selected mask $M_t^s$ contribute a
foreground--background pair using the prototype generator,
\begin{equation}
p_t^{+}=\psi(F_t^c,M_t^s),
\qquad
p_t^{-}=\psi(F_t^c,1-M_t^s). 
\end{equation}
The first frame pair is always retained. Later pairs are admitted only when
the model's predicted-IoU score and objectness logit exceed 0.5 and 0, respectively. To avoid repeatedly storing near-duplicate target appearances, the resulting foreground prototype is further required to pass a cosine novelty gate:
\begin{equation}
\begin{gathered}
\max_{m\in\mathcal{T}}
\cos\!\left(p_t^{+},p_m^{+}\right)
< \tau_{\mathrm{nov}}
\\[-1pt]
\Longrightarrow\quad
\mathcal{T}\leftarrow
\mathcal{T}\cup\{(p_t^{+},p_t^{-})\}.
\end{gathered}
\label{eq:novelty-gate}
\end{equation}
We set $\tau_{\mathrm{nov}}$ to 0.98. These mechanisms help us reject unreliable and
redundant updates while preserving appearance changes needed to distinguish the
target from its surroundings.

\subsubsection{Search Components}
During video propagation, CDS searches for plausible distractors from a set of signals exposed by the clean and attacked trajectories. We consider three sources of evidence.

\noindent\textbf{A. Semantic Distractor Search.}
We use the target prototype bank $\mathcal{T}$ and the current clean encoder feature map $F_t^c$ to measure whether each spatial location resembles
the target more than its surrounding background.

For each stored target prototype $m$, $d_m(u)$ measures how much location \(u\) resembles the target more than the background:
\begin{equation}
d_m(u)
=
\Big[
\cos\!\left(F_t^c(u),p_m^{+}\right)
-
\cos\!\left(F_t^c(u),p_m^{-}\right)
\Big]_{+}.
\label{eq:semantic-margin}
\end{equation}
We then normalize each $d_m$ by its maximum response and average across stored
target prototypes:
\begin{equation}
S_t(u)
=
\frac{1}{|\mathcal{T}|}
\sum_{m\in\mathcal{T}}
\frac{d_m(u)}
     {\max_v d_m(v)+\epsilon}.
\label{eq:semantic-score}
\end{equation}
$S_t$ provides a spatial map of regions that remain similar to the target across its stored appearances ~\cite{samite}. We mask the selected target using $M_t^s$ and retain the remaining spatially distinct high-response regions as semantic distractor proposals, $\mathcal{P}_t^{\mathrm{sem}}$. 

Proposals in \textbf{$\mathcal{P}_t^{\mathrm{sem}}$} capture plausible appearance-based distractors, but not the decoder's evolving memory or longer-term temporal behavior. Following Observations I and II, we therefore examine SAM2's tracking behavior to identify emerging mask-decoder vulnerabilities.

\noindent\textbf{B. Hypothesis Disagreement Observation.}
CDS compares the clean decoder's alternative mask hypotheses with the selected mask $M_t^s$. If an alternative predicts a distinct region outside $M_t^s$, we retain that region as an alternative distractor proposal in the set $\mathcal{P}_t^{\mathrm{alt}}$. These proposals are particularly informative because they expose alternative regions the decoder already considers plausible.
 However, they may be unavailable when SAM2 is confident, leading to its mask hypotheses largely agreeing.

\noindent\textbf{C. Adversarial Attention-Diffuse Search.}
CDS further examines mask-token attention along
a parallel attacked trajectory. We find that this attention begins to spread toward
off-target regions before the predicted mask visibly fails. After scaling each
attention map by its uniform spatial expectation, we define
\begin{equation}
H_t(u)=
\max_k
\log\!\left(
A_{t,k}^{a}(u)
\right),
\label{eq:attention-field}
\end{equation}
where $A_{t,k}^{a}$ denotes the attention of mask token $k$ outside the target.
Positive values of $H_t$ therefore indicate stronger attention. The resulting off-target regions enter the attention diffuse set $\mathcal{P}_t^{\mathrm{att}}$.

\begin{figure}[t]
\centering
\begin{minipage}{0.97\columnwidth}
\hrule
\vspace{2pt}
\refstepcounter{algorithm}
\textbf{Algorithm~\thealgorithm. Effective Distractor Bank Update Policy}
\vspace{2pt}
\hrule
\label{alg:distractor-bank-algo}
\vspace{3pt}

\begin{algorithmic}
\Require $\mathcal{P}_t$: distractor proposal set, $F_t^a$: attacked feature map,
$\mathcal{E}=\{e_s\}$: effective distractor bank
\Require $K$: bank capacity, $\theta_{\mathrm{merge}}$: merge threshold,
$\eta$: EMA update rate

\For{$P_t^j\in\mathcal{P}_t$}
    \State Form a prototype: $q \gets \psi(F_t^a,P_t^j)$
    \State Find its closest pair in $\mathcal{E}$:
    $i \gets \arg\max_s \cos(q,e_s)$

    \If{$\cos(q,e_i)\geq\theta_{\mathrm{merge}}$}
        \State Merge: $e_i \gets (1-\eta)e_i+\eta q$
    \ElsIf{$|\mathcal{E}|<K$}
        \State Insert $q$ into $\mathcal{E}$
    \Else
        \State Replace with the least used slot.
    \EndIf
\EndFor
\end{algorithmic}

\vspace{2pt}
\hrule
\end{minipage}
\end{figure}

\subsubsection{Effective Distractor Bank}

The three searches provide complementary distractor evidence in the form of binary masks. After removing
overlapping regions, CDS combines all available proposals into
\begin{equation}
\mathcal{P}_t=
\mathcal{P}_t^{\mathrm{sem}}
\cup
\mathcal{P}_t^{\mathrm{alt}}
\cup
\mathcal{P}_t^{\mathrm{att}}.
\label{eq:proposal-set}
\end{equation}
Much like the target prototype bank $\mathcal{T}$, an \emph{Effective Distractor Bank}
$\mathcal{E}=\{e_s\}_{s=1}^{K}$ admits a compact history of competing
appearances using $\mathcal{P}_t$ and the adversarial encoder feature $F^a_t$. Unlike the target prototype bank, whose identity is initiated by the initial prompt, these proposals carry no annotation anchors. Naive updates may therefore
split one distractor across multiple slots or merge distinct distractors. We
use conservative association and updates to maintain stable histories.

\noindent\textbf{Effective Distractor Bank Update Policy.} Each proposal in
$\mathcal{P}_t$ forms a prototype from gradient-detached $F_t^a$ for comparison
with occupied slots in $\mathcal{E}$. If its
closest slot is sufficiently similar, the two are merged through an exponential
moving average (EMA) ~\cite{ren2023focusdetailsonlinemultiobject}. Otherwise, the candidate fills an empty slot or replaces the least-used one when the bank is full. A least used slot is the one that has the least EMA update. This turns repeated evidence into a compact set of stable distractors, preventing the bank from growing with redundant regions and focusing the attack on a few \emph{effective} alternatives (See Algorithm~\ref{alg:distractor-bank-algo}).

\subsection{Attack Objectives}
\label{sec:objectives}

The attack objective set is designed to induce persistent tracking failure. Using the target prototype and effective
distractor banks accumulated over time, it suppresses foreground support on the
target, promotes CDS-selected competing regions, and corrupts
target distinctiveness along the attacked trajectory. 

\noindent\textbf{Foreground suppression.}
Given the attacked mask $M_t^a$ and the clean predicted target $M_t^s$ from the parallel clean trajectory, we
first suppress the attacked prediction $M_t^a$ over the clean target $M_t^s$:
\begin{equation}
\mathcal{L}_{\mathrm{mask}}
=
\operatorname{BCE}_{M_t^s}
\left(
M_t^a,\mathbf{0}
\right),
\label{eq:mask-loss}
\end{equation}
where $\operatorname{BCE}_{R}$ evaluates binary cross entropy only within $R$. Unlike objectives
that also suppress the background ~\cite{uapsam2, attacksam}, this leaves off-target regions available to
receive foreground support.

\noindent\textbf{Distractor-weighted foreground displacement.}
Foreground suppression removes support from the target but does not specify
where it should move. A complete displacement objective can instead promote
foreground to be over the active CDS proposals:
\begin{equation}
{\mathcal{L}}_{\mathrm{disp}}
=
\frac{1}{|\mathcal{P}_t|}
\sum_{j\in\mathcal{P}_t}
\operatorname{BCE}_{P_t^j}
\left(M_t^a,\mathbf{1}\right),
\label{eq:uniform-disp-loss}
\end{equation}
where $\mathcal{P}_t$ is the current distractor proposal set. Treating all proposals equally, however, ignores which regions repeatedly expose the strongest vulnerabilities along the attacked trajectory~\cite{Maag_2024}. Promoting multiple unrelated regions also conflicts with the tracker’s single-target nature and weakens the attack. We therefore leverage the effective distractor bank which preserves evidence from the strongest proposal trajectories, providing long-term precedent and a coherent attack direction. For each proposal $P_t^j$, we pool $z_t^j=\psi(F_t^a,P_t^j)$ and define $a_t^j$ as its maximum cosine similarity to the occupied distractor slots in $\mathcal{E}$. We then apply softmax:
\begin{equation}
w_t^j
=
\frac{e^{\,a_t^j/\tau_w}}
{\sum_{q\in\mathcal{P}_t}e^{\,a_t^q/\tau_w}},
\qquad j\in\mathcal{P}_t ,
\label{eq:proposal-routing}
\end{equation}
so that $w_t^j$ concentrates foreground on proposals most strongly
supported by the accumulated distractor evidence. $\tau_w$ denotes temperature. The resulting enhanced objective is
\begin{equation}
\mathcal{L}_{\mathrm{disp}}
=
\sum_{j\in\mathcal{P}_t}
w_t^j\,
\operatorname{BCE}_{P_t^j}
\left(M_t^a,\mathbf{1}\right).
\label{eq:disp-loss}
\end{equation}

\noindent\textbf{Multi-granular clutter corruption. }
Mask displacement biases foreground support toward competing regions, but the
target may remain sufficiently distinct
in feature space, deeming it sufficient enough for SAM2 to continue tracking the true target. We therefore contrast attacked feature maps against the effective distractor bank $\mathcal{E}$ and
the clean target foreground prototypes $\mathcal{T}^{+}=\{p_m^{+}\}$.

For an arbitrary feature $q$, we directly define a multi-positive contrastive
loss as
\begin{equation}
\mathcal{L}_{\mathrm{con},q}
=
-\log
\frac{
\displaystyle
\sum_{e\in\mathcal{E}}
e^{\cos(q,e)/\tau_c}
}{
\displaystyle
\sum_{e\in\mathcal{E}}
e^{\cos(q,e)/\tau_c}
+
\sum_{p\in\mathcal{T}^{+}}
e^{\cos(q,p)/\tau_c}
}.
\label{eq:clutter-contrast}
\end{equation}

Minimizing this loss pulls the representation toward accumulated distractors
while pushing it away from the clean target history ~\cite{khosla2020supervised, dai2022unimoco}. $\tau_c$ denotes contrastive temperature.

From $F_t^a$, we form a global query $q_t^G$ by average pooling and local
queries $q_t^u$ from its patch features, and evaluate the contrast at both
granularities.
The resulting objective is
\begin{equation}
\mathcal{L}_{\mathrm{clut}}
=
\eta_G\,\mathcal{L}_{\mathrm{con},q_t^G}
+
\eta_L
\sum_u
\alpha_t^u\,
\mathcal{L}_{\mathrm{con},q_t^u}.
\end{equation}
where $\alpha_t^u$ is weight over target-region features, focusing the local contrastive loss on the target region to more effectively weaken its representation. The complete objective is
\begin{equation}
\mathcal{L}
=
\lambda_M\mathcal{L}_{\mathrm{mask}}
+
\lambda_D\mathcal{L}_{\mathrm{disp}}
+
\lambda_C\mathcal{L}_{\mathrm{clut}}.
\label{eq:full-objective}
\end{equation}
where $\lambda_M$, $\lambda_D$, $\lambda_C$ are loss coefficients.
Using the computed gradient
${g}_{\mathcal{W}}$, the perturbation is a signed gradient update ~\cite{goodfellow2014explaining, madry2017towards}:
\begin{equation}
\delta
\leftarrow
\Pi_{\lVert\delta\rVert_\infty\leq\epsilon}
\left(
\delta
-
\alpha\operatorname{sign}
\left(
{g}_{\mathcal{W}}
\right)
\right),
\label{eq:update}
\end{equation}
where $\alpha$ is the step size and the scope of $\delta$ follows one of the following perturbation settings:
\begin{itemize}
\item \textbf{Video Perturbation (VP):}
A separate perturbation $\delta_v$ is optimized for each video $v$.
Once optimized, the same perturbation is applied across all frames of that video.

\item \textbf{One-shot Perturbation (OP):}
We consider a more practical setting in which the adversary can manipulate only the initially prompted frame \cite{jiang2023exploring, chen2020one}. The attacked sequence is defined as
\begin{equation}
\widetilde{x}_t=
\begin{cases}
\operatorname{clip}(x_0+\delta_o), & t=0,\\
x_t, & t>0,
\end{cases}
\qquad
\|\delta_o\|_{\infty}\leq\epsilon .
\label{eq:op-setting}
\end{equation}
At each optimization step, clean and attacked trajectories are initialized from
the same prompt and propagated over a rollout of $N$ frames to feed banks
$\mathcal{T}$ and $\mathcal{E}$. Here, $N$ defines only the observation horizon
for bank construction and introduces no additional prompts, annotations, or
perturbations to later frames. Both banks are detached and used as temporal
context for the attack objective, while gradients are propagated only to
$\delta_o$. Once optimized, $\delta_o$ is applied once to the prompted frame,
after which the tracker proceeds on the remaining clean frames without further
adversarial intervention.
\end{itemize}
Importantly, in both settings, only the initial mask and video frames are known to the adversary.

\section{Experiments}
\label{sec:experiments}

\noindent \textbf{Setup protocol.}
We evaluate VP and OP on visual object tracking (VOT) and video object segmentation (VOS). For VOT, we use LaSOT~\cite{lasot}, UTB180~\cite{utb180}, GOT-10k~\cite{got10k}, and LaSOT-ext~\cite{lasot}, including the distractor-distilled DiDi~\cite{dam4sam}. Each is initialized by the ground-truth box. Following VOTS~\cite{kristan2023first}, we macro-average tracking quality $Q$, accuracy $Acc$, robustness $Rob$, DRE, and NRE across sequences. For VOS, we use DAVIS~2017 (DAVIS)~\cite{davis2017}, YouTube-VOS~2018 (YT)~\cite{youtubevos2018}, MOSEv2~\cite{mose}, and the SA-V test set~\cite{sam2}. The ground-truth mask initializes frame 0, and we report the standard $\mathcal{J}\&\mathcal{F}$. During optimization, OP may use $N\in\{1,10,20\}$ rollout frames, while its evaluation is on the complete sequence. We use the official SAM2.1 checkpoint and none of the dataset sequences were used to train it. VP uses $\epsilon=8/255$ and $\alpha=2/255$ while OP uses $\epsilon=4/255$ and $\alpha=1/255$. Both use 10 iterations, and a weight of $1.0$ for all loss terms. We cap $\mathcal{T}$ and $\mathcal{E}$ at 20 and 10 prototypes, respectively, and set $\theta_{\mathrm{merge}}$ 0.85 and the EMA rate $\eta$ 0.1. Single-image and universal baselines retain their defining objectives and mechanisms when adapted to sample-wise VP and OP. Stronger attacks correspond to lower $Q$, accuracy, robustness, and $\mathcal{J}\&\mathcal{F}$, and higher DRE and NRE. Experiments run on an NVIDIA A100 80GB GPU. Further protocol and baseline adaptation details are provided in Appendix~\ref{supp:implementation}.

\begin{table}[t]
    \centering
    \caption{Video (VP) and One-shot (OP) attack perturbation mechanisms on SAM2.1-L.
    VOS results are given in $\mathcal{J}\&\mathcal{F}$. Arrows indicate a stronger attack.}
    \label{tab:attack_results}
    \scriptsize
    \setlength{\tabcolsep}{1.5pt}

    \resizebox{\columnwidth}{!}{%
    \begin{tabular}{@{}llccccc|ccccc@{}}
        \toprule
        & &
        \multicolumn{5}{c|}{VOT}
        & \multicolumn{5}{c}{VOS} \\
        \cmidrule(lr){3-7}
        \cmidrule(lr){8-12}

        Setting & Method
        & $Q\downarrow$
        & Acc.$\downarrow$
        & Rob.$\downarrow$
        & DRE\%$\uparrow$
        & NRE\%$\uparrow$
        & DAVIS$\downarrow$
        & YT$\downarrow$
        & MOSEv2$\downarrow$
        & SA-V$\downarrow$
        & Avg.$\downarrow$ \\
        \midrule

        & \textcolor{gray}{Clean inference}
        & \textcolor{gray}{0.836}
        & \textcolor{gray}{0.842}
        & \textcolor{gray}{0.989}
        & \textcolor{gray}{0.68}
        & \textcolor{gray}{0.42}
        & \textcolor{gray}{91.92}
        & \textcolor{gray}{86.66}
        & \textcolor{gray}{82.57}
        & \textcolor{gray}{87.43}
        & \textcolor{gray}{87.15} \\

        \addlinespace[2pt]

        \multirow{8}{*}{VP}

        & AttackSAM~\cite{attacksam}
        & 0.798 & 0.807 & 0.981 & 1.10 & 0.83
        & 80.34 & 75.92 & 72.88 & 77.10 & 76.56 \\

        & DarkSAM
        & 0.772 & 0.784 & 0.976 & 1.38 & 1.05
        & 77.42 & 70.18 & 69.55 & 73.84 & 72.75 \\

        & CPA~\cite{long2025robust}
        & 0.756 & 0.769 & 0.972 & 1.56 & 1.22
        & 74.90 & 67.62 & 66.81 & 71.33 & 70.17 \\

        & SegPGD~\cite{segpgd}
        & 0.735 & 0.748 & 0.966 & 1.89 & 1.48
        & 72.88 & 64.95 & 65.72 & 69.04 & 68.15 \\

        & UAD~\cite{uad}
        & 0.718 & 0.732 & 0.962 & 2.04 & 1.70
        & 71.56 & 63.72 & 64.40 & 67.85 & 66.88 \\
        
        & FAN~\cite{fan}
        & 0.701 & 0.720 & 0.959 & 2.16 & 1.94
        & 69.31 & 60.44 & 61.87 & 65.26 & 64.22 \\

        & UAP-SAM2~\cite{uapsam2}
        & 0.689
        & 0.709
        & 0.954
        & 2.47
        & 2.16
        & 62.81 & 51.37 & 62.55 & 57.94 & 58.67 \\

        & \textbf{DistractAttack}
        & \textbf{0.420}
        & \textbf{0.554}
        & \textbf{0.734}
        & \textbf{10.83}
        & \textbf{15.77}
        & \textbf{59.38}
        & \textbf{44.46}
        & \textbf{49.42}
        & \textbf{50.62}
        & \textbf{50.97} \\

        \midrule
        \addlinespace[2pt]

        \multirow{12}{*}{OP}

        & DarkSAM
        & 0.803 & 0.811 & 0.971 & 1.15 & 1.78
        & 61.48 & 53.21 & 58.75 & 56.84 & 57.57 \\

        & AttackSAM~\cite{attacksam}
        & 0.816 & 0.822 & 0.976 & 1.04 & 1.36
        & 66.12 & 57.44 & 62.08 & 60.35 & 61.50 \\

        & CPA~\cite{long2025robust}
        & 0.791 & 0.802 & 0.865 & 1.29 & 12.21
        & 56.72 & 47.63 & 54.30 & 51.18 & 52.46 \\

        & SegPGD~\cite{segpgd}
        & 0.775 & 0.787 & 0.952 & 1.52 & 3.26
        & 50.83 & 41.57 & 48.66 & 45.21 & 46.57 \\

        & UAD~\cite{uad}
        & 0.764 & 0.778 & 0.668 & 1.88 & 31.31
        & 47.35 & 38.72 & 45.19 & 42.60 & 43.47 \\

        & FAN~\cite{fan}
        & 0.354 & 0.369 & 0.520 & 2.10 & 45.87
        & 43.92 & 34.88 & 42.75 & 39.46 & 40.25 \\

        & UAP-SAM2~\cite{uapsam2}
        & 0.245
        & 0.266
        & 0.481
        & 1.39
        & 50.54
        & 36.84 & 26.95 & 37.88 & 31.75 & 33.36 \\

        & ARA-Attack~\cite{li2023adversarial}
        & 0.213 & 0.235 & 0.423 & 2.74 & 54.92
        & 32.14 & 22.81 & 34.57 & 28.93 & 29.61 \\

        & VOSPGD~\cite{jiang2023exploring}
        & 0.172
        & 0.193
        & 0.281
        & 2.55
        & \textbf{69.45}
        & 19.43 & 16.59 & 37.31 & 24.73 & 24.52 \\

        & \textbf{DistractAttack$_{N=1}$}
        & 0.131
        & 0.148
        & 0.270
        & 8.71
        & 64.34
        & 16.45
        & 11.12
        & 33.34
        & 27.09
        & 22.00 \\

        & \textbf{DistractAttack$_{N=10}$}
        & 0.101
        & 0.126
        & 0.278
        & 9.09
        & 63.14
        & 16.11
        & 10.12
        & 31.24
        & 25.76
        & 20.81 \\

        & \textbf{DistractAttack$_{N=20}$}
        & \textbf{0.076}
        & \textbf{0.115}
        & \textbf{0.250}
        & \textbf{9.28}
        & 65.73
        & \textbf{15.85}
        & \textbf{9.38}
        & \textbf{31.02}
        & \textbf{21.99}
        & \textbf{19.56} \\

        \bottomrule
    \end{tabular}%
    }
\end{table}

\newcommand{\atkclean}[2]{%
    \begin{tabular}[c]{@{}c@{}}
        #1\\[-1.2pt]
        {\color{gray}\tiny (#2)}
    \end{tabular}%
}

\begin{figure}[!b]
    \centering
    \includegraphics[width=\columnwidth]{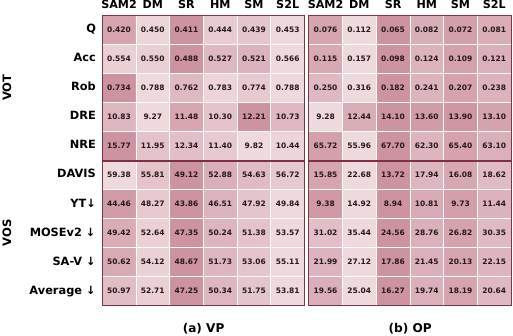}
    \caption{\textbf{DistractAttack across tracker variants} under (a) VP and
(b) OP. DM, SR, HM, SM, and S2L denote DAM4SAM, SAMURAI, HiM2SAM, SAMITE,
and SAM2Long. VOS entries report $\mathcal{J}\&\mathcal{F}$. }
    \label{fig:VP_trackers}
\end{figure}

\begin{figure*}[t]
    \centering

    \begin{subfigure}[t]{0.21\textwidth}
        \centering
        \includegraphics[width=\linewidth]{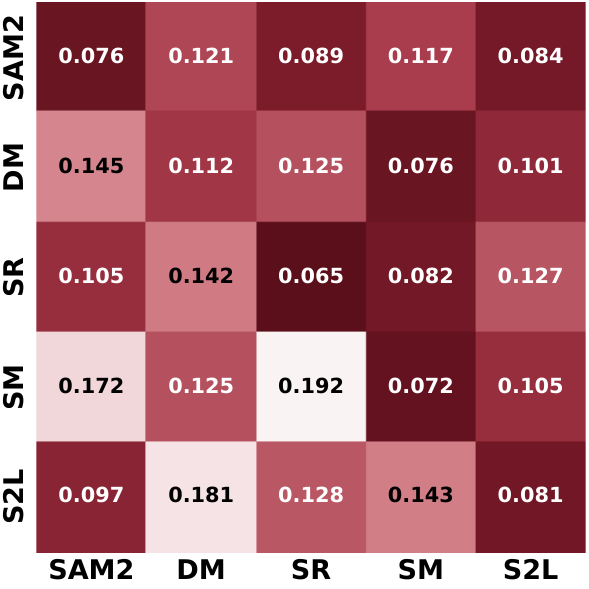}
        \caption{Cross-variant OP}
        \label{fig:transfer_variant_op}
    \end{subfigure}%
    \hspace{0.012\textwidth}%
    \begin{subfigure}[t]{0.21\textwidth}
        \centering
        \includegraphics[width=\linewidth]{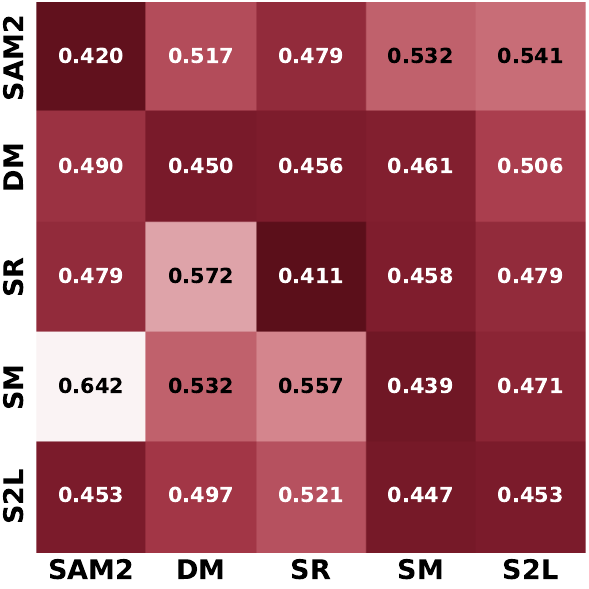}
        \caption{Cross-variant VP}
        \label{fig:transfer_variant}
    \end{subfigure}%
    \hspace{0.012\textwidth}%
    \begin{subfigure}[t]{0.21\textwidth}
        \centering
        \includegraphics[width=\linewidth]{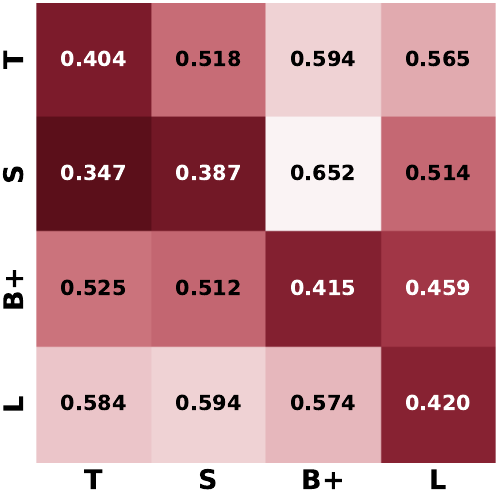}
        \caption{Cross-size}
        \label{fig:transfer_size}
    \end{subfigure}%
    \hspace{0.012\textwidth}%
    \begin{subfigure}[t]{0.21\textwidth}
        \centering
        \includegraphics[width=\linewidth]{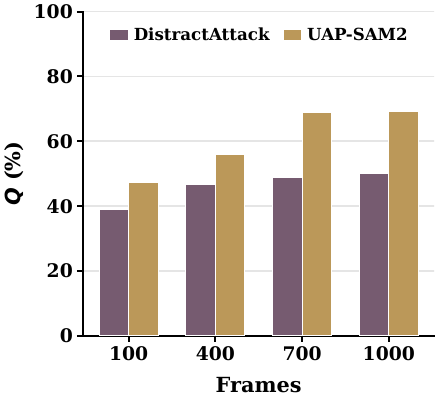}
        \caption{Cross-time}
        \label{fig:transfer_dataset}
    \end{subfigure}

\caption{\textbf{Black-box transferability of DistractAttack.}
Panels show cross-variant transfer under (a) OP and (b) VP, (c) cross-size
transfer among SAM2.1 models, and (d) zero-shot temporal transfer. Rows denote
source models and columns denote targets. Lower off-diagonal $Q$ indicates
stronger transfer. In (d), perturbations optimized on frames 0--100 are evaluated on longer windows}
\label{fig:transferability}
\end{figure*}

\begin{figure*}[t]
  \centering

  \definecolor{qualblue}{RGB}{35,76,112}

  \newcommand{\qualImageSix}[1]{%
    \raisebox{-0.5\height}{%
      \includegraphics[
        width=\dimexpr(\linewidth-18pt)/6\relax
      ]{#1}%
    }%
  }

  \newcommand{\qualRowSix}[7]{%
    \noindent\hbox to \linewidth{%
      \makebox[12pt][c]{%
        \rotatebox[origin=c]{90}{\scriptsize\bfseries #1}%
      }\hspace{1pt}%
      \qualImageSix{#2}\hspace{1pt}%
      \qualImageSix{#3}\hspace{1pt}%
      \qualImageSix{#4}\hspace{1pt}%
      \qualImageSix{#5}\hspace{1pt}%
      \qualImageSix{#6}\hspace{1pt}%
      \qualImageSix{#7}\hss
    }\par\vspace{0.5pt}%
  }

  \resizebox{0.82\textwidth}{!}{%
    \begin{minipage}{\textwidth}

      {\scriptsize\bfseries
      (a) Video Object Segmentation (VOS)}
      \par\vspace{-1pt}

   \qualRowSix{{\tiny VOSPGD~OP}}%
        {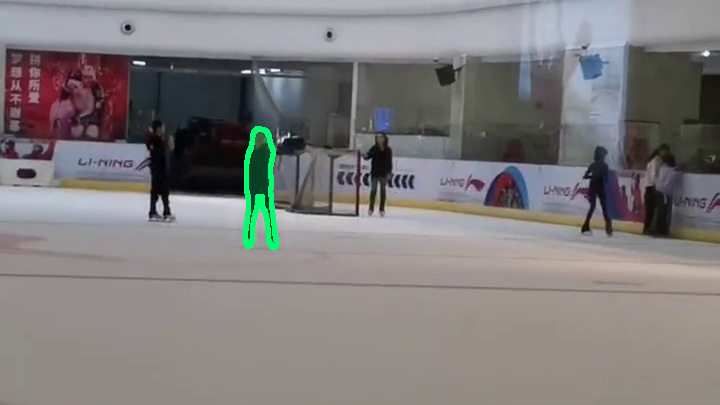}%
        {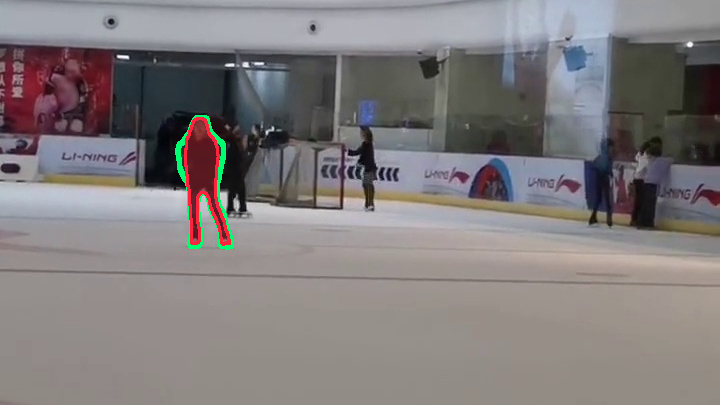}%
        {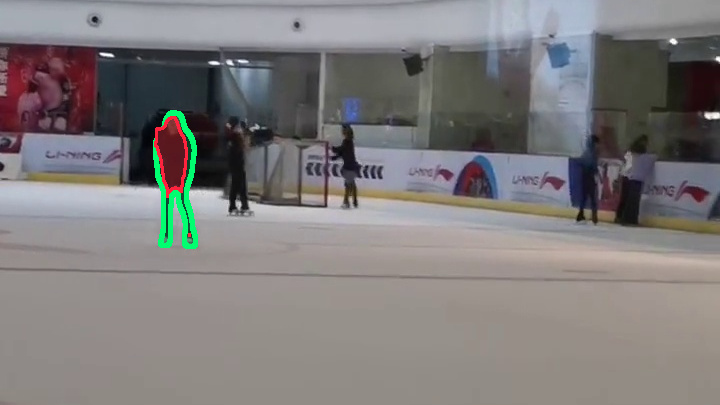}%
        {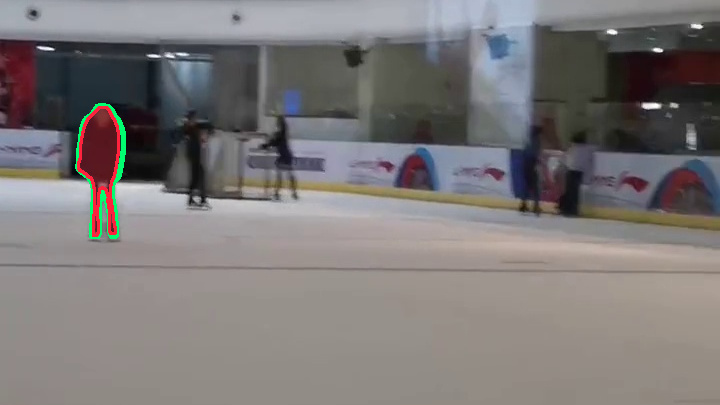}%
        {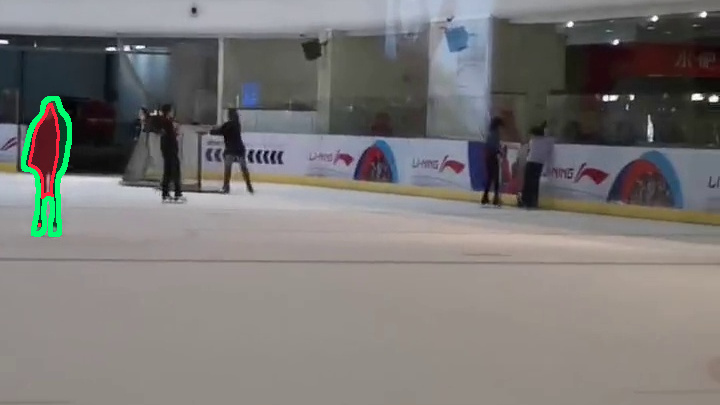}%
        {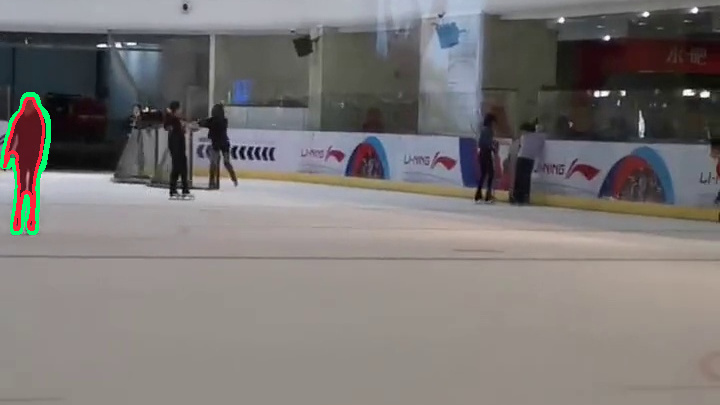}

        \qualRowSix{Ours~OP}%
        {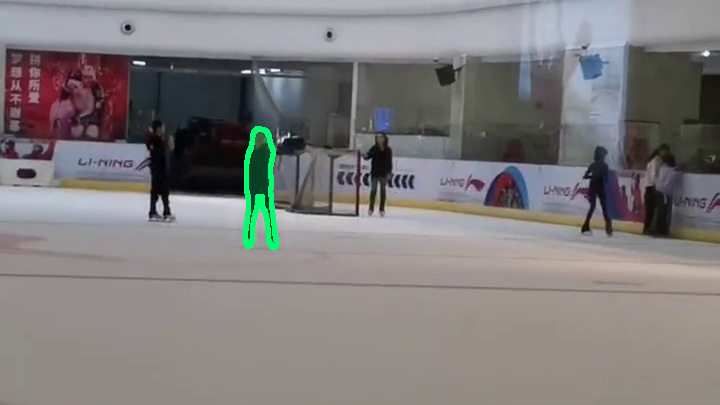}%
        {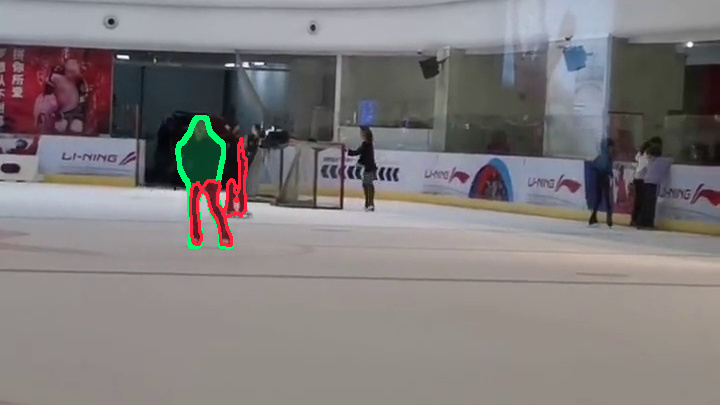}%
        {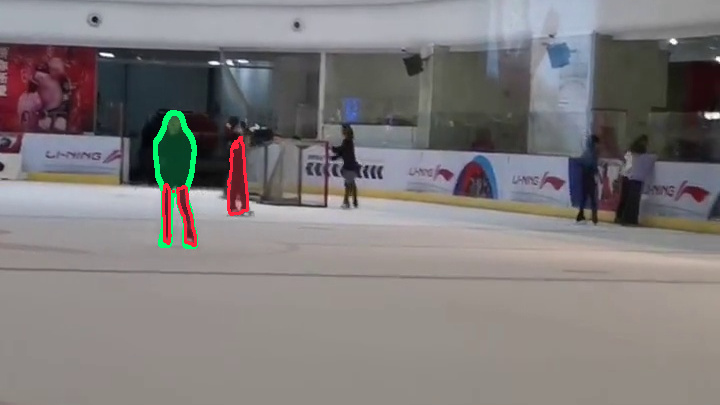}%
        {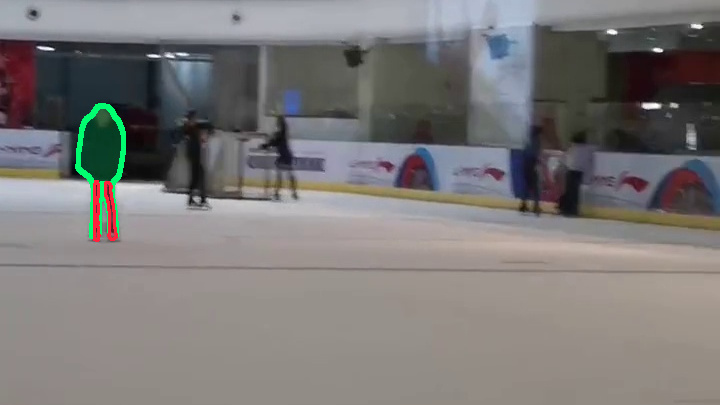}%
        {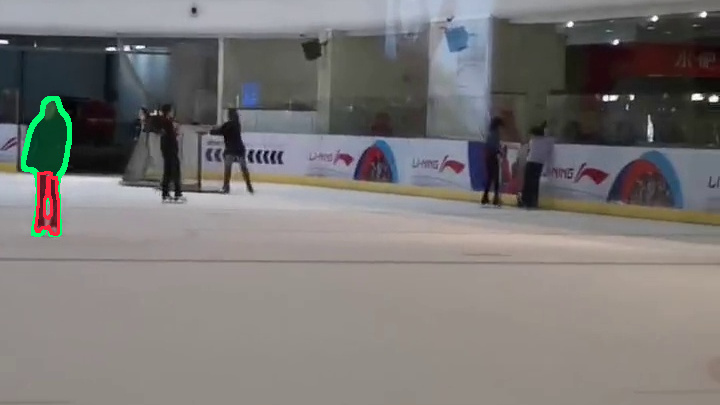}%
        {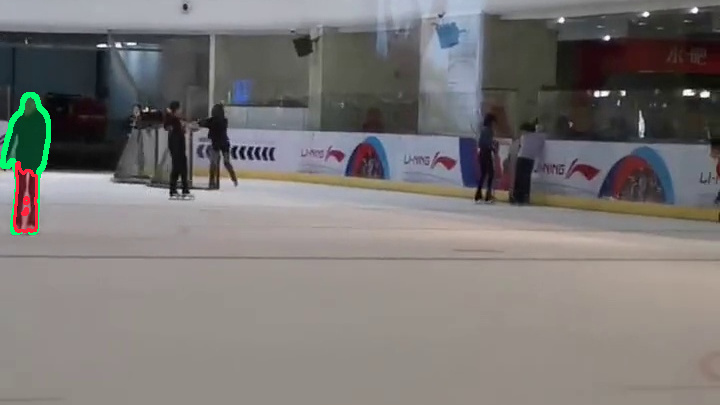}

        \qualRowSix{Ours~VP}%
        {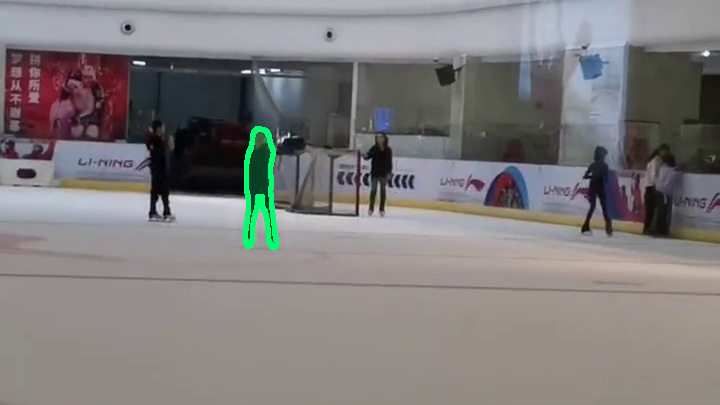}%
        {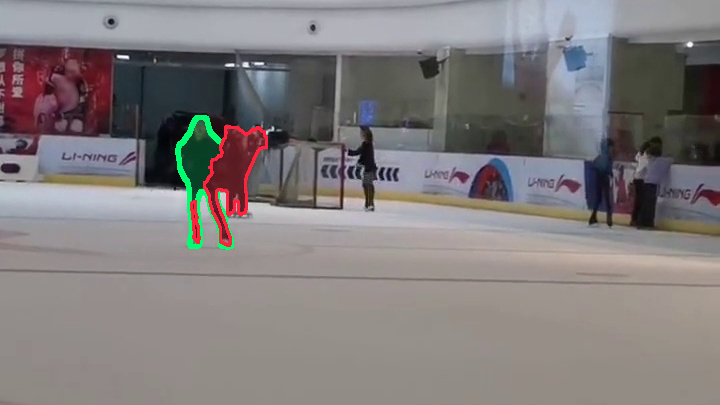}%
        {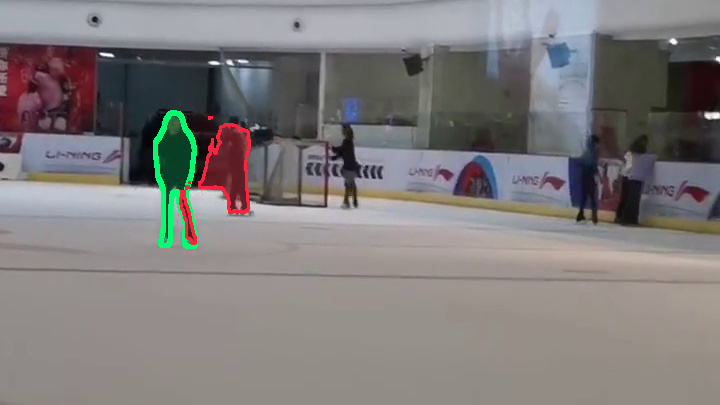}%
        {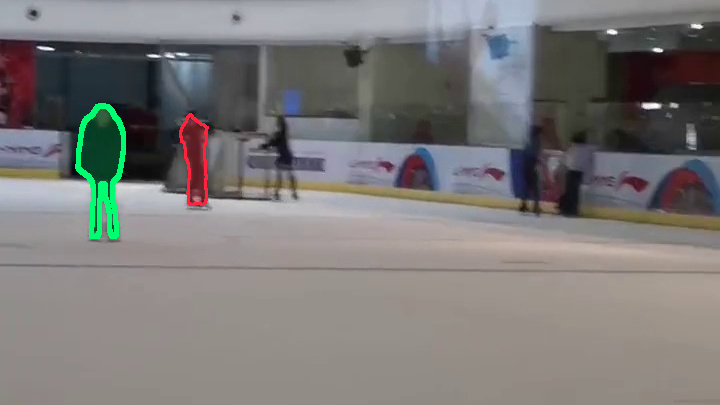}%
        {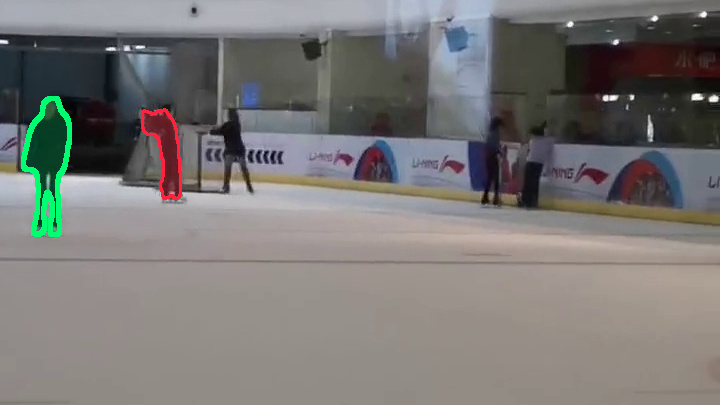}%
        {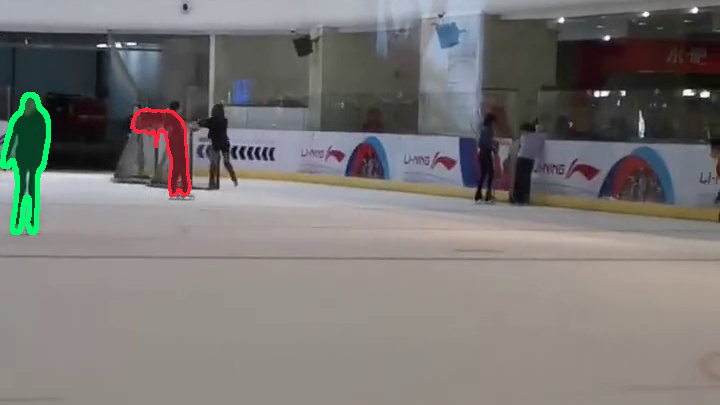}

      \vspace{0.5pt}

      {\scriptsize\bfseries
      (b) Visual Object Tracking (VOT)}
      \par\vspace{-1pt}

      \qualRowSix{UAD~VP}%
        {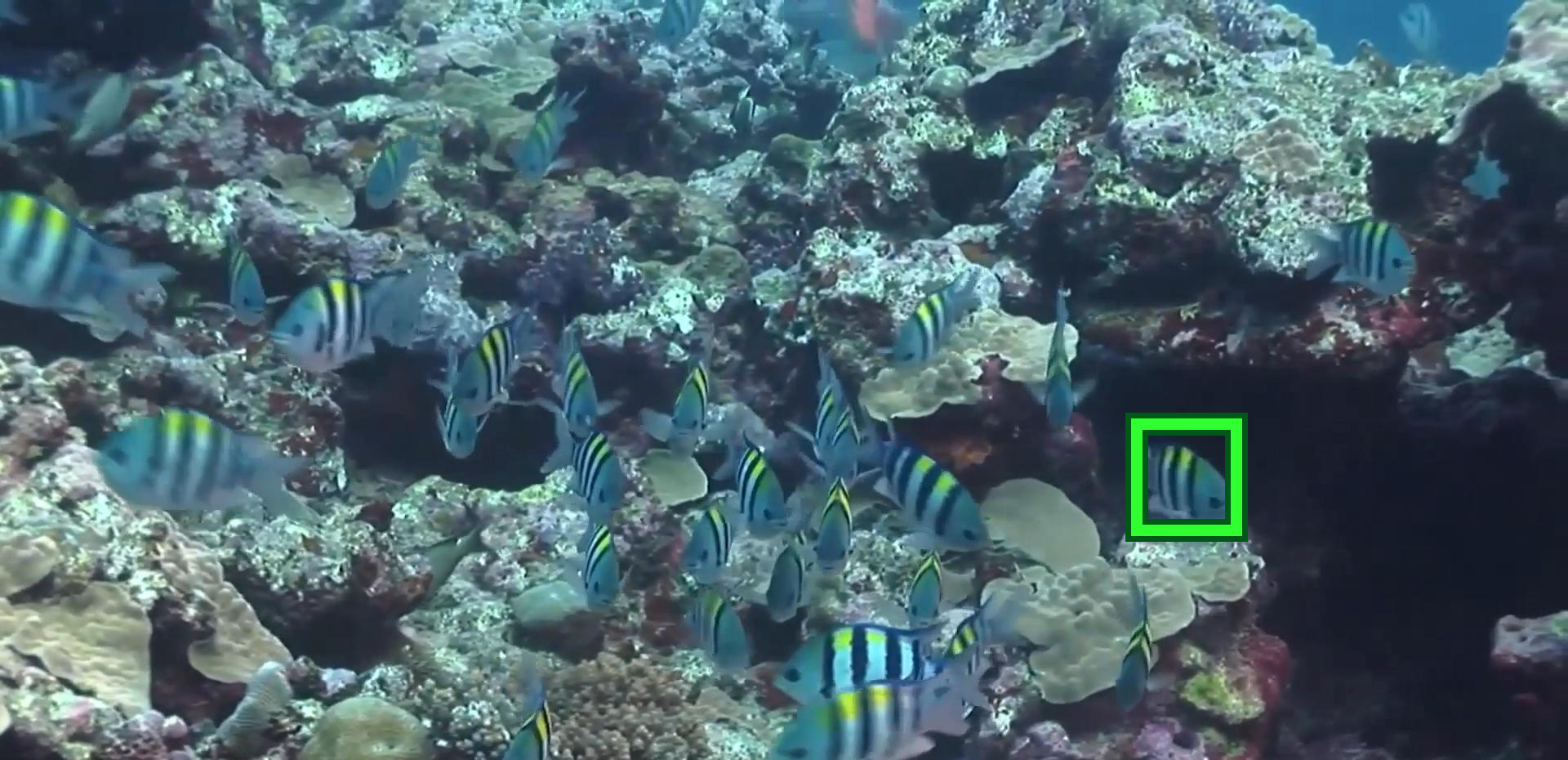}%
        {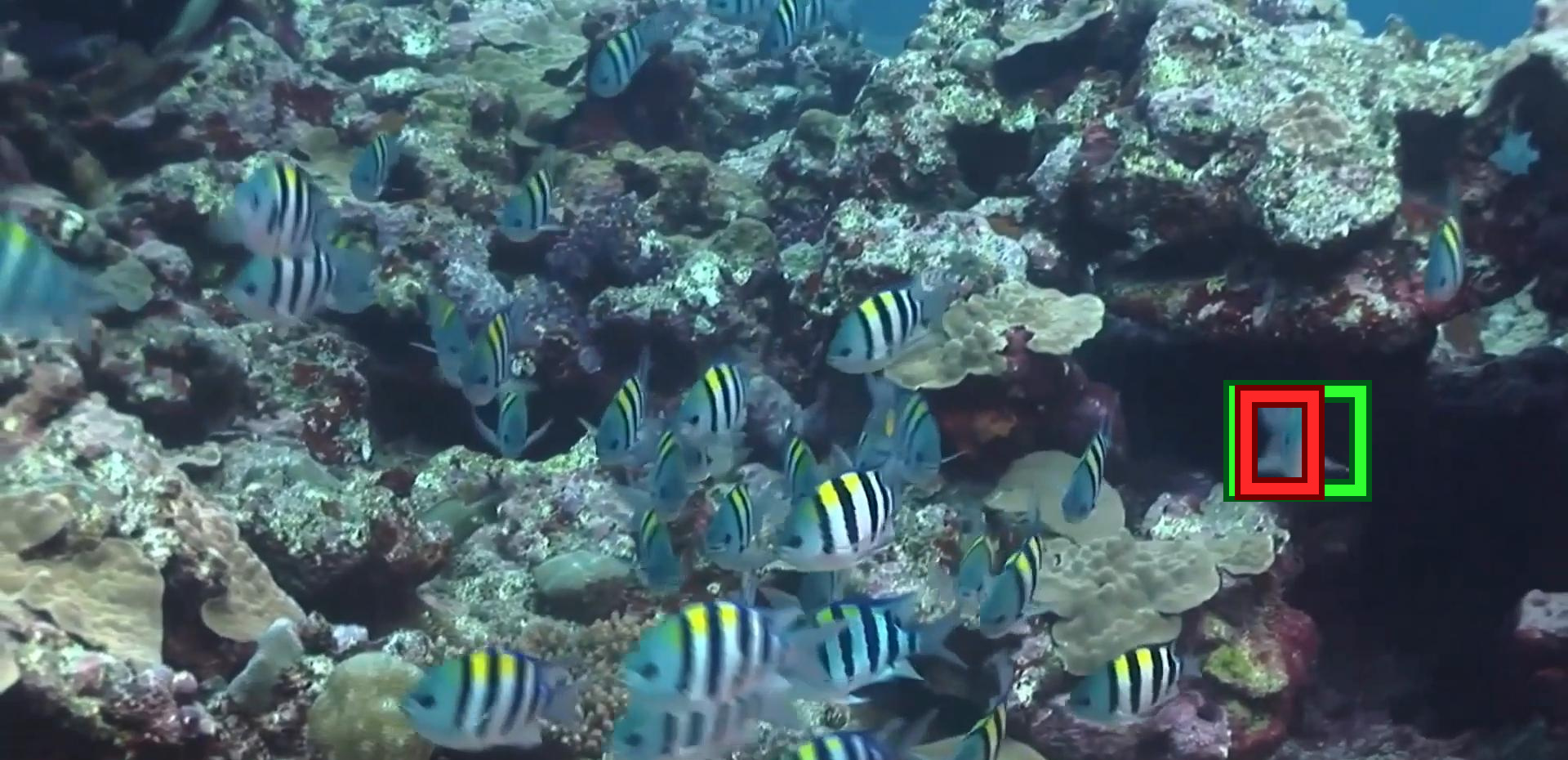}%
        {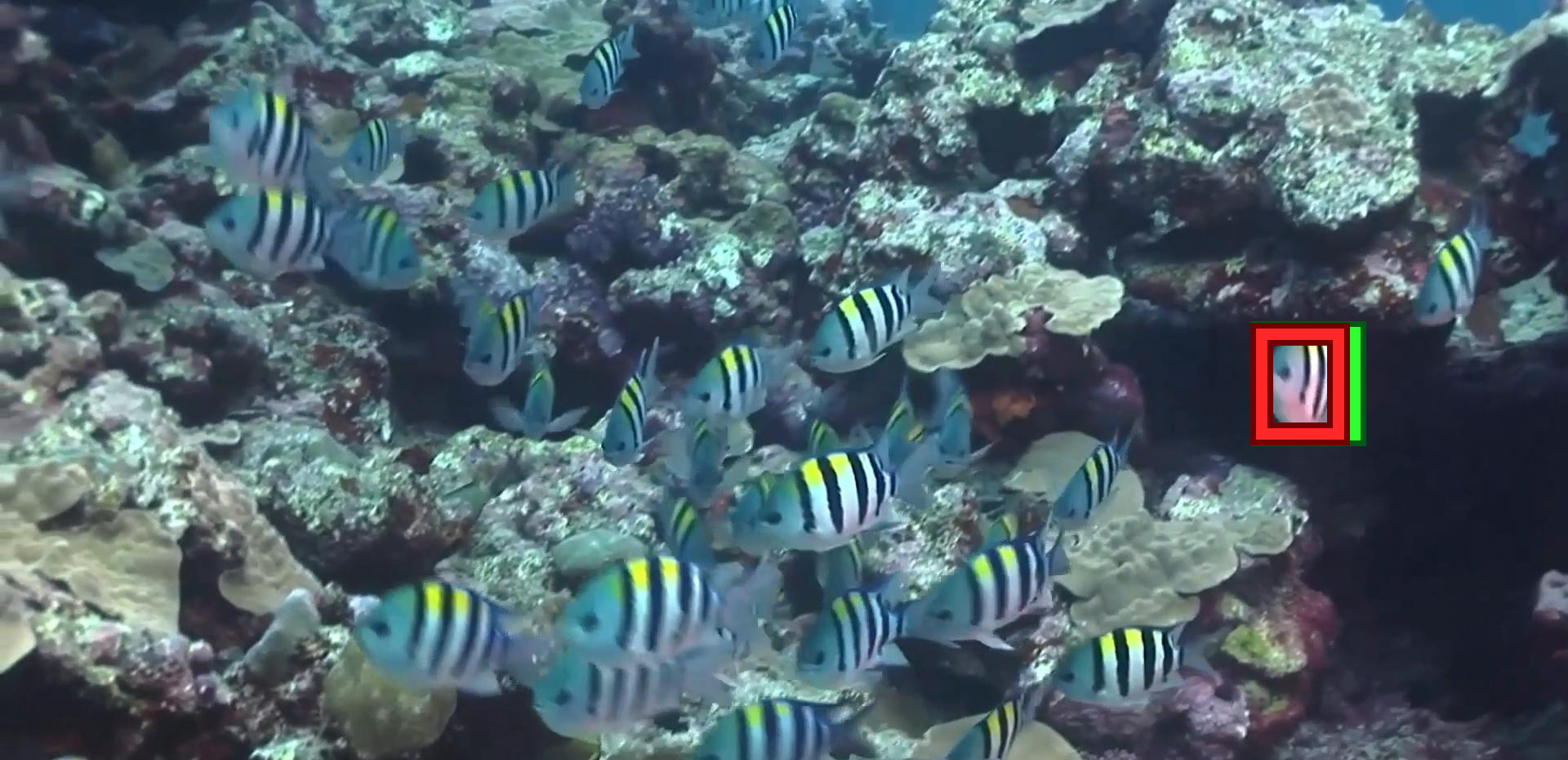}%
        {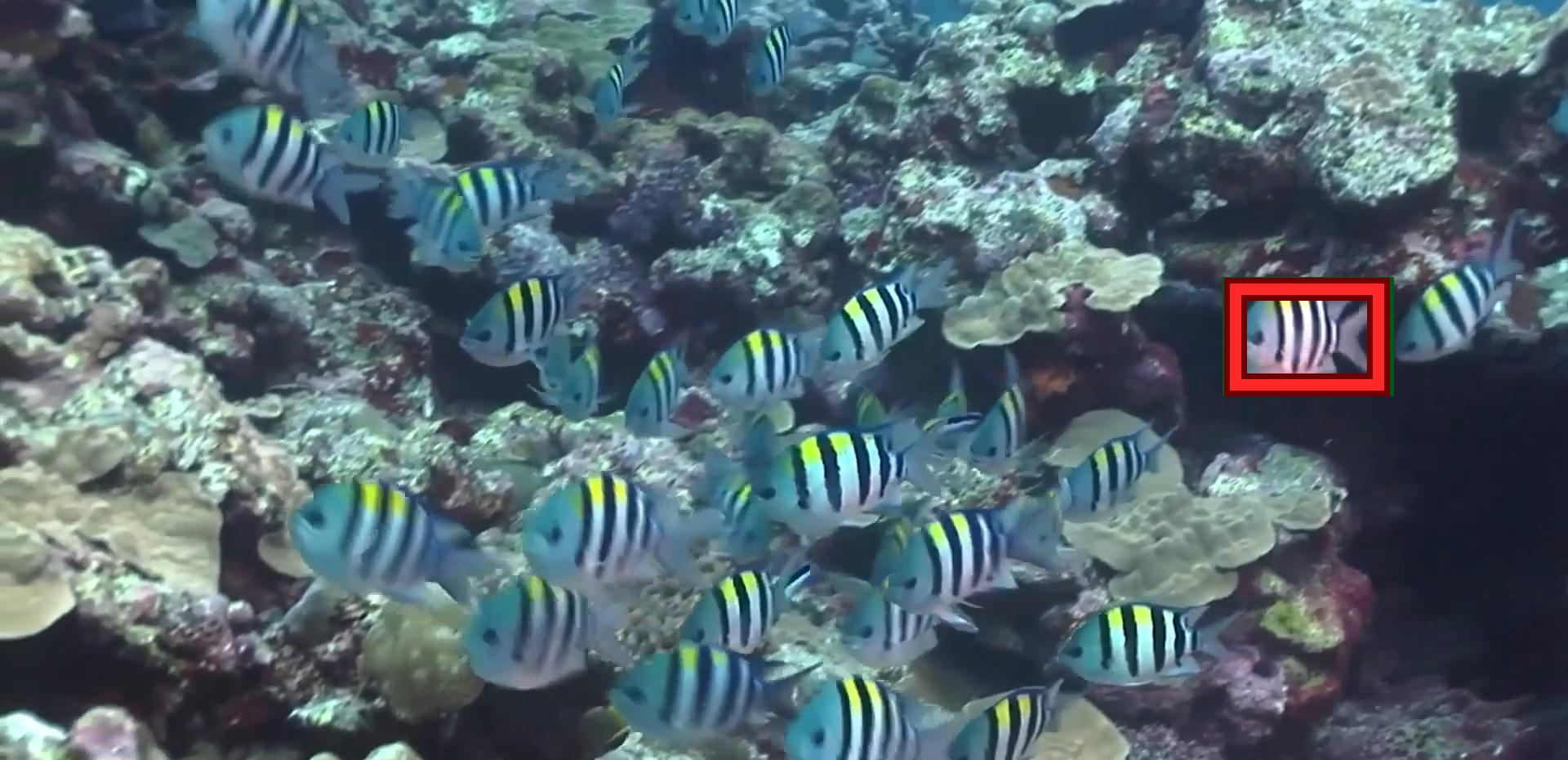}%
        {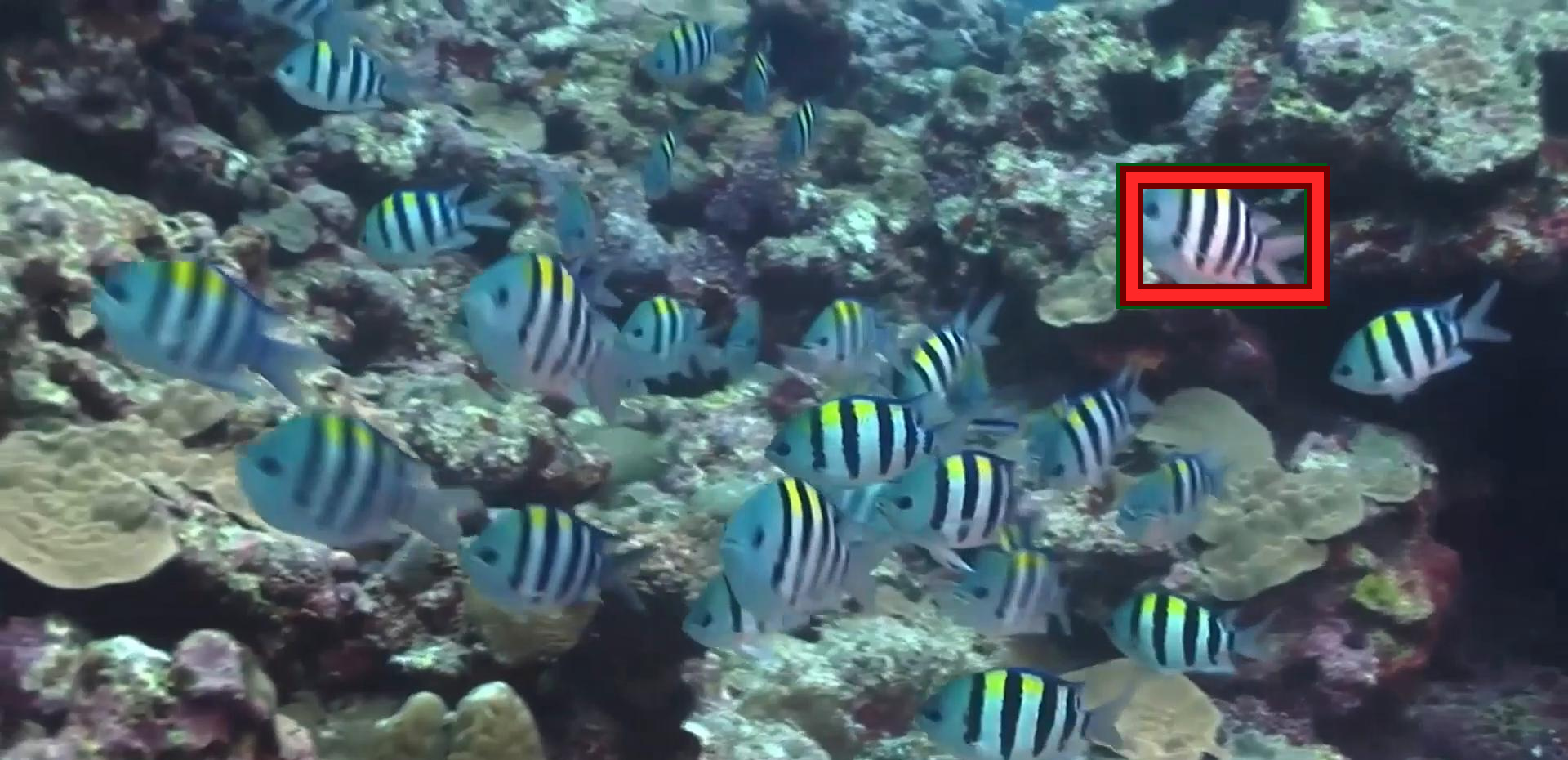}%
        {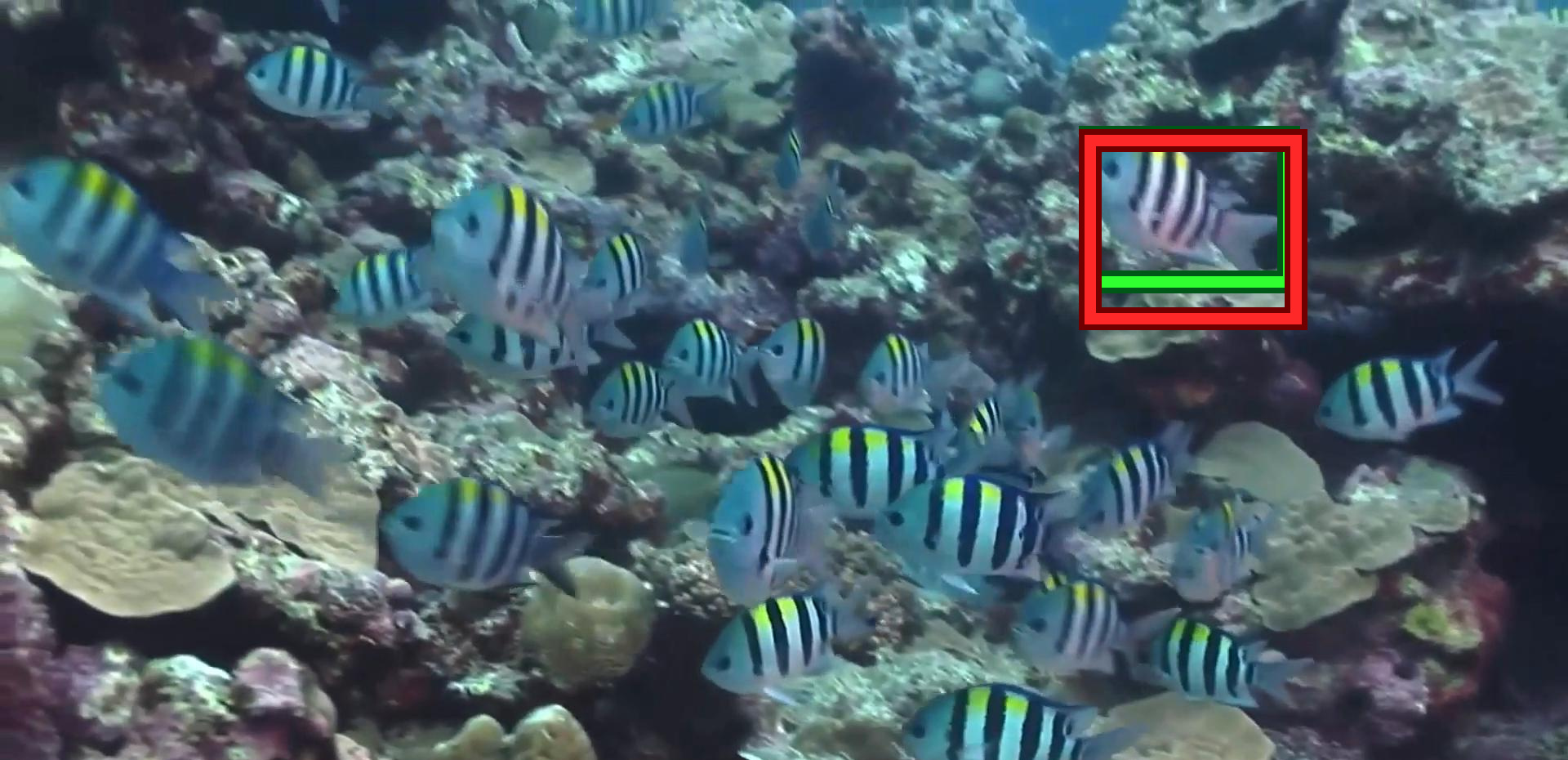}

      \qualRowSix{{\tiny UAP-SAM2~VP}}%
        {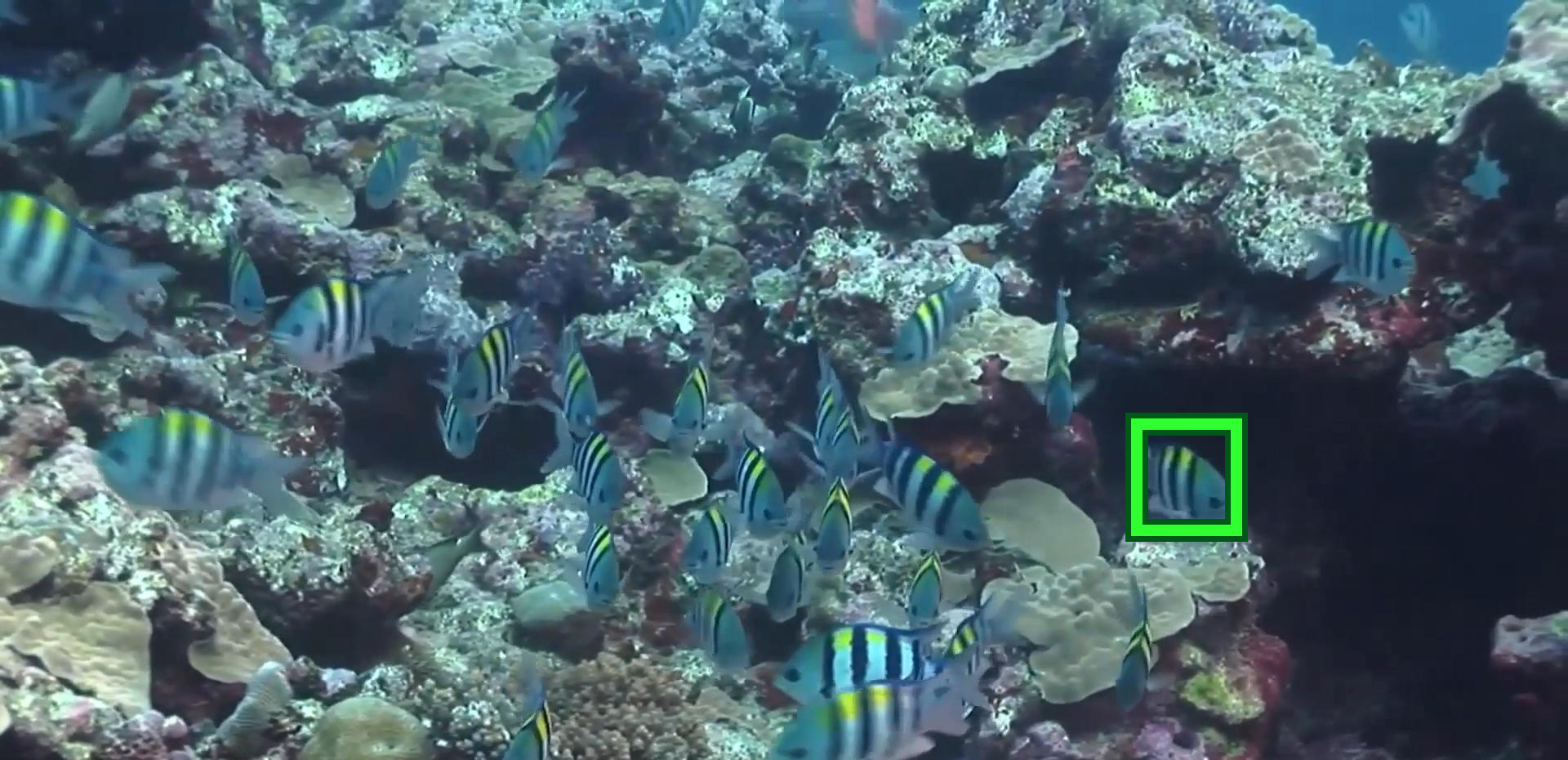}%
        {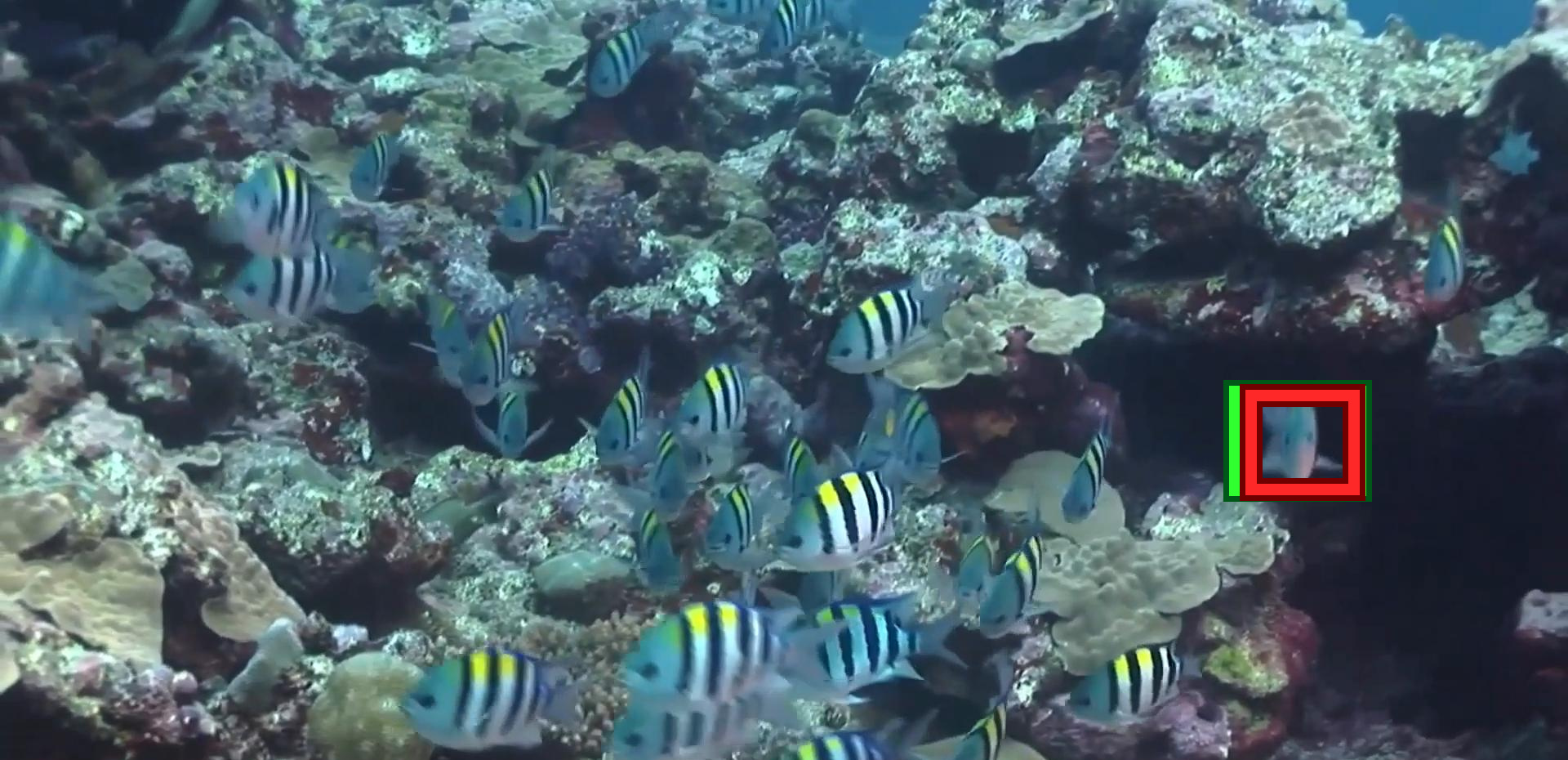}%
        {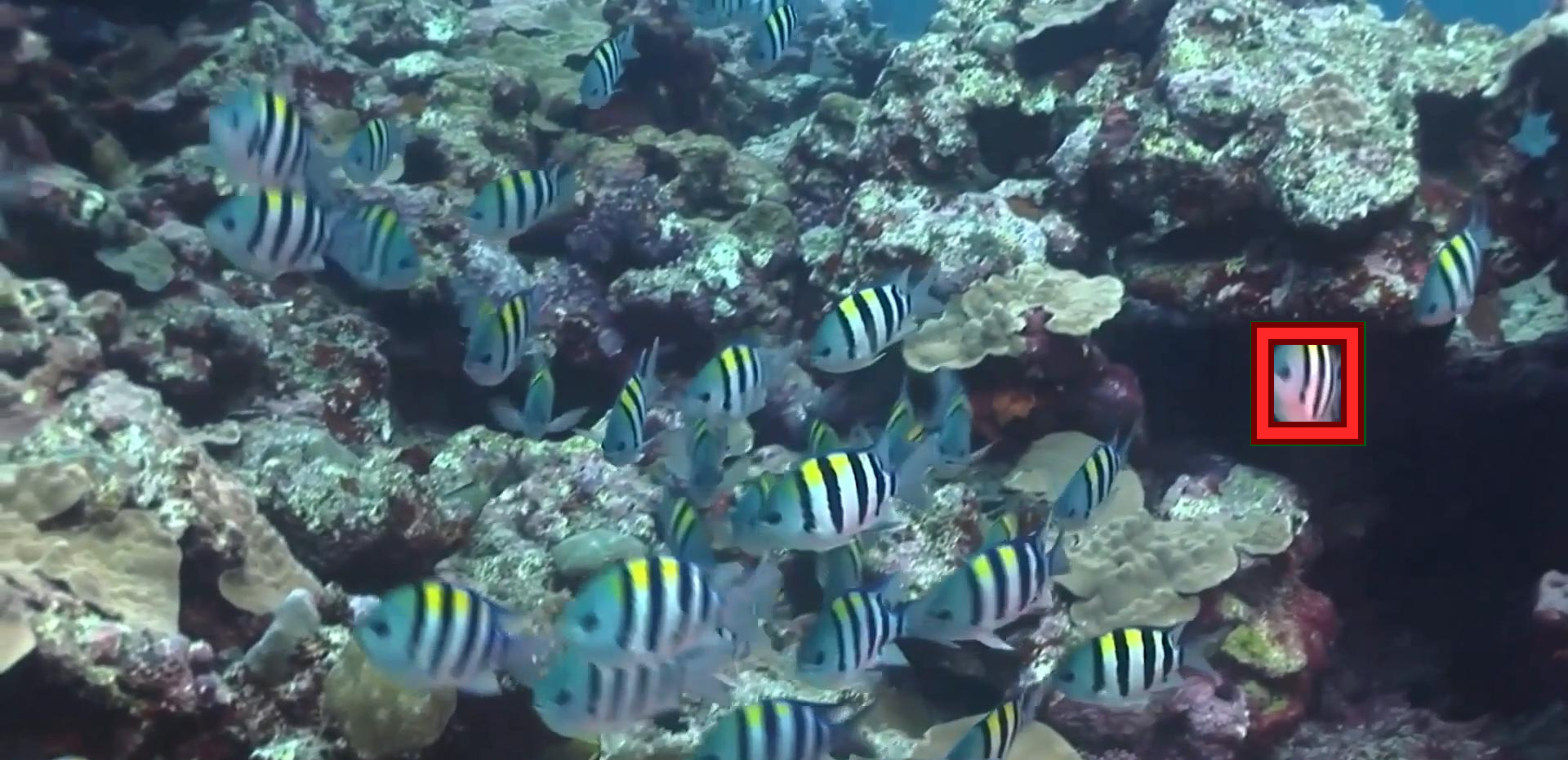}%
        {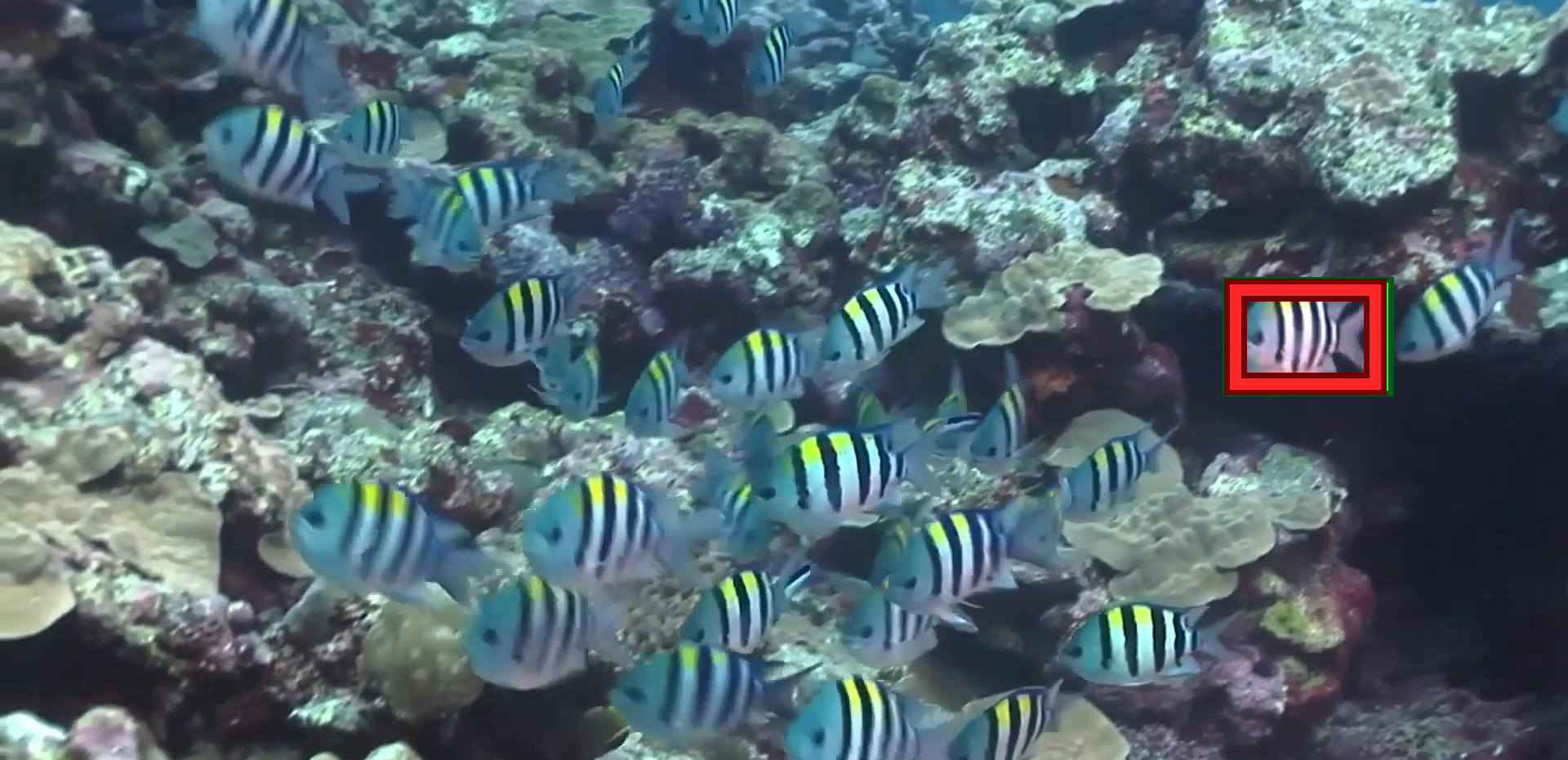}%
        {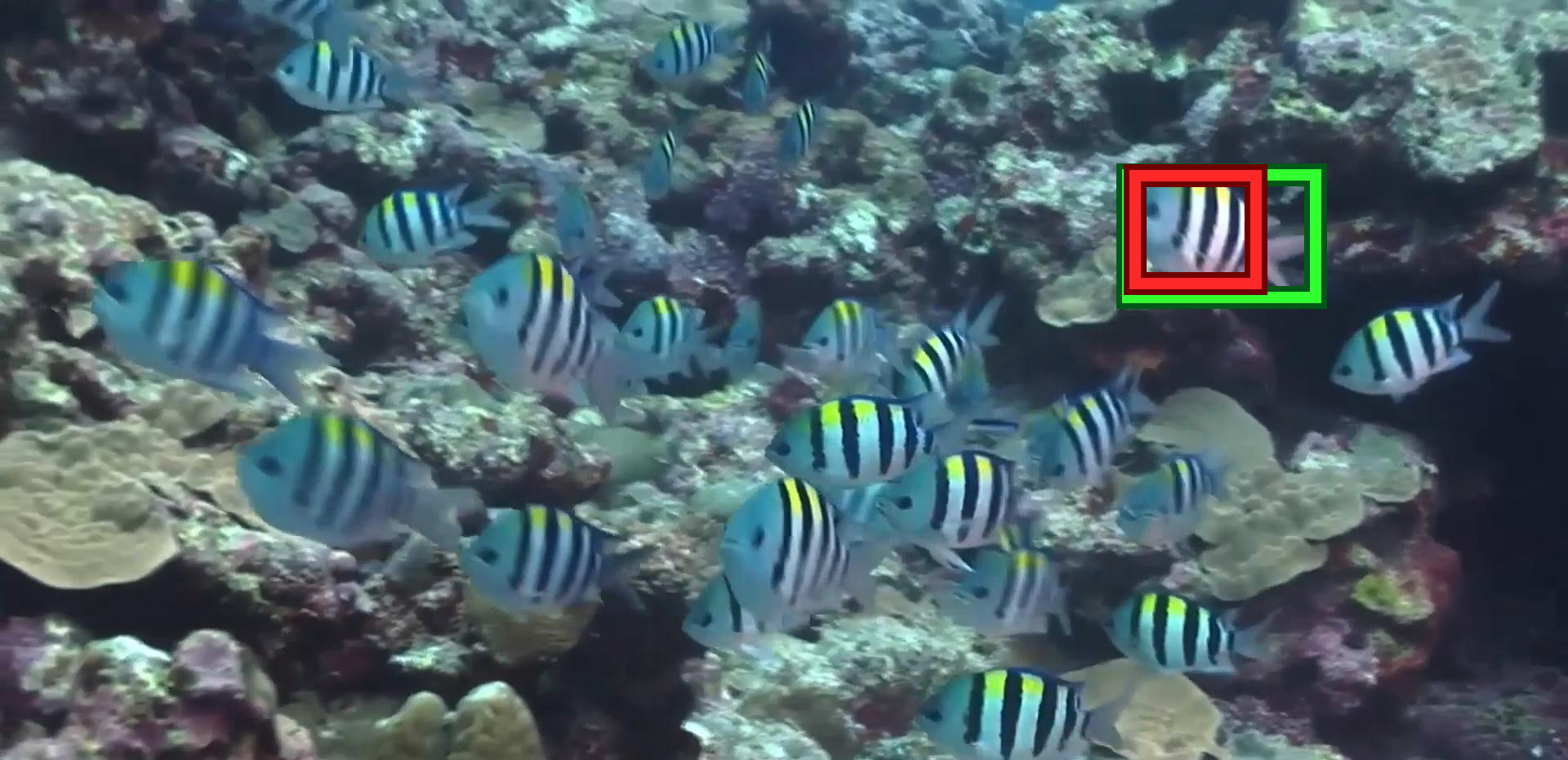}%
        {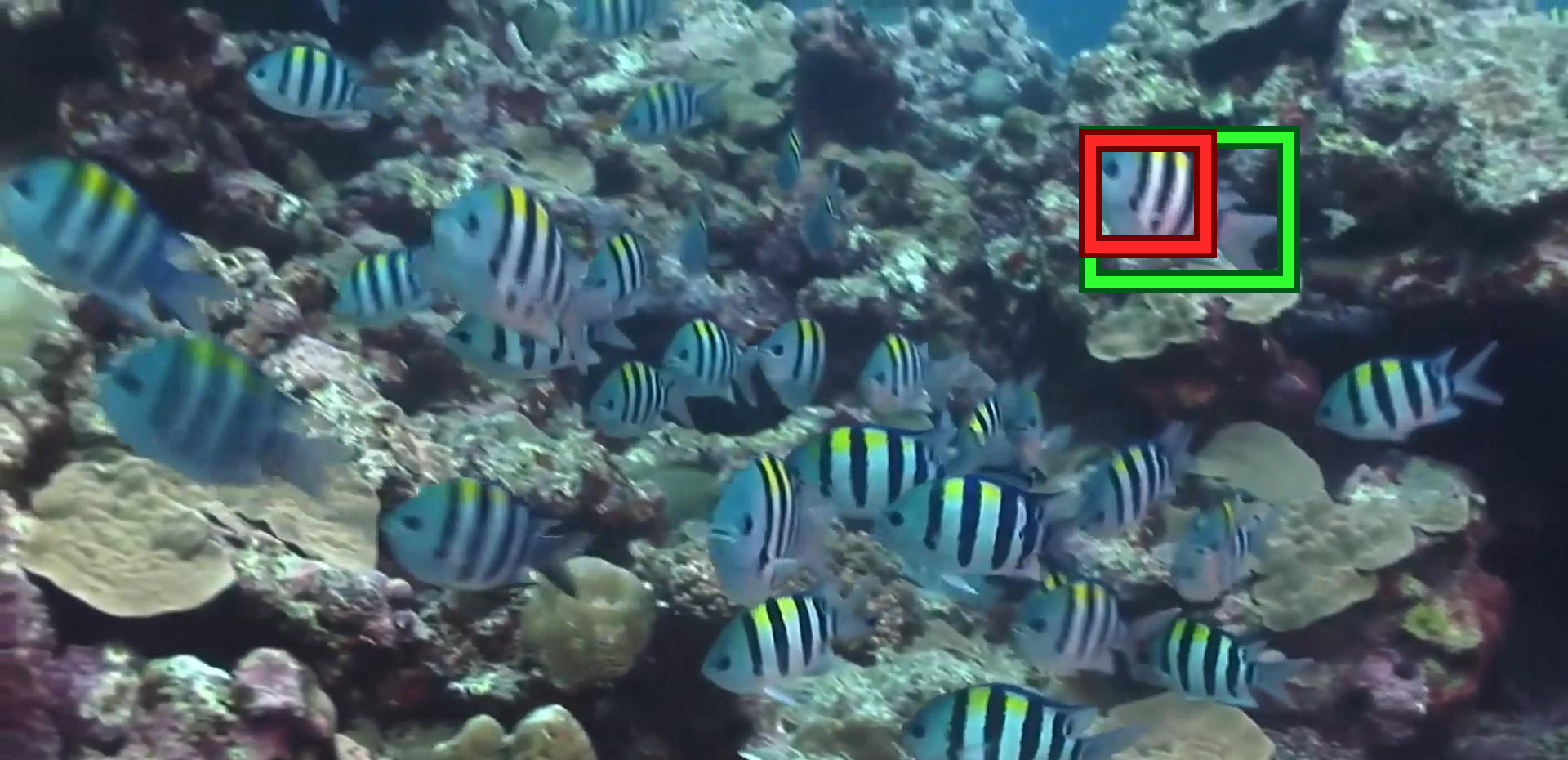}

      \qualRowSix{Ours~VP}%
        {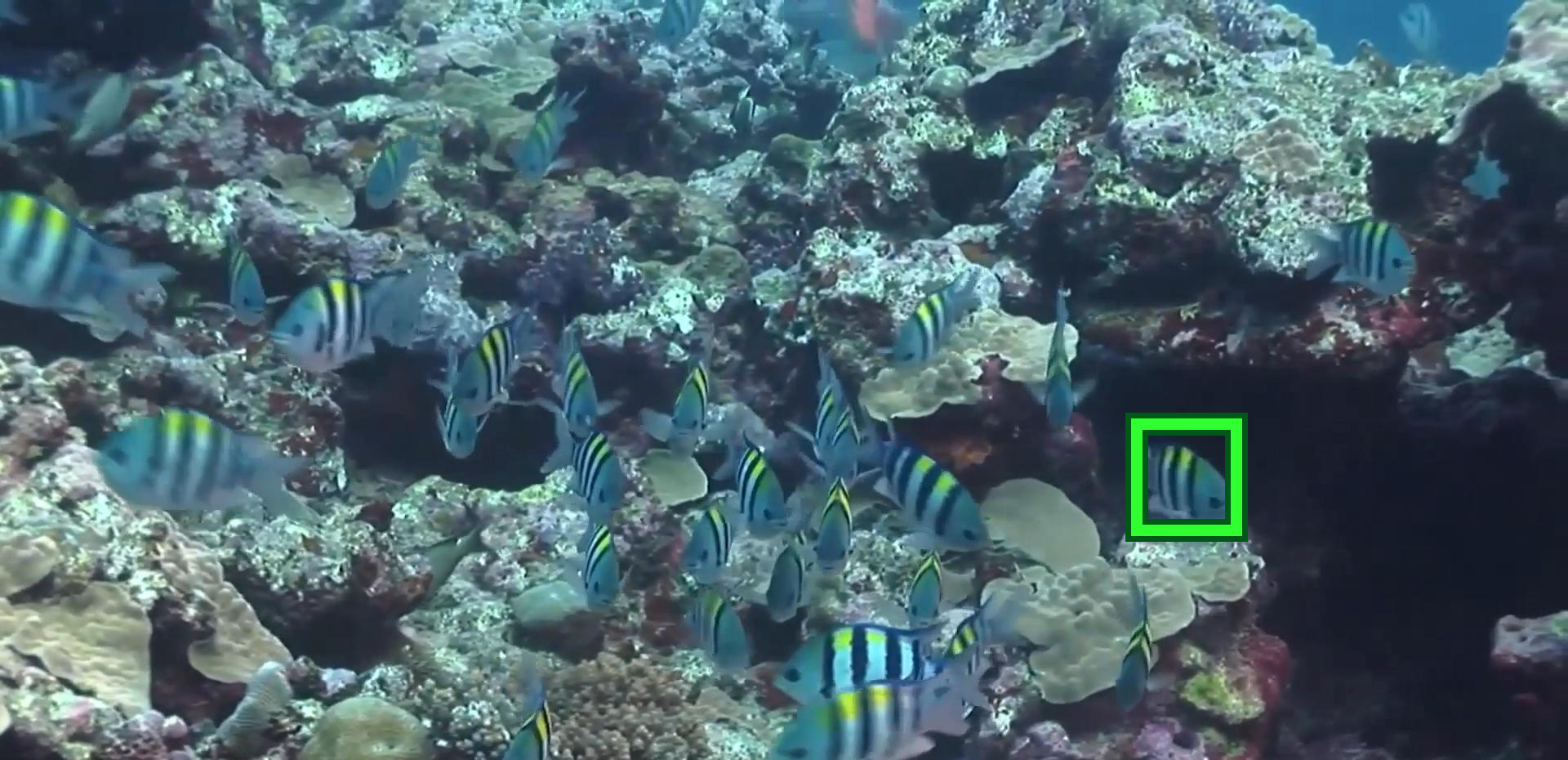}%
        {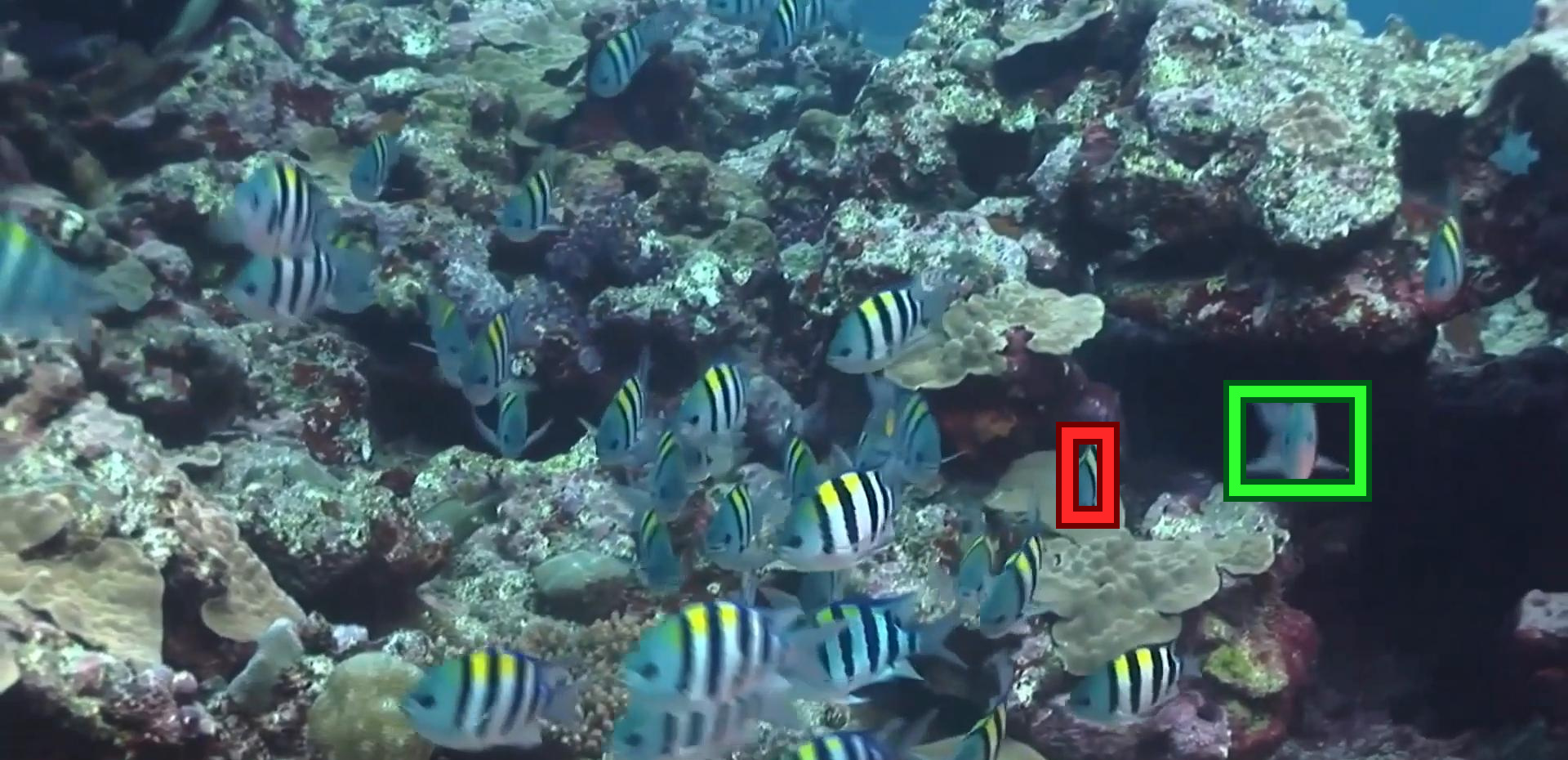}%
        {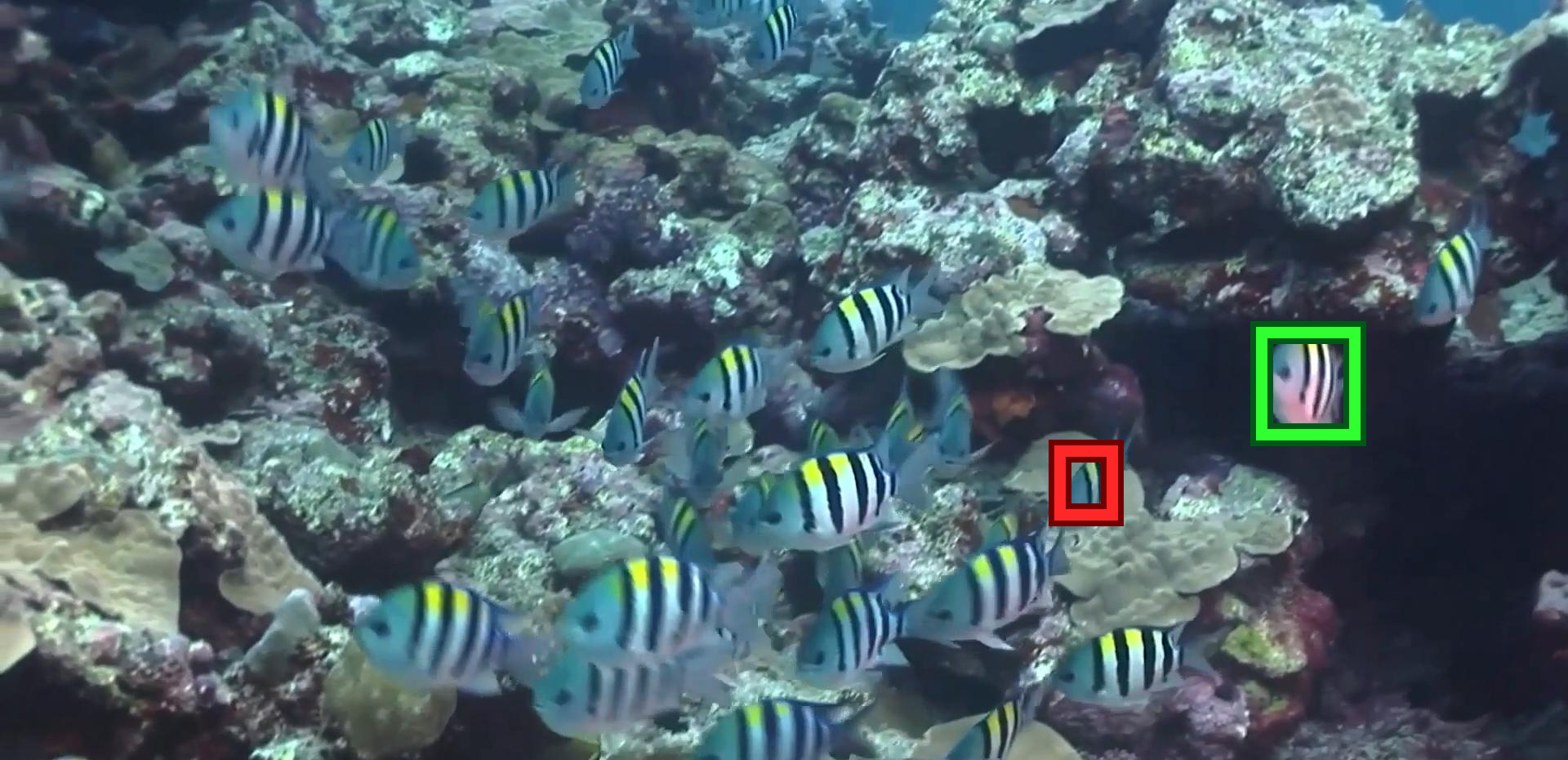}%
        {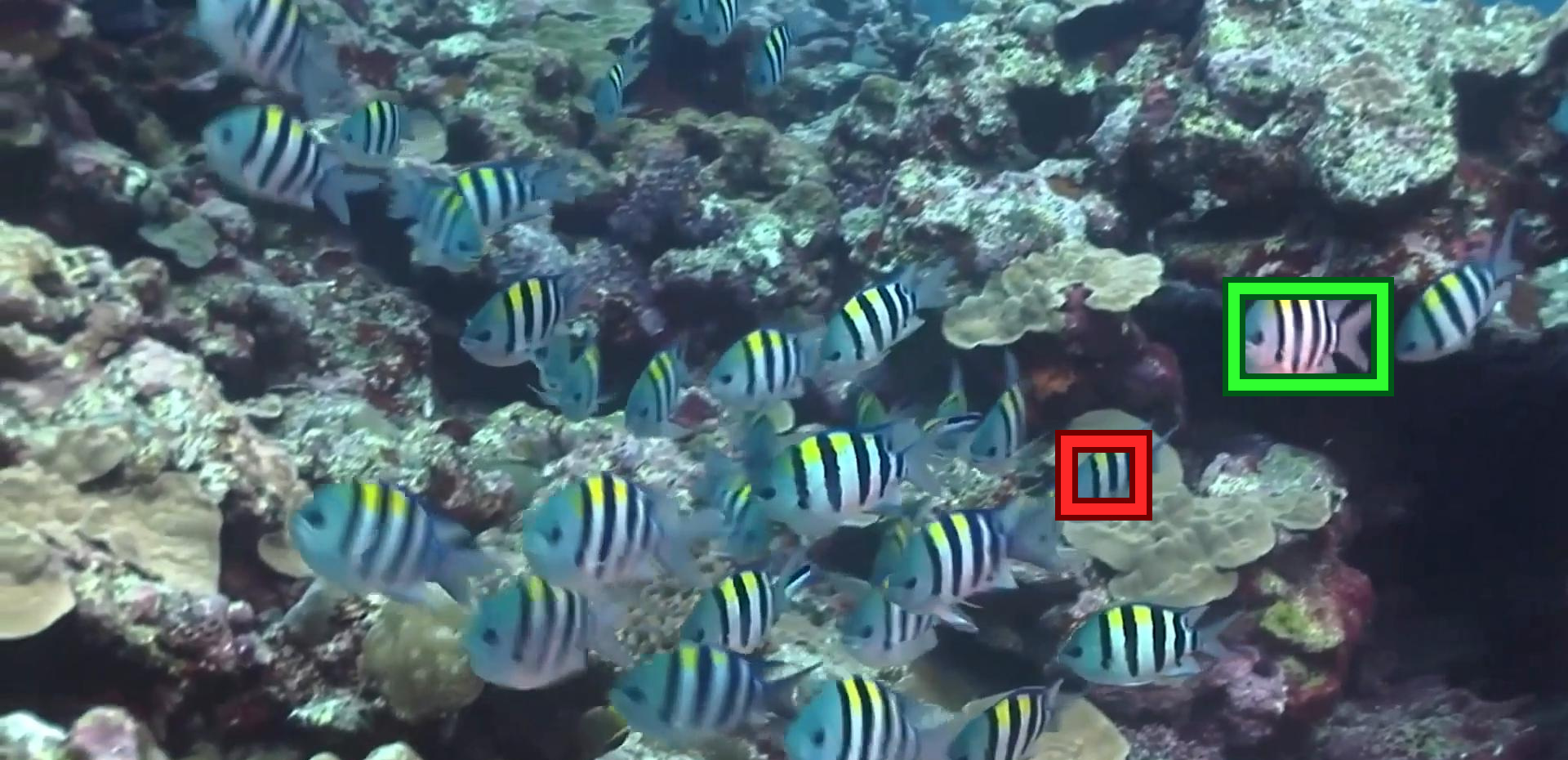}%
        {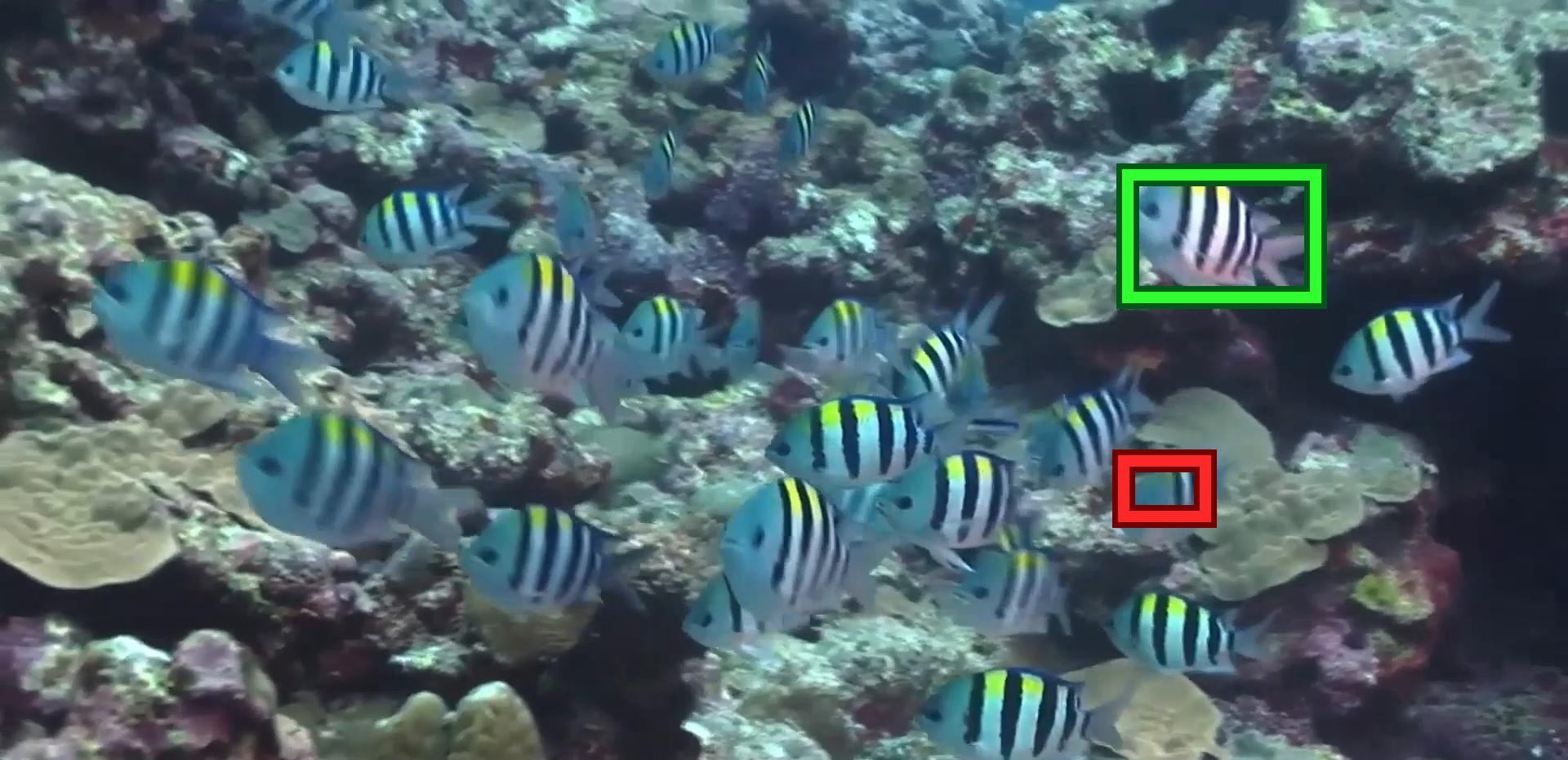}%
        {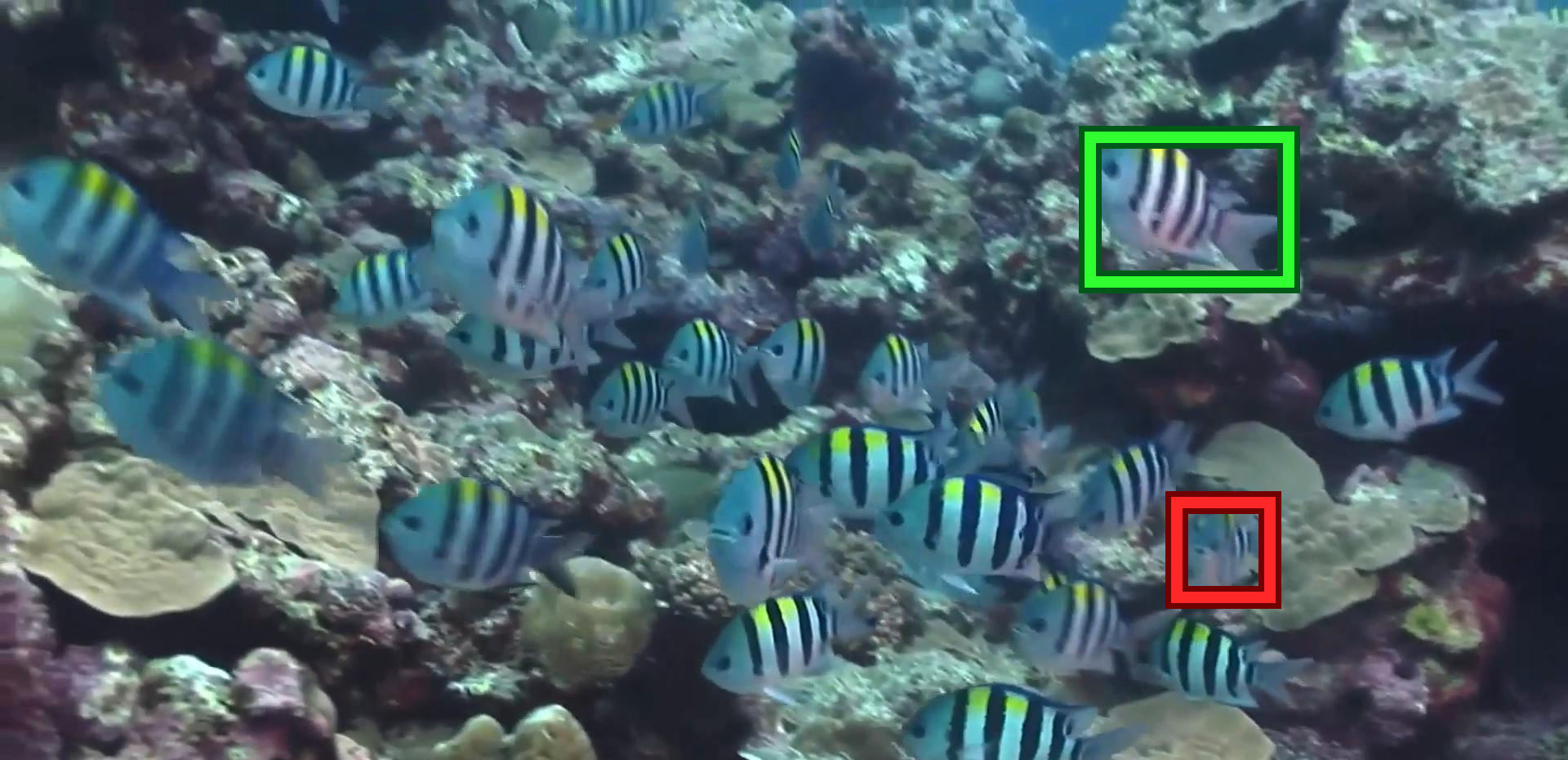}

    \end{minipage}%
  }

  \caption{\textbf{Qualitative comparison of adversarial methods on (a) VOS and (b) VOT tasks.} Columns represent propagation from left to right, where the first column is the prompted frame. Green denotes the clean SAM2.1-L prediction, while red denotes its prediction under the corresponding perturbation and setting. Across both tasks, DistractAttack causes more severe and persistent failures than baselines.}
  \label{fig:qualitative-vos-vot-six}
\end{figure*}

\noindent\textbf{Comparison against other methods.}
We compare image-segmentation, tracking, and VOS attacks on SAM2.1-Large in
\cref{tab:attack_results}. ARA-Attack and VOSPGD are reported only under OP
because they perturb the annotated initialization
frame~\cite{li2023adversarial,jiang2023exploring}. DistractAttack is strongest
overall in both settings. Under VP, it raises DRE from UAP-SAM2's 2.47\% to
10.83\%. Under OP, $N$ denotes the rollout used to construct the target and
distractor banks. At $N=1$, CDS uses only initialization-frame evidence and
lowers $Q$ from 0.172 to 0.131 against VOSPGD. Increasing $N$ to 20 adds
unannotated frames and lowers the VOS average by a further 2.44 points,
showing the benefit of temporal context. OP propagates severe initialization
errors through memory, yielding more empty masks and higher NRE, while VP
sustains misdirection and higher DRE. These complementary failure modes are
also visible in \cref{fig:qualitative-vos-vot-six}.

\noindent\textbf{DistractAttack across tracker variants.}
We evaluate DistractAttack on SAMURAI~\cite{samurai},
DAM4SAM~\cite{dam4sam}, HiM2SAM~\cite{him2sam}, SAMITE~\cite{samite}, and
SAM2Long~\cite{sam2long}, all based on SAM2.1-L. Under VP, SAM2Long and
DAM4SAM are slightly more resistant, while SAMURAI is most affected. OP with
$N=20$ is stronger across every variant. DAM4SAM remains the most resistant
but still degrades sharply. Specialized distractor and memory mechanisms
therefore offer only partial protection against tracking and segmentation
failures. Across SAM2.1 sizes, $Q$ falls to 0.387--0.420 with only modest
variation across scales (\cref{fig:transferability}).

\noindent\textbf{Black-box transferability.}
Since DistractAttack uses white-box optimization, we assess its applicability in the black-box setting. Following standard transfer-based
black-box evaluation~\cite{jiang2023exploring, uad, fan},
each perturbation is optimized on a source model and applied unchanged to a
target. Fig.~\ref{fig:transferability} evaluates four complementary axes.
Cross-variant transfer among SAM2.1, DAM4SAM, SAMURAI, SAMITE, and SAM2Long
yields off-diagonal $Q$ ranges of 0.076--0.192 under OP and 0.447--0.642 under
VP. All variants use SAM2.1-L, with $N=20$ for OP. Transfer across the T, S,
B+, and L sizes also remains effective, showing generalization across scales.
For long-horizon transfer, perturbations optimized over frames 0--100
on ten randomly selected samples are evaluated unchanged at frames 400, 700,
and 1000. On SAM2.1-L, $Q$ increases by 28.7\% for DistractAttack versus
46.1\% for UAP-SAM2, supporting the long-term effect of our trajectory-focused optimization over the consecutive-frame optimization of UAP-SAM2. Complementary VOS transfer is provided in
Appendix~\ref{supp:evaluation:transfer}.

\suppressfloats[t]
\begin{table}[t]
    \centering
    \caption{Ablation of DistractAttack on SAM2.1-L under video (VP) and
    one-shot (OP) perturbations. }
    \label{tab:ablation}
    \scriptsize
    \setlength{\tabcolsep}{1.8pt}
    \renewcommand{\arraystretch}{0.88}

    \textbf{(a) Attack objectives and optimization.}
    \vspace{2pt}

    \resizebox{\columnwidth}{!}{%
    \begin{tabular}{@{}lccc ccc|ccc@{}}
        \toprule
        & & & &
        \multicolumn{3}{c|}{VP}
        & \multicolumn{3}{c}{OP} \\
        \cmidrule(lr){5-7}
        \cmidrule(lr){8-10}
        Setting
        & $\mathcal{L}_{\rm mask}$
        & $\mathcal{L}_{\rm clut}$
        & $\mathcal{L}_{\rm disp}$
        & $Q\downarrow$
        & $\mathrm{DRE}\uparrow$
        & VOS Avg.$\downarrow$
        & $Q\downarrow$
        & $\mathrm{DRE}\uparrow$
        & VOS Avg.$\downarrow$ \\
        \midrule

        \multicolumn{10}{l}{\emph{Attack objectives}} \\

                $\mathcal{L}_{\rm mask}$ only
        & $\checkmark$ & &
        & 0.631 & 2.78 & 68.42
        & 0.497 & 0.78 & 47.83 \\

        w/o $\mathcal{L}_{\rm mask}$
        & & $\checkmark$ & $\checkmark$
        & 0.561 & 9.89 & 62.76
        & 0.244 & 11.96 & 34.12 \\

        w/o $\mathcal{L}_{\rm disp}$
        & $\checkmark$ & $\checkmark$ &
        & 0.502 & 6.99 & 58.61
        & 0.291 & 3.89 & 37.64 \\

        w/o $\mathcal{L}_{\rm clut}$
        & $\checkmark$ & & $\checkmark$
        & 0.491 & 7.78 & 56.94
        & 0.202 & 4.05 & 30.87 \\

        Full
& $\checkmark$ & $\checkmark$ & $\checkmark$
& 0.420 & 10.83 & 50.97
& 0.076 & 9.28 & 19.56 \\

        \midrule
        \multicolumn{10}{l}{\emph{Iteration length}} \\

        5 passes
        & $\checkmark$ & $\checkmark$ & $\checkmark$
        & 0.501 & 5.77 & 58.87
        & 0.119 & 8.78 & 24.83 \\

        10 passes
        & $\checkmark$ & $\checkmark$ & $\checkmark$
        & 0.420 & 10.83 & 50.97
        & 0.076 & 9.28 & 19.56 \\

        15 passes
        & $\checkmark$ & $\checkmark$ & $\checkmark$
        & 0.343 & 13.67 & 49.84
        & 0.071 & 7.45 & 20.41 \\

        \midrule
        \multicolumn{10}{l}{\emph{Perturbation magnitude}} \\

        $\epsilon=4/255$
        & $\checkmark$ & $\checkmark$ & $\checkmark$
        & 0.562 & 10.44 & 62.03
        & 0.076 & 9.28 & 19.56 \\

        $\epsilon=8/255$
        & $\checkmark$ & $\checkmark$ & $\checkmark$
        & 0.420 & 10.83 & 50.97
        & 0.069 & 8.49 & 18.74 \\

        $\epsilon=16/255$
        & $\checkmark$ & $\checkmark$ & $\checkmark$
        & 0.390 & 9.51 & 49.91
        & 0.046 & 9.95 & 17.82 \\

        \bottomrule
    \end{tabular}%
    }

    \vspace{0.35em}

    \textbf{(b) CDS evidence and temporal design.}
    \vspace{2pt}

    \resizebox{\columnwidth}{!}{%
    \begin{tabular}{@{}lccc ccc|ccc@{}}
        \toprule
        & & & &
        \multicolumn{3}{c|}{VP}
        & \multicolumn{3}{c}{OP} \\
        \cmidrule(lr){5-7}
        \cmidrule(lr){8-10}
        Setting
        & $\mathcal{P}_t^{\mathrm{sem}}$
        & $\mathcal{P}_t^{\mathrm{alt}}$
        & $\mathcal{P}_t^{\mathrm{att}}$
        & $Q\downarrow$
        & $\mathrm{DRE}\uparrow$
        & VOS Avg.$\downarrow$
        & $Q\downarrow$
        & $\mathrm{DRE}\uparrow$
        & VOS Avg.$\downarrow$ \\
        \midrule

        \multicolumn{10}{l}{\emph{Distractor evidence}} \\

        $\mathcal{P}_t^{\mathrm{sem}}$ only
        & $\checkmark$ & &
        & 0.651 & 7.42 & 69.47
        & 0.312 & 7.34 & 39.52 \\

        w/o $\mathcal{P}_t^{\mathrm{sem}}$
        & & $\checkmark$ & $\checkmark$
        & 0.553 & 6.55 & 63.38
        & 0.215 & 4.56 & 31.74 \\

        w/o $\mathcal{P}_t^{\mathrm{alt}}$
        & $\checkmark$ & & $\checkmark$
        & 0.591 & 8.91 & 65.96
        & 0.290 & 7.94 & 37.21 \\

        w/o $\mathcal{P}_t^{\mathrm{att}}$
        & $\checkmark$ & $\checkmark$ &
        & 0.501 & 9.34 & 58.79
        & 0.255 & 8.12 & 34.03 \\

        \midrule
        \multicolumn{10}{l}{\emph{History and routing}} \\

        FIFO $\mathcal{E}$
        & $\checkmark$ & $\checkmark$ & $\checkmark$
        & 0.611 & 4.55 & 67.14
        & 0.211 & 5.44 & 31.02 \\

        Uniform routing
        & $\checkmark$ & $\checkmark$ & $\checkmark$
        & 0.591 & 6.55 & 64.81
        & 0.155 & 7.43 & 26.41 \\

        First-frame banks
        & $\checkmark$ & $\checkmark$ & $\checkmark$
        & 0.742 & 1.44 & 69.91
        & 0.131 & 8.71 & 22.00 \\

        Current-frame banks
        & $\checkmark$ & $\checkmark$ & $\checkmark$
        & 0.581 & 7.40 & 58.34
        & -- & -- & -- \\

        Full CDS
& $\checkmark$ & $\checkmark$ & $\checkmark$
& 0.420 & 10.83 & 50.97
& 0.076 & 9.28 & 19.56 \\

        \bottomrule
    \end{tabular}%
    }
\end{table}

\section{Ablation Studies}
\label{sec:ablations}
Unless varied, both settings use 10 optimization passes, with
$\epsilon=8/255$ for VP and $\epsilon=4/255$ for OP. The default OP rollout
uses $N=20$ frames.

\noindent\textbf{Across the attack objectives,} mask suppression weakens both
tasks but induces little drift. Displacement ($\mathcal{L}_{\rm disp}$)
redirects this failure toward spatial misdirection, while clutter corruption
($\mathcal{L}_{\rm clut}$) strengthens the attack across both tasks. Their
combination reduces $Q$ to 0.420 under VP and 0.076 under OP. Removing mask
suppression raises OP DRE but is substantially weaker in $Q$ and VOS Avg.
\textbf{For iteration length,} ten passes substantially outperform five in
both settings. Fifteen further improves VP but gains only 0.005 in OP $Q$ while
worsening DRE and VOS performance. We therefore retain ten passes.
\textbf{With increasing perturbation magnitude,} a larger budget strengthens
the attack but permits more visible perturbations. Both $Q$ and VOS Avg.
decrease with $\epsilon$, while VP DRE peaks at the selected $8/255$.
\textbf{Across sources of distractor evidence,} semantic proposals provide
broad coverage, while alternative masks and adversarial attention maps add
decoder-specific cues. Removing any source weakens both tasks, and every
reduced variant remains behind full CDS. \textbf{For historical banks and the
update policy,} restricting $\mathcal{E}$ to the first or current frame weakens
the attack. The current-frame variant applies only to VP because OP optimizes
the initialization frame alone. FIFO replaces entries by arrival order,
whereas uniform updates ignore consistency with existing distractors, so
neither effectively preserves or prioritizes recurring evidence. Full CDS
performs best by maintaining a persistent $\mathcal{E}$ and prioritizing
proposals supported by its most effective entries. Supporting ablations are given
in Appendix~\ref{supp:evaluation:ablations}.

\noindent\textbf{Defense Case Study.}
We evaluate complementary feature- and memory-level defenses against
DistractAttack. Robust-SAM~\cite{long2025robust} adversarially adapts the
singular values of selected convolutional weights. We further
repurpose SAM2Long's attention modulation~\cite{sam2long} to attenuate
low-confidence memory keys and limit uncertain time steps and initialization
errors. Robust-SAM and our memory defense raise $Q$ from 0.420 to 0.515 and
0.476, respectively. Combined, they reach 0.532 $Q$ and reduce DRE from
10.83\% to 8.45\%, but remain well below the clean $Q$ of 0.836. Existing
feature and memory controls therefore offer only partial protection. Details
are provided in Appendix~\ref{sec:supp-defense}.

\section{Conclusion}
\label{sec:Conclusion}

We examined how adversarial perturbations compromise target identity in SAM2.
Although recurrent memory promotes temporal consistency, it can propagate
perturbation-induced local ambiguity into persistent tracking failure. We
introduced DistractAttack, a distractor-centric attack that exploits plausible
competing regions rather than relying solely on foreground suppression. Its
Counterfactual Distractor Search jointly follows clean and attacked
trajectories, identifies model-supported distractors, and accumulates their
evidence over time to sustain failure. Across VOT and VOS, DistractAttack
remains effective under both perturbation settings and long temporal horizons,
while generalizing and transferring across model scales and tracker variants.
These results identify distractor-induced failure and initialization
sensitivity as key vulnerabilities of SAM2-based trackers, motivating
stronger robustness mechanisms in video foundation models.

\section*{Broader Societal Impact}
This work provides new insight into the vulnerabilities of SAM2. We hope these findings can inform future efforts to better understand and mitigate such failure modes, contributing to safer and more reliable deployment.

\section*{Acknowledgments}
This research was funded by Khalifa University
of Science and Technology through the Faculty Start-Ups under Project ID: KU-INT-FSU-2005-8474000775. This research was also supported by KU Research Center for Advanced Intelligent Systems (AIS), Khalifa University of Science and Technology (KU-AIS).

{
    \small
    \bibliographystyle{ieeenat_fullname}
    \bibliography{main}
}

\clearpage

\clearpage
\pagenumbering{arabic}
\setcounter{page}{1}

\renewcommand{\thesection}{\arabic{section}}
\setcounter{section}{7}

\maketitlesupplementary

\setlength{\columnsep}{0.25in}

\definecolor{acceptedqualblue}{RGB}{35,76,112}
\newlength{\acceptedQualLabelWidth}
\newlength{\acceptedQualImageWidth}

\newcommand{\acceptedQualImage}[1]{%
  \raisebox{-0.5\height}{\includegraphics[width=\acceptedQualImageWidth]{#1}}%
}

\newcommand{\acceptedQualRowEight}[9]{%
  \noindent\hbox to \linewidth{%
  \makebox[\acceptedQualLabelWidth][c]{%
    \rotatebox[origin=c]{90}{\scriptsize\bfseries #1}}\hspace{1pt}%
  \acceptedQualImage{#2}\hspace{1pt}%
  \acceptedQualImage{#3}\hspace{1pt}%
  \acceptedQualImage{#4}\hspace{1pt}%
  \acceptedQualImage{#5}\hspace{1pt}%
  \acceptedQualImage{#6}\hspace{1pt}%
  \acceptedQualImage{#7}\hspace{1pt}%
  \acceptedQualImage{#8}\hspace{1pt}%
  \acceptedQualImage{#9}\hss}\par\vspace{0.5pt}%
}

\newcommand{\AcceptedWarmupGrid}{%
  \resizebox{0.98\textwidth}{!}{%
  \begin{minipage}{\textwidth}
  \setlength{\acceptedQualLabelWidth}{12pt}%
  \setlength{\acceptedQualImageWidth}{%
    \dimexpr(\linewidth-\acceptedQualLabelWidth-8pt)/8\relax}%
  \acceptedQualRowEight{DAM4SAM}%
    {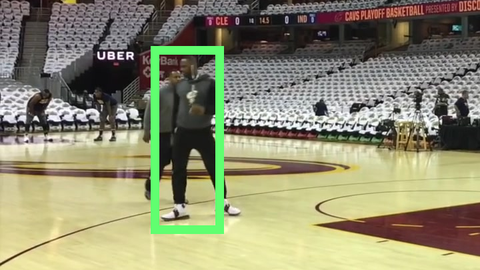}%
    {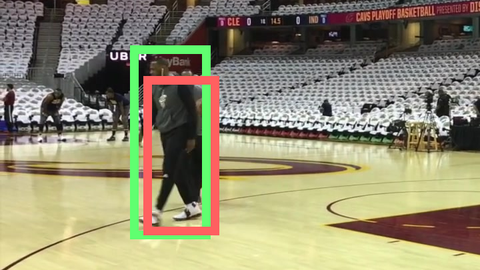}%
    {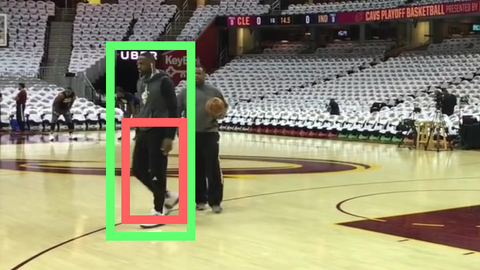}%
    {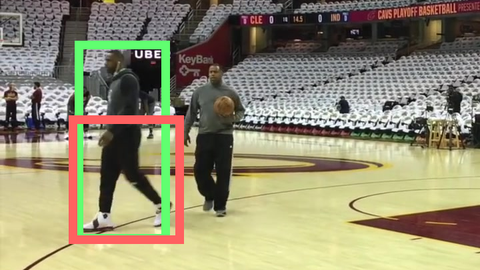}%
    {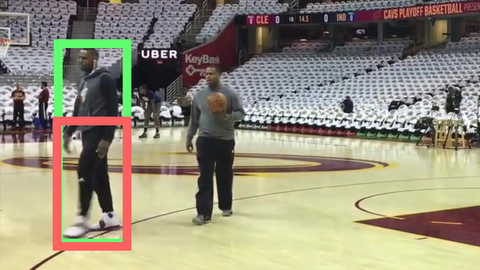}%
    {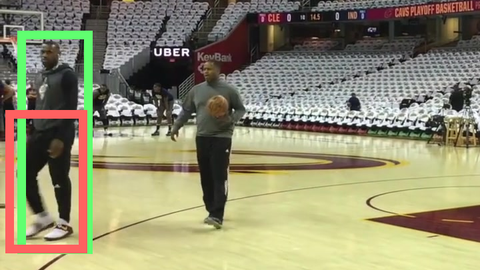}%
    {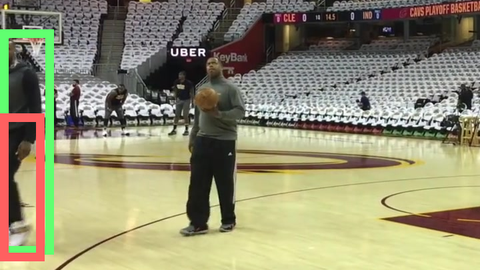}%
    {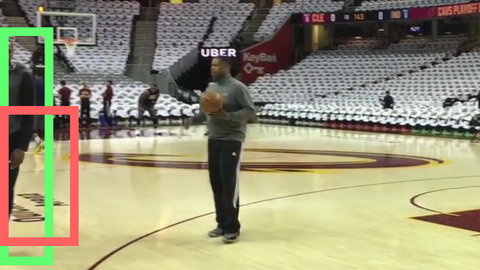}
  \acceptedQualRowEight{SAM2Long}%
    {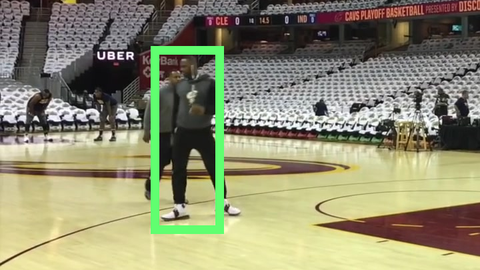}%
    {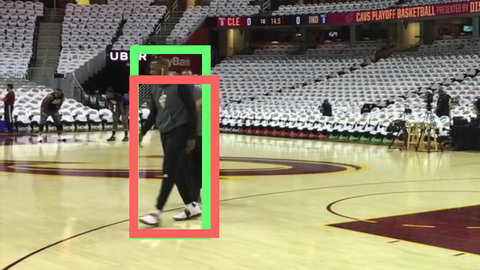}%
    {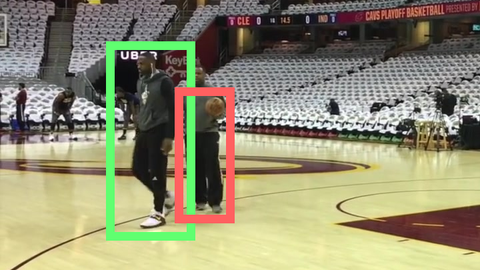}%
    {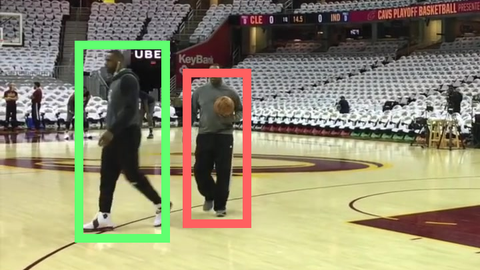}%
    {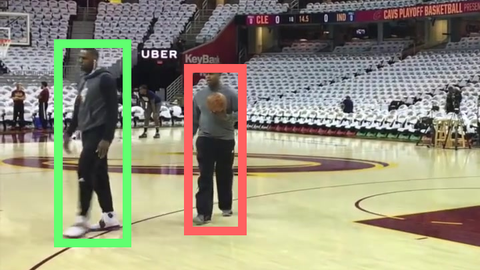}%
    {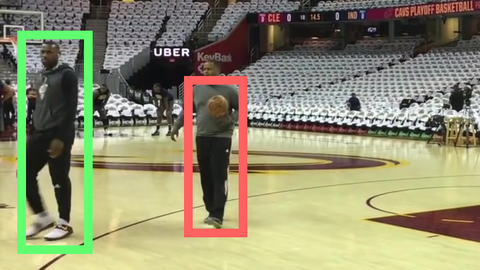}%
    {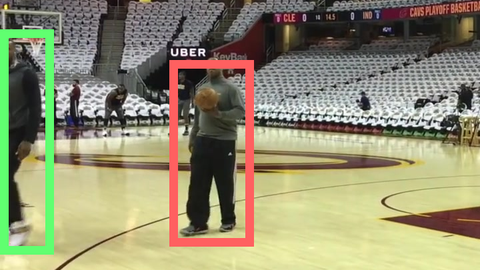}%
    {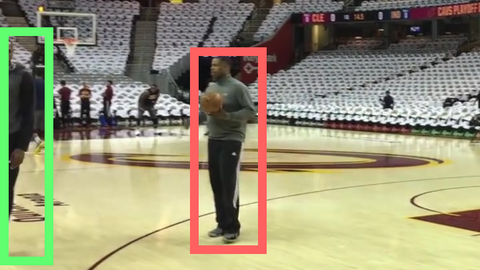}
  \acceptedQualRowEight{HiM2SAM}%
    {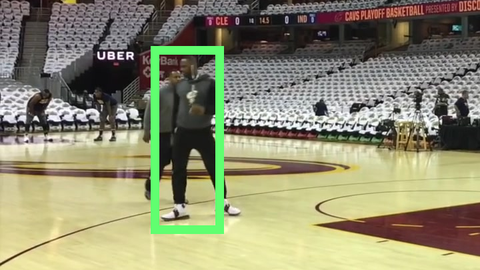}%
    {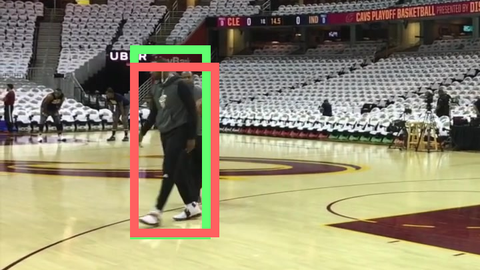}%
    {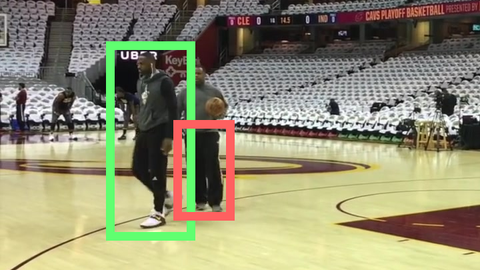}%
    {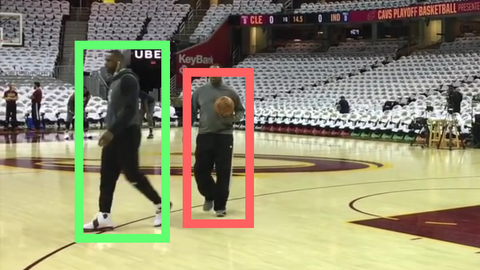}%
    {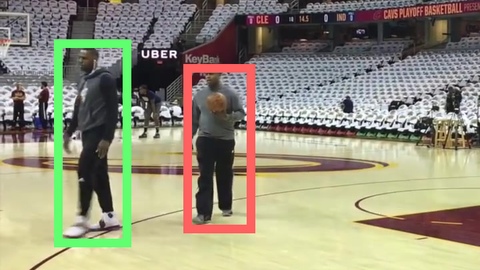}%
    {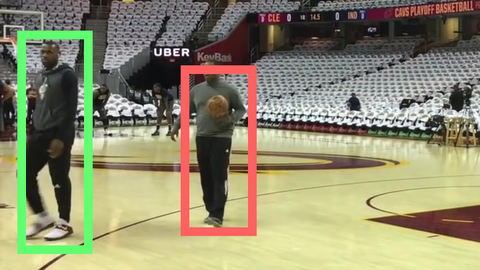}%
    {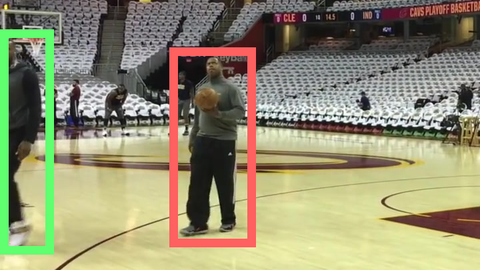}%
    {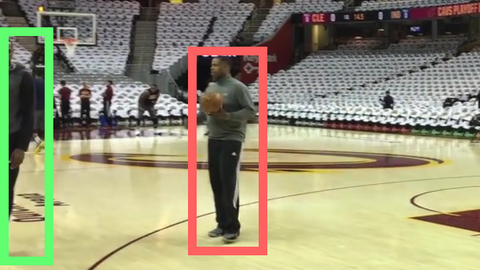}
  \acceptedQualRowEight{SAMITE}%
    {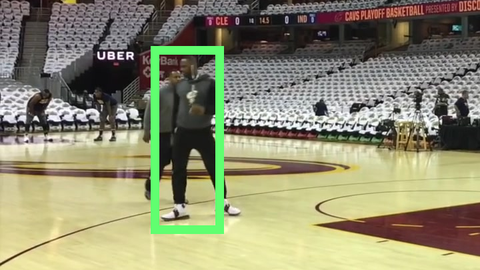}%
    {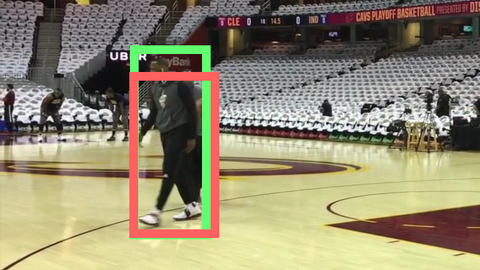}%
    {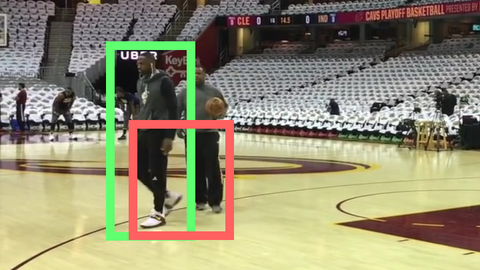}%
    {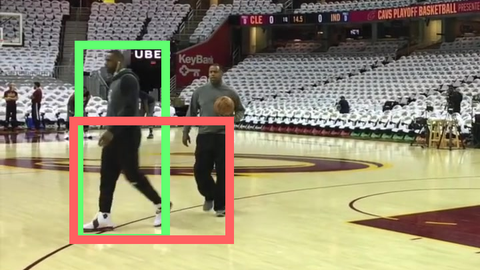}%
    {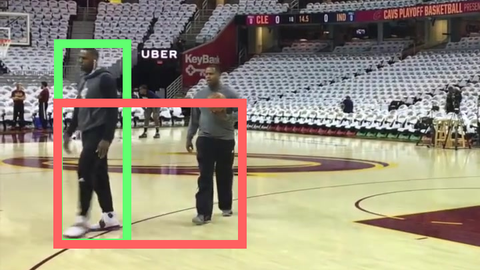}%
    {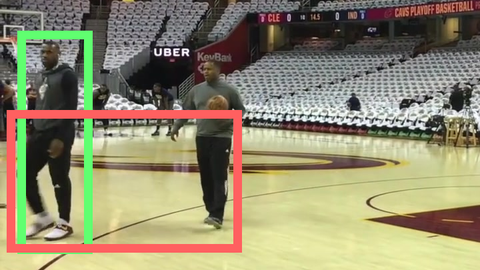}%
    {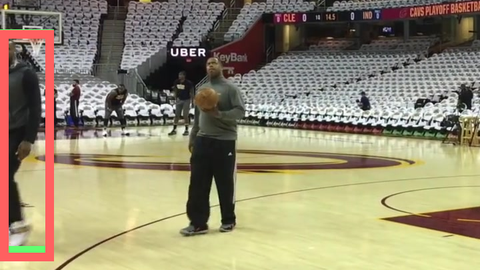}%
    {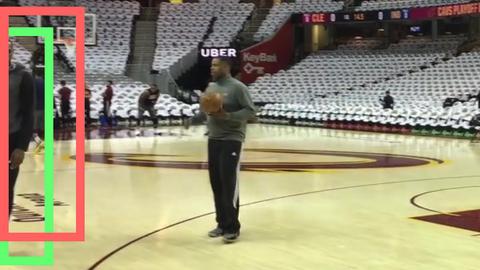}
  \acceptedQualRowEight{SAMURAI}%
    {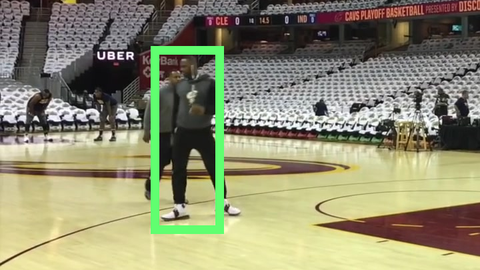}%
    {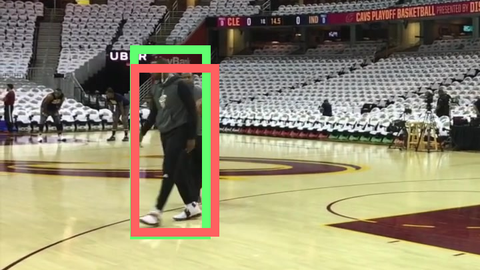}%
    {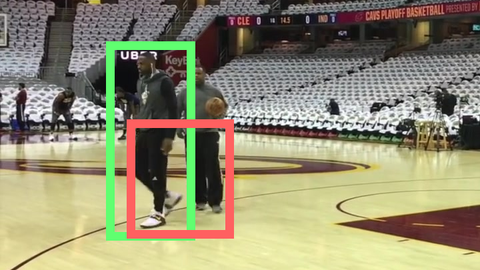}%
    {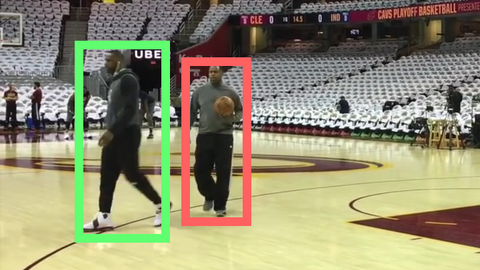}%
    {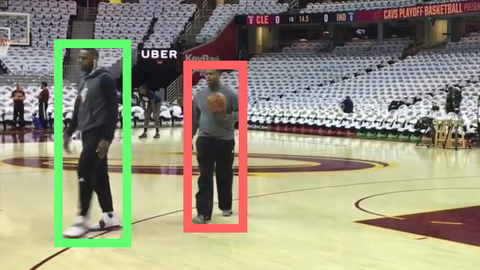}%
    {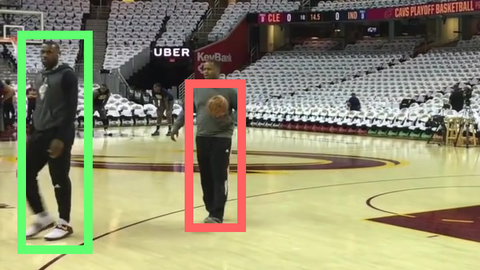}%
    {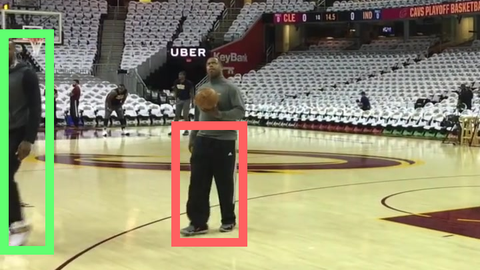}%
    {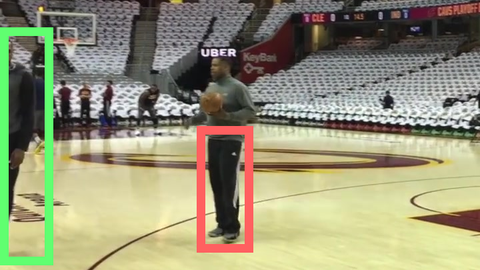}
  \end{minipage}}%
}

\newcommand{\AcceptedSharkGrid}{%
  \resizebox{0.98\textwidth}{!}{%
  \begin{minipage}{\textwidth}
  \setlength{\acceptedQualLabelWidth}{12pt}%
  \setlength{\acceptedQualImageWidth}{%
    \dimexpr(\linewidth-\acceptedQualLabelWidth-8pt)/8\relax}%
  \acceptedQualRowEight{FAN}%
    {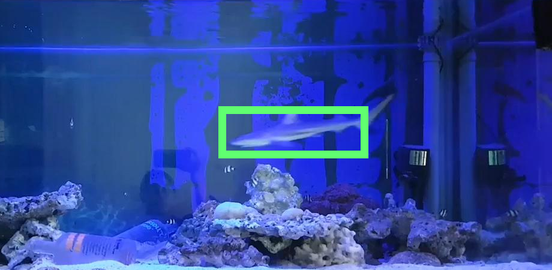}%
    {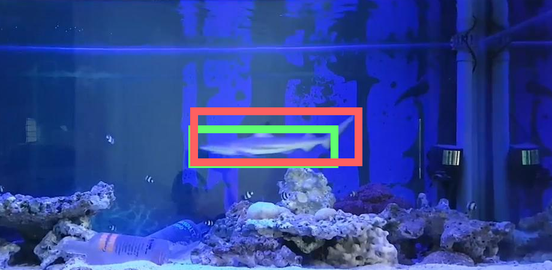}%
    {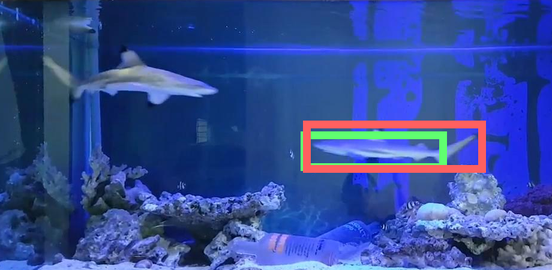}%
    {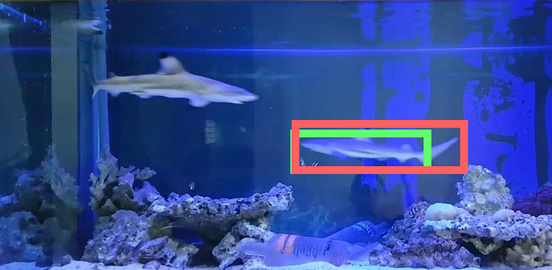}%
    {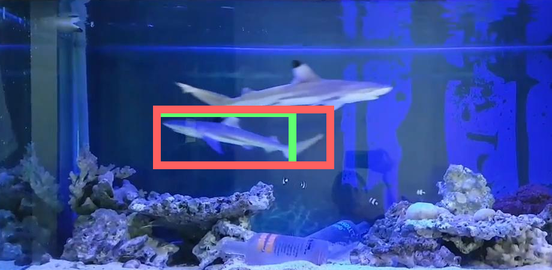}%
    {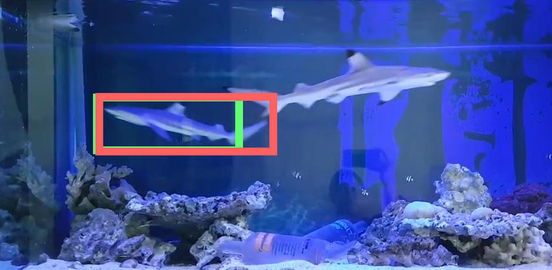}%
    {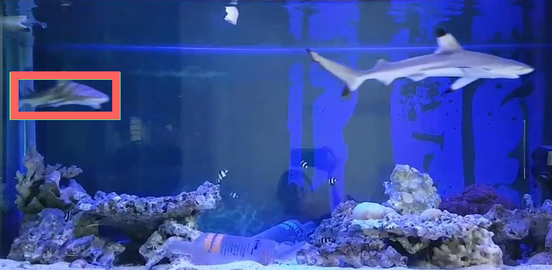}%
    {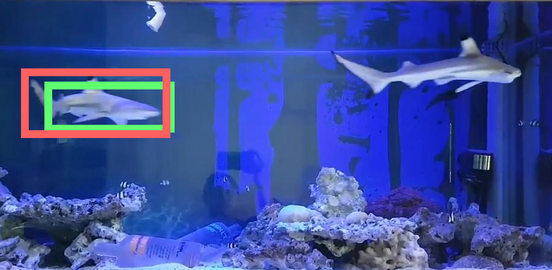}
  \acceptedQualRowEight{\tiny DarkSAM}%
    {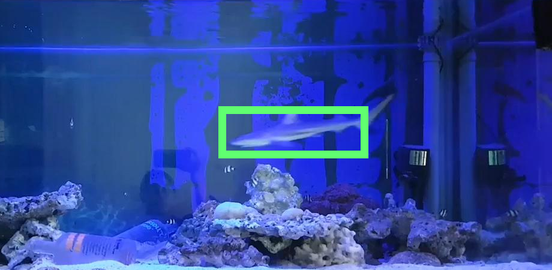}%
    {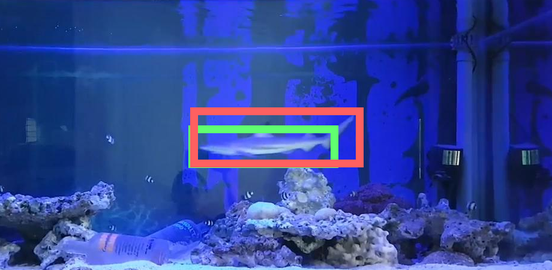}%
    {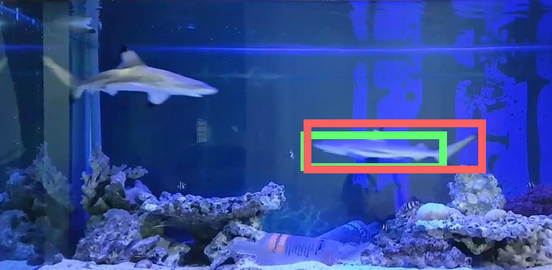}%
    {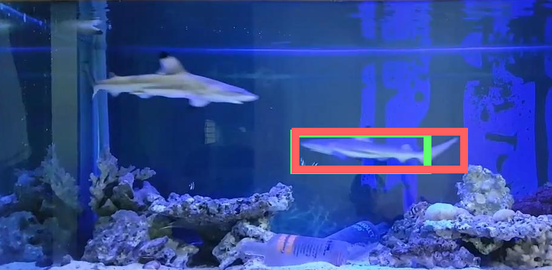}%
    {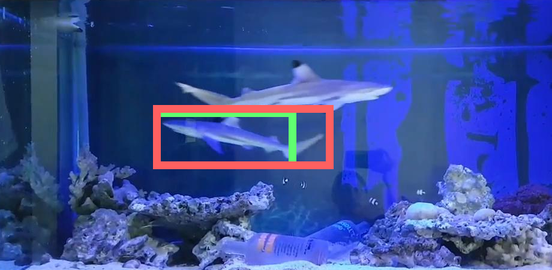}%
    {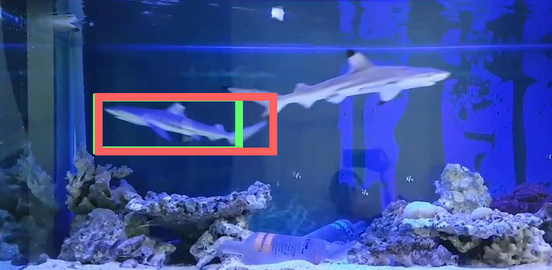}%
    {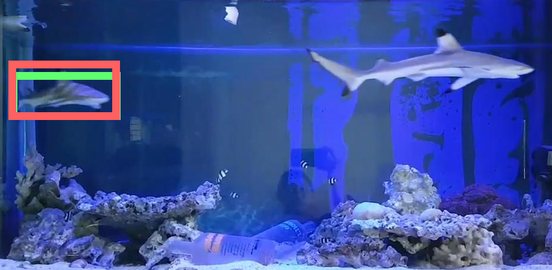}%
    {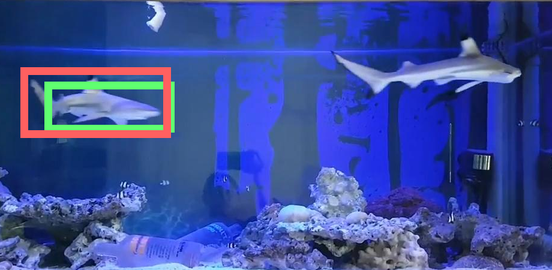}
  \acceptedQualRowEight{SegPGD}%
    {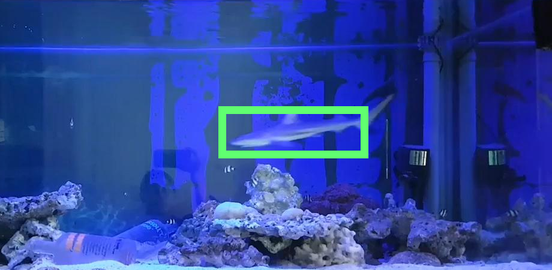}%
    {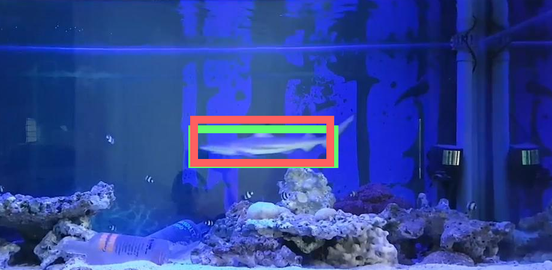}%
    {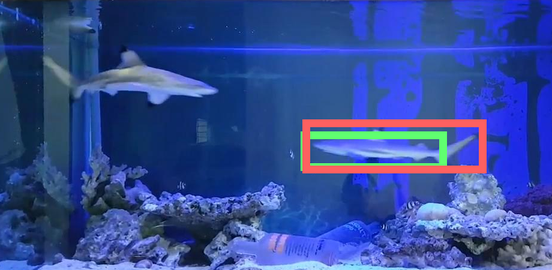}%
    {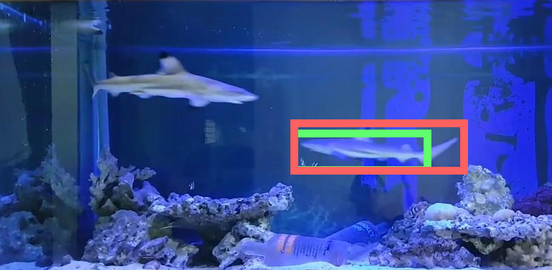}%
    {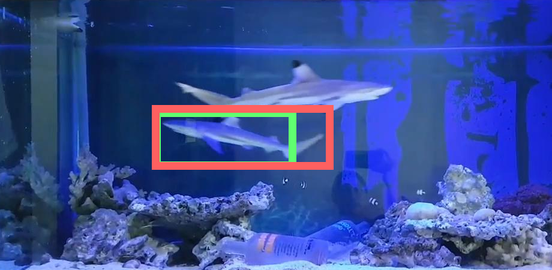}%
    {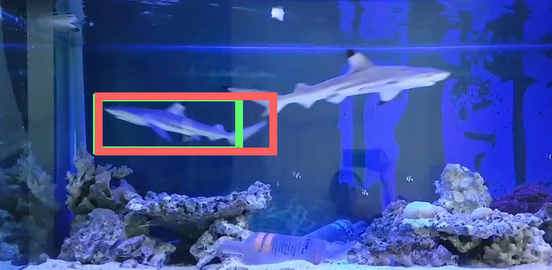}%
    {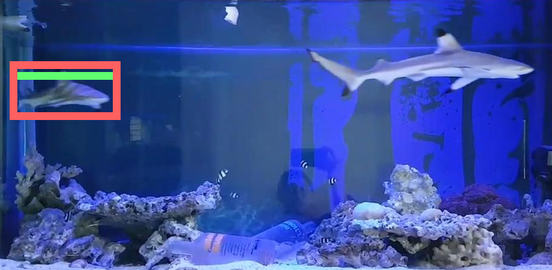}%
    {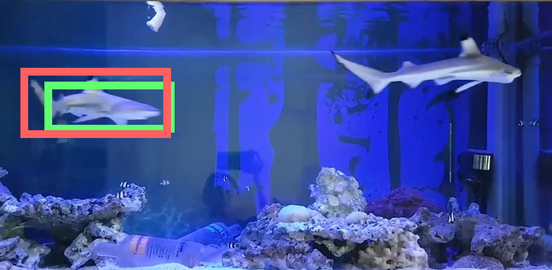}
  \acceptedQualRowEight{Ours}%
    {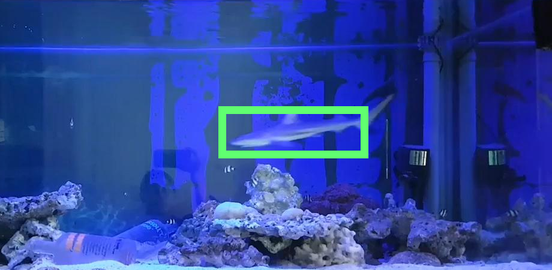}%
    {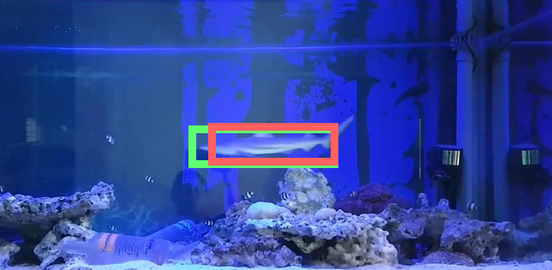}%
    {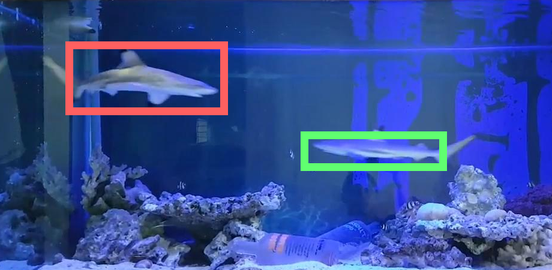}%
    {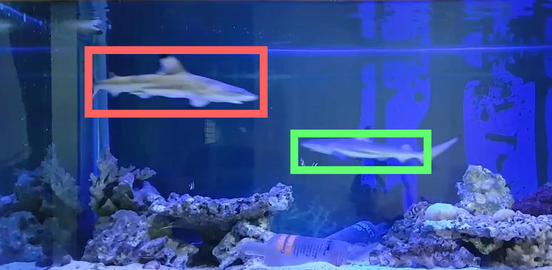}%
    {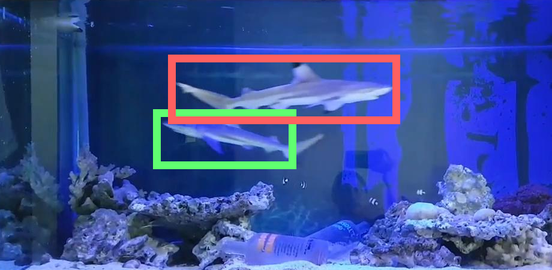}%
    {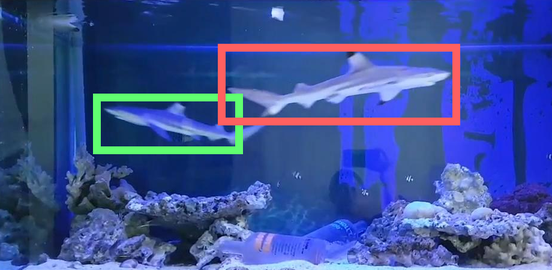}%
    {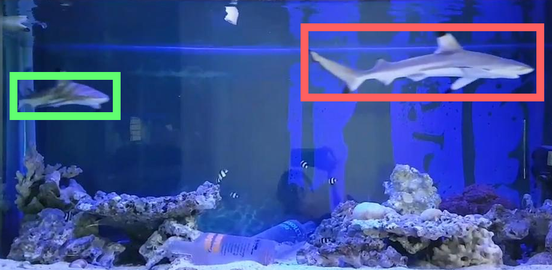}%
    {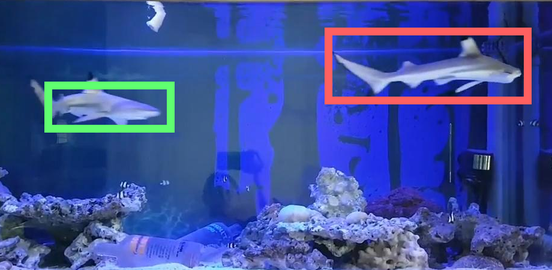}
  \end{minipage}}%
}

\newcommand{\AcceptedWingsuitGrid}{%
  \resizebox{0.98\textwidth}{!}{%
  \begin{minipage}{\textwidth}
  \setlength{\acceptedQualLabelWidth}{12pt}%
  \setlength{\acceptedQualImageWidth}{%
    \dimexpr(\linewidth-\acceptedQualLabelWidth-8pt)/8\relax}%
  \acceptedQualRowEight{Tiny}%
    {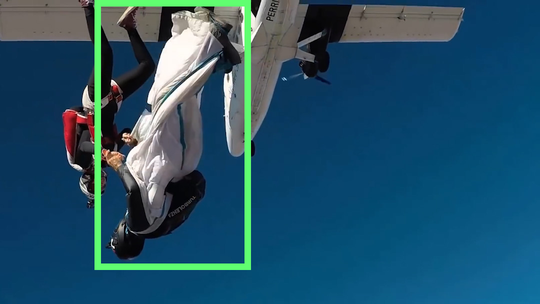}%
    {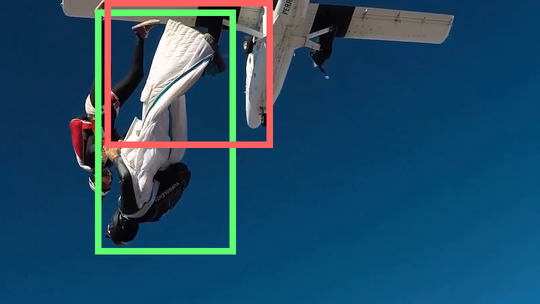}%
    {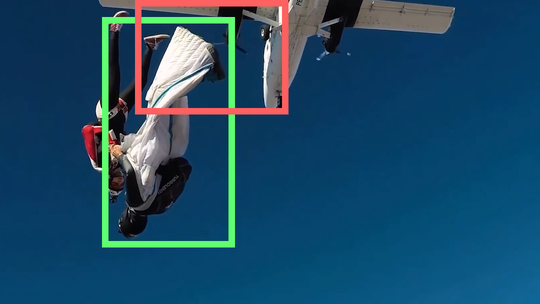}%
    {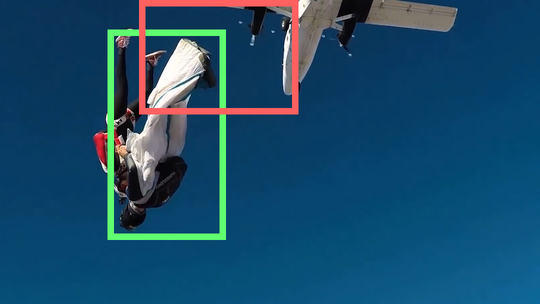}%
    {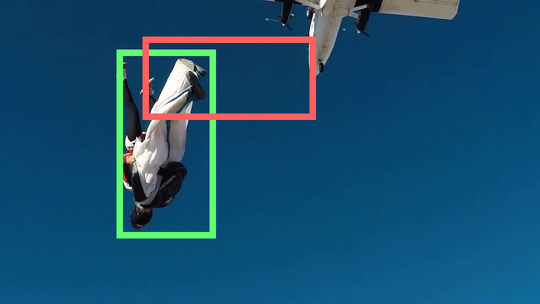}%
    {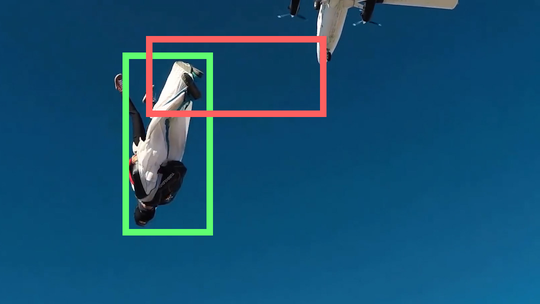}%
    {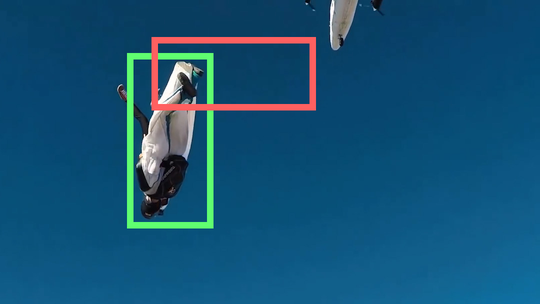}%
    {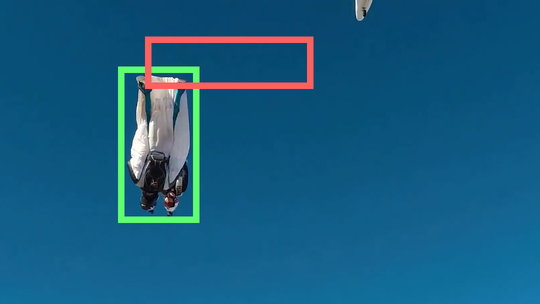}
  \acceptedQualRowEight{Small}%
    {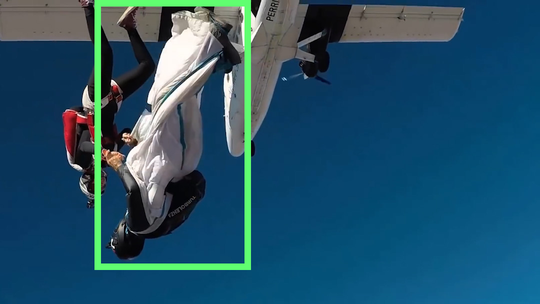}%
    {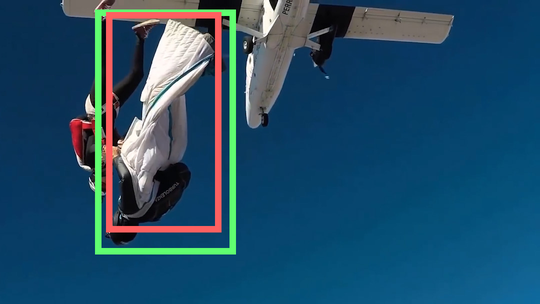}%
    {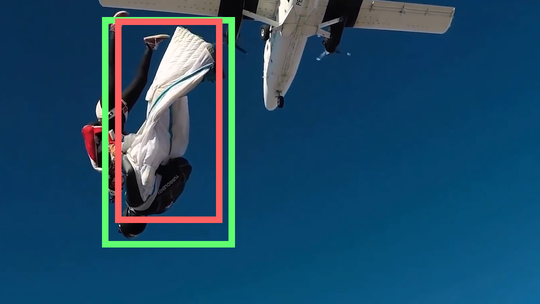}%
    {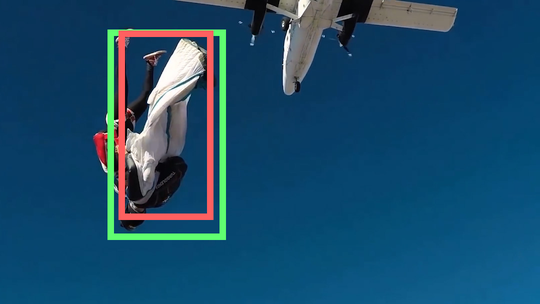}%
    {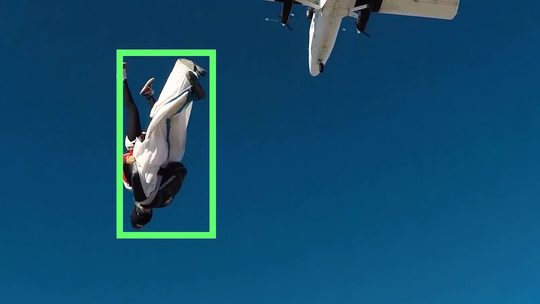}%
    {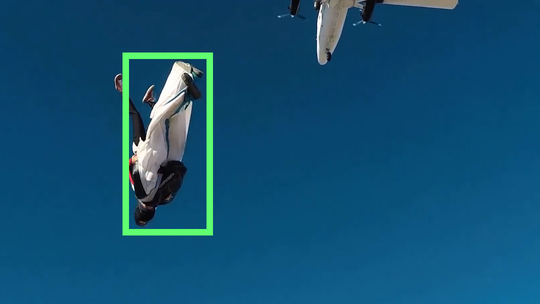}%
    {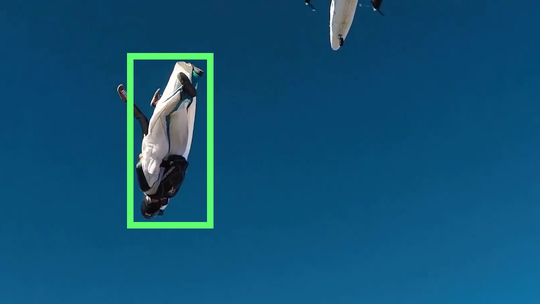}%
    {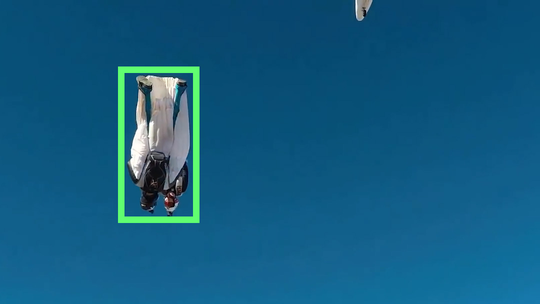}
  \acceptedQualRowEight{Base+}%
    {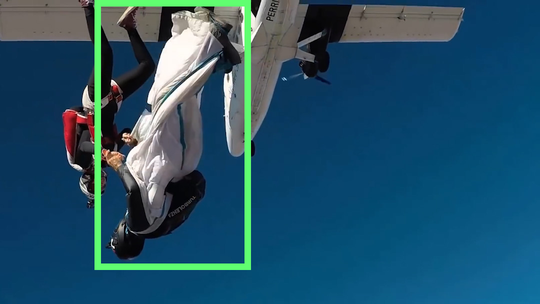}%
    {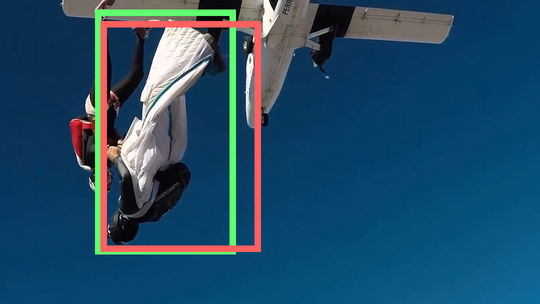}%
    {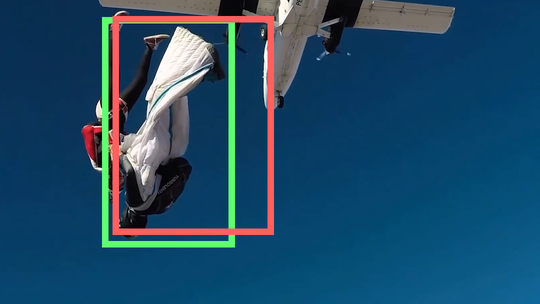}%
    {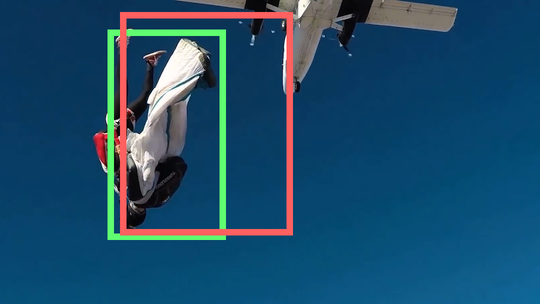}%
    {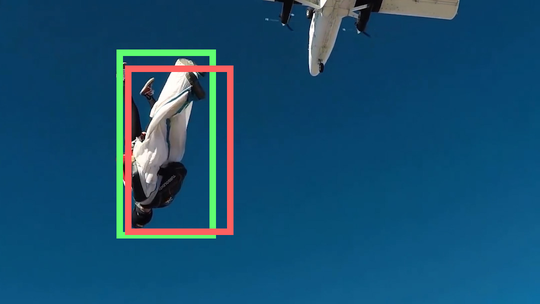}%
    {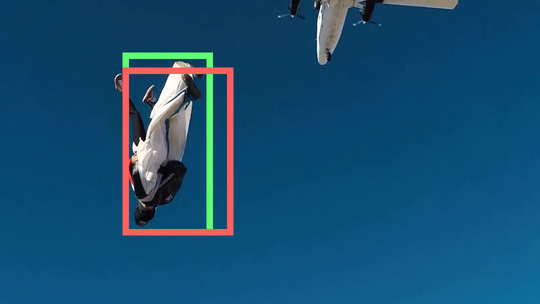}%
    {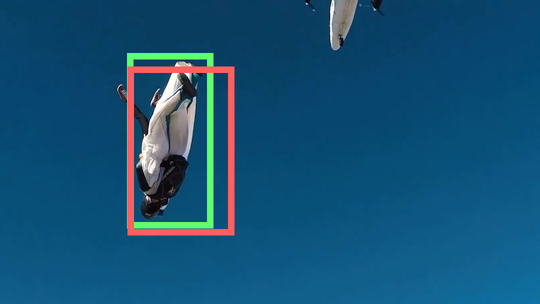}%
    {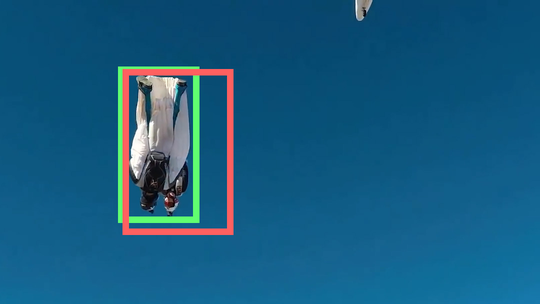}
  \acceptedQualRowEight{Large}%
    {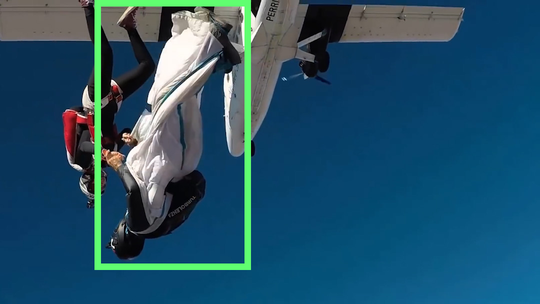}%
    {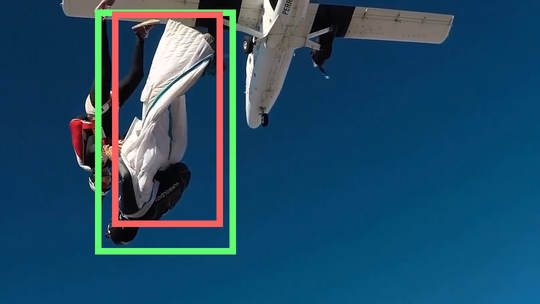}%
    {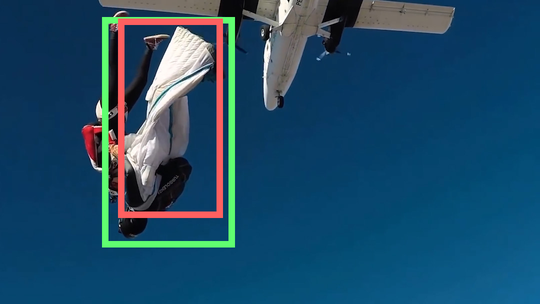}%
    {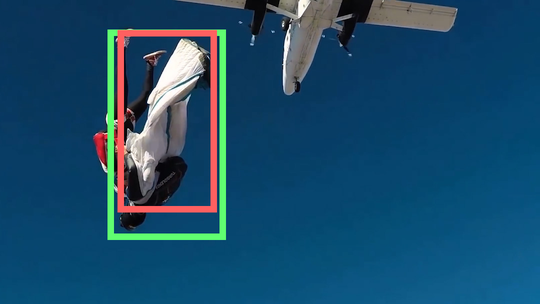}%
    {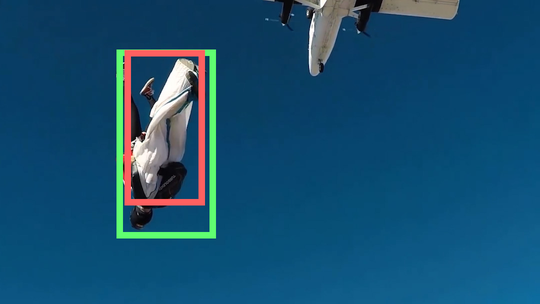}%
    {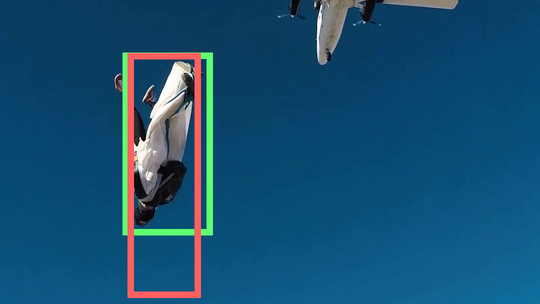}%
    {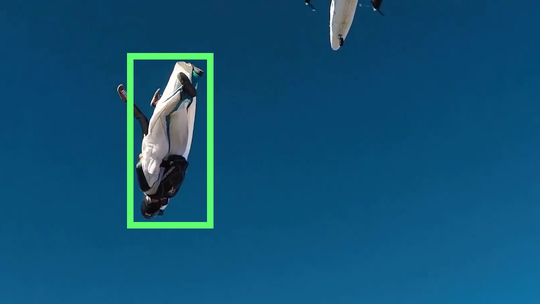}%
    {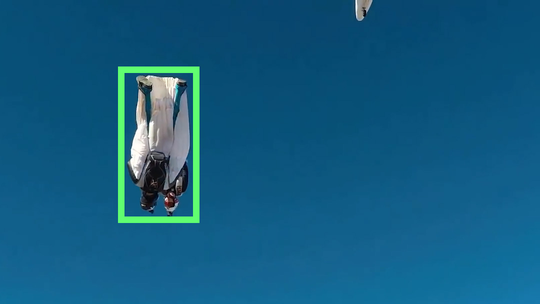}
  \end{minipage}}%
}

\definecolor{judoblue}{RGB}{35,76,112}

\newlength{\judoLabelWidth}
\newlength{\judoImageWidth}

\newcommand{\judoImage}[1]{%
  \raisebox{-0.5\height}{%
    \includegraphics[width=\judoImageWidth]{#1}%
  }%
}

\newcommand{\judoRowSix}[7]{%
  \noindent\hbox to \linewidth{%
    \makebox[\judoLabelWidth][c]{%
      \rotatebox[origin=c]{90}{\scriptsize\bfseries #1}%
    }\hspace{1pt}%
    \judoImage{#2}\hspace{1pt}%
    \judoImage{#3}\hspace{1pt}%
    \judoImage{#4}\hspace{1pt}%
    \judoImage{#5}\hspace{1pt}%
    \judoImage{#6}\hspace{1pt}%
    \judoImage{#7}\hss%
  }\par\vspace{0.5pt}%
}

\newcommand{\JudoCleanSix}[1]{%
  \resizebox{#1\textwidth}{!}{%
    \begin{minipage}{\textwidth}
      \setlength{\judoLabelWidth}{12pt}%
      \setlength{\judoImageWidth}{%
        \dimexpr(\linewidth-\judoLabelWidth-6pt)/6\relax
      }%

      \par\vspace{2pt}

      \judoRowSix{Clean SAM2}%
        {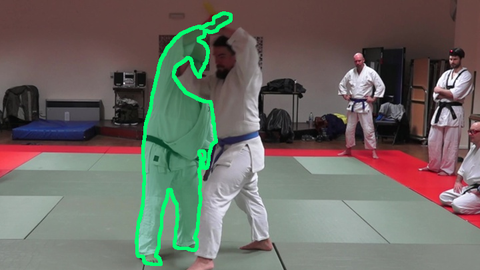}%
        {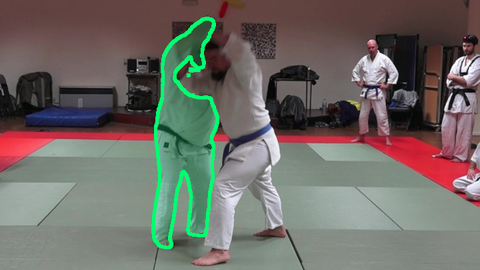}%
        {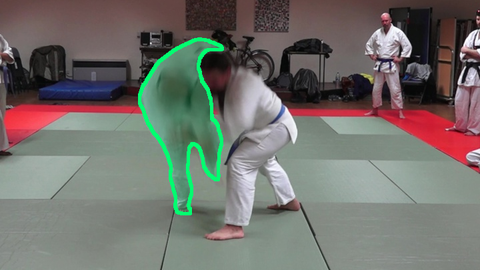}%
        {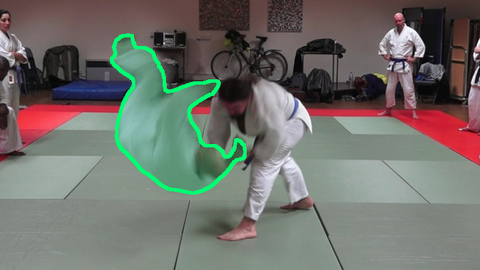}%
        {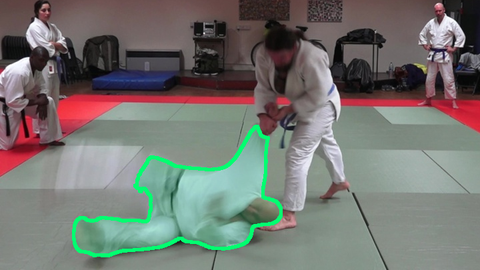}%
        {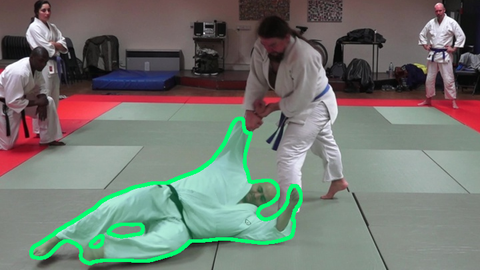}

      \judoRowSix{UAP-SAM2}%
        {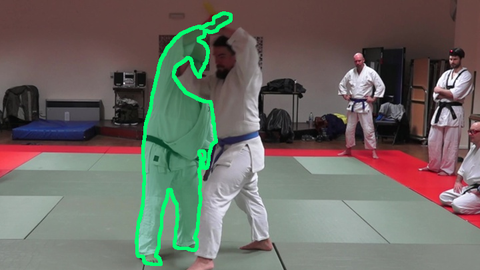}%
        {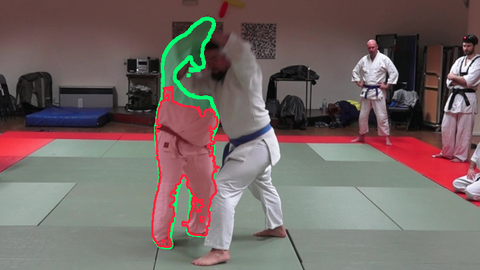}%
        {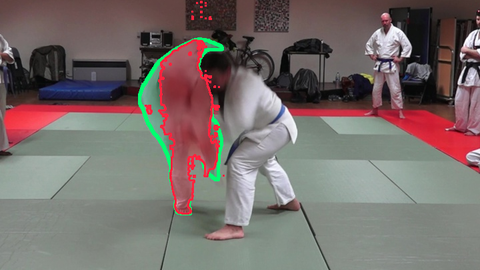}%
        {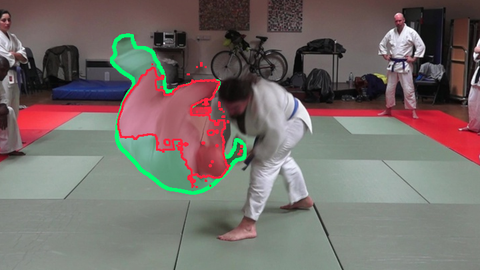}%
        {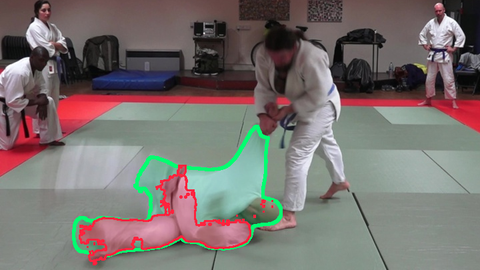}%
        {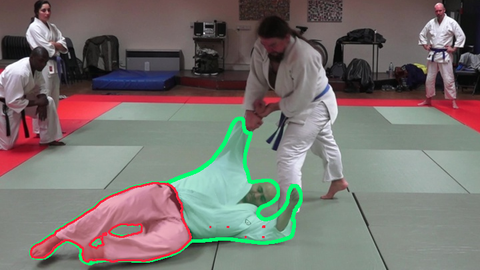}

      \judoRowSix{UAD}%
        {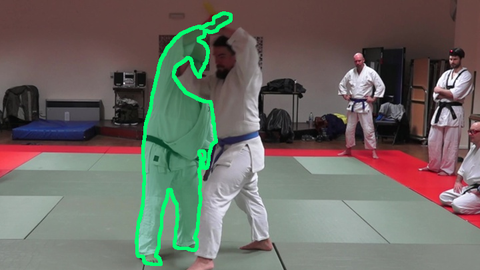}%
        {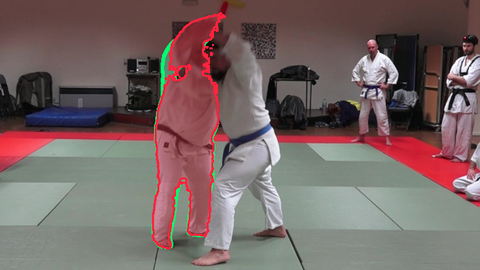}%
        {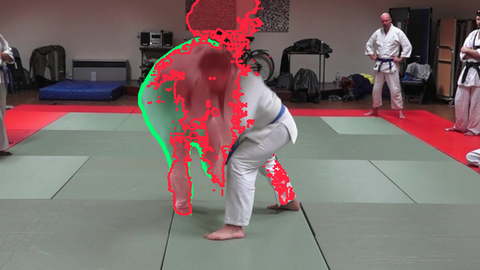}%
        {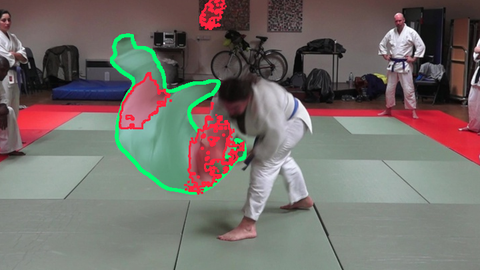}%
        {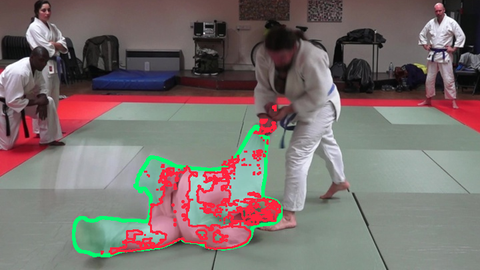}%
        {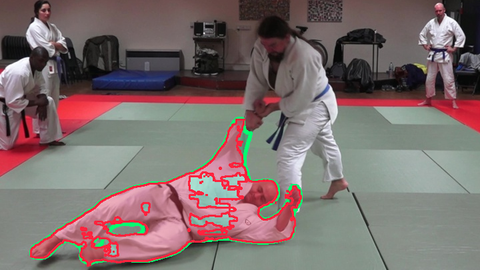}

      \judoRowSix{Ours}%
        {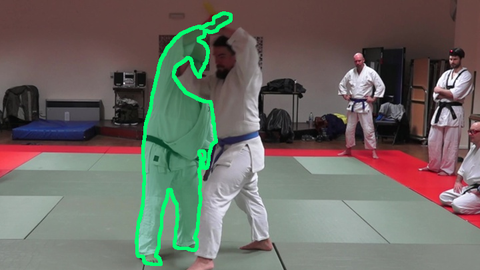}%
        {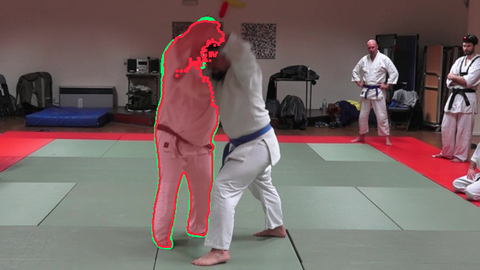}%
        {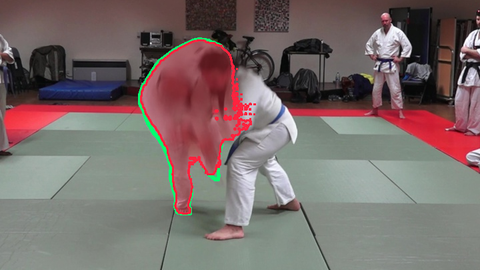}%
        {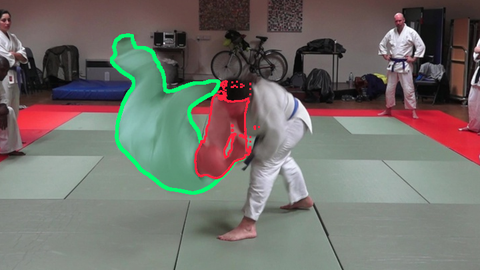}%
        {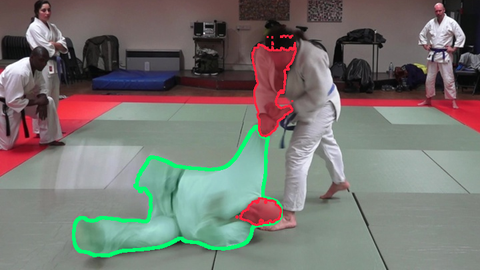}%
        {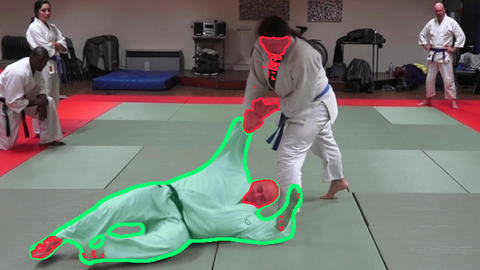}
    \end{minipage}%
  }%
}

\newcommand{\judoRowEight}[9]{%
  \noindent\hbox to \linewidth{%
    \makebox[\judoLabelWidth][c]{%
      \rotatebox[origin=c]{90}{\scriptsize\bfseries #1}%
    }\hspace{1pt}%
    \judoImage{#2}\hspace{1pt}%
    \judoImage{#3}\hspace{1pt}%
    \judoImage{#4}\hspace{1pt}%
    \judoImage{#5}\hspace{1pt}%
    \judoImage{#6}\hspace{1pt}%
    \judoImage{#7}\hspace{1pt}%
    \judoImage{#8}\hspace{1pt}%
    \judoImage{#9}\hss%
  }\par\vspace{0.5pt}%
}

\vspace{-0.4cm}

\begin{itemize}
    \setlength{\itemsep}{0.25em}

    \item \textbf{8. Implementation Details}
    \begin{itemize}
        \setlength{\itemsep}{0em}
        \item 8.1. Attack Optimization
        \item 8.2. CDS Configuration
        \item 8.3. Datasets and Metrics
        \item 8.4. Baseline Adaptations
    \end{itemize}

    \item \textbf{9. Supplementary Evaluation}
    \begin{itemize}
        \setlength{\itemsep}{0em}
        \item 9.1. Per-Dataset VOT: Generic vs. Distractor-heavy
        \item 9.2. VOS Transferability
        \item 9.3. Additional Qualitative Results
        \item 9.4. Additional Ablations
    \end{itemize}

    \item \textbf{10. Defense Study}
    \begin{itemize}
        \setlength{\itemsep}{0em}
        \item 10.1. Robust-SAM
        \item 10.2. Training-Free Attention Modulation
    \end{itemize}
\end{itemize}

\vspace{0.2cm}

\section{Implementation Details}
\label{supp:implementation}

This section gives the settings used to reproduce DistractAttack. We follow the
VP and OP definitions in \cref{sec:method}. The victim model and its tracker
state remain differentiable, while all model parameters are frozen.

\subsection{Attack Optimization}

\noindent\textbf{Shared setup.}
We use the official pretrained SAM~2.1-L checkpoint~\cite{sam2}.  Frames follow the official SAM~2 preprocessing and are
resized to $1024\times1024$. The perturbation is initialized from
$\delta\sim\mathcal{U}(-\epsilon,\epsilon)$, with $\epsilon$ defaulting to 8/255 and 4/255, for VP and OP respectively,  and updated with the projected signed
gradient rule in \cref{eq:update}. After every update, $\delta$ is projected onto the $\ell_\infty$ ball of radius $\epsilon$. When composing an attacked frame, $x+\delta$ is clipped to $[0,1]$ before normalization.

\noindent\textbf{Video perturbation.}
VP optimizes one perturbation for each video. The same perturbation is applied
to every frame in its fixed window. Clean and attacked trajectories are
propagated together during each pass. Their detached banks provide the temporal
context used by the attack losses.

\noindent\textbf{One-shot perturbation.}
OP changes only the prompted frame. Later frames remain clean and provide
evidence for the target and effective-distractor banks.  The clutter objective acts
on the perturbed frame 0 encoder features, while later clean frames contribute
detached evidence to the banks. Neither setting uses ground truth beyond the
initialization prompt during optimization.

\subsection{CDS Configuration}

\noindent\textbf{Banks.}
The target bank stores at most 20 foreground and background prototype pairs.
Later entries require predicted-IoU and objectness scores above 0.5 and 0,
respectively. We use a novelty threshold of $\tau_{\mathrm{nov}}=0.98$. These checks
prevent closely repeated observations from filling the bank.
The
effective-distractor bank has capacity $K=10$. Its merge threshold is
$\theta_{\mathrm{merge}}=0.85$ and its EMA rate is $\eta=0.1$. We set the routing temperature to $\tau_w=0.10$ and the clutter-loss contrastive temperature to $\tau_c=0.07$.

\noindent\textbf{Loss.}
A grid search over $\{0.25,0.5,1.0\}$ for each loss weight selected
$\lambda_M=\lambda_D=\lambda_C=1.0$. The global
and local clutter weights are $\eta_G=1.0$ and $\eta_L=1.0$. 

\subsection{Datasets and Metrics}

\noindent\textbf{Visual object tracking.}
The VOT set contains 104 randomly sampled sequences from LaSOT~\cite{lasot},
UTB180~\cite{utb180}, GOT-10k~\cite{got10k}, and
LaSOT-ext~\cite{lasot}. The LaSOT and UTB180 subset follows the
distractor-distilled DiDi selection~\cite{dam4sam}. GOT-10k and LaSOT-ext add
broader tracking conditions including videos with no distractors. Each source dataset contributes 22-31 randomly sampled sequences. Each
sequence contributes a long 100-frame window, in contrast to 15 frame window used by published UAP-SAM2. None of these evaluation splits
was used to train the original SAM~2 model.

\noindent\textbf{Video object segmentation.}
The VOS set contains 100 videos. We sample 25 videos from each source and select
one target object. DAVIS~2017~\cite{davis2017} provides
accurate boundaries. YouTube-VOS~2018 validation~\cite{youtubevos2018} adds
broad appearance and category variation. The MOSEv2 ~\cite{mose}
contains crowded scenes and severe occlusion. SA-V test~\cite{sam2} covers motion and disappearance. Similarily, each video is 100 frame long. None of these evaluation splits
was used to train the original SAM~2 model.

\noindent\textbf{VOT metrics.}
Each predicted mask is converted to its tight box, with an empty
mask denoting target absence. Let $\mathcal{I}_s$ contain the evaluated frames
of sequence $s$, excluding initialization and unavailable annotations, and let
$N_s=|\mathcal{I}_s|$. We denote the predicted box, ground-truth box, and their
IoU by $\widehat B_{s,t}$, $B_{s,t}$, and $o_{s,t}$, respectively. We follow VOTS2023~\cite{kristan2023first} for all our VOT metrics. We use $o_{s,t}=1$ when both boxes are empty and $o_{s,t}=0$ when only one is
empty. Tracking quality is
\begin{equation}
Q=
\frac{1}{S}\sum_{s=1}^{S}
\frac{1}{N_s}\sum_{t\in\mathcal{I}_s}o_{s,t}.
\end{equation}

Let $\mathcal{V}_s\subseteq\mathcal{I}_s$ contain the visible-target frames.
Within this set, $T_s$ counts predictions with nonzero overlap, $D_s$ counts
non-empty predictions with zero overlap, and $M_s$ counts empty predictions.
The remaining metrics are
\begin{align}
\mathrm{Acc}
&=
\frac{1}{S}\sum_{s=1}^{S}
\frac{1}{T_s}
\sum_{\substack{t\in\mathcal{V}_s\\o_{s,t}>0}}o_{s,t},
\\
\mathrm{Rob}
&=
\frac{1}{S}\sum_{s=1}^{S}
\frac{T_s}{T_s+D_s+M_s},
\\
\mathrm{DRE}
&=
\frac{1}{S}\sum_{s=1}^{S}
\frac{D_s}{T_s+D_s+M_s},
\\
\mathrm{NRE}
&=
\frac{1}{S}\sum_{s=1}^{S}
\frac{M_s}{T_s+D_s+M_s}.
\end{align}
 Tables report DRE and NRE
as percentages. A stronger attack lowers $Q$, accuracy, and robustness while
increasing DRE and NRE.

\noindent\textbf{VOS metrics.}
Following SAM~2~\cite{sam2}, we report the standard
$\mathcal{J}\&\mathcal{F}$ score, which averages mask IoU and contour accuracy
over annotated non-initialization frames and then over videos. Lower is
stronger.

\subsection{Baseline Adaptations}

\noindent\textbf{Common protocol.}
We retain each baseline's defining objective and apply it under the same VP or
OP constraint as DistractAttack. VP aggregates each method's native loss over the
sequence. OP perturbs only the initialization frame. When a method requires a reference
after frame 0, we use detached clean predictions rather than future
annotations. All methods share the same preprocessing, prompts, temporal
windows, random seeds, $\ell_\infty$ bounds.

\noindent\textbf{Image-based attacks.}
We adapt AttackSAM, DarkSAM, CPA, SegPGD, and UAD from static image segmentation
to the recurrent setting. AttackSAM~\cite{attacksam} applies its ClipMSE
objective to the selected SAM~2 mask prediction.
DarkSAM~\cite{darksam} retains its semantic-decoupling and wavelet-texture terms. 
CPA~\cite{long2025robust} retains its negative-feature objective over the top responses shared by point and box prompts. SegPGD~\cite{segpgd} treats each SAM~2 output
as binary segmentation, uses the detached clean mask as its pseudo-label, and
retains its iteration-dependent weighting schedule. UAD~\cite{uad} preserves
its two-stage feature objective with deformation fields. VP aggregates
this objective across the rollout, while OP applies it to the initialization entering memory.

\noindent\textbf{UAP-SAM2.}
UAP-SAM2~\cite{uapsam2} combines target scanning, semantic confusion, encoder
feature shift, and temporal feature misalignment. We retain these terms while
changing setting to VP and OP. 

\noindent\textbf{One-shot VOS attacks.}
ARA-Attack~\cite{li2023adversarial} and
VOSPGD~\cite{jiang2023exploring} remain in their native one-shot setting.
ARA-Attack recomputes the initialization gradient map and applies its
hard-region learner.
VOSPGD writes the perturbed prompt frame to memory and replays its clean
counterpart as a memory-conditioned query. For VOT, the clean mask obtained
from the initialization box provides the binary reference, while VOS uses the
given initialization mask.

\section{Supplementary Evaluation}
\label{supp:evaluation}

\subsection{Per-Dataset VOT: Generic vs. Distractor-heavy}

\begin{table}[t]
    \centering
    \caption{Per-dataset VOT results extending Table~\ref{tab:attack_results}}
    \label{tab:vot_dataset_results}

    \resizebox{\columnwidth}{!}{%
    \begin{tabular}{l c ccc ccc}
        \toprule
        & &
        \multicolumn{3}{c}{VP} &
        \multicolumn{3}{c}{OP} \\
        \cmidrule(lr){3-5}
        \cmidrule(lr){6-8}

        Dataset
        & Seq.
        & $Q\downarrow$
        & DRE (\%)$\uparrow$
        & NRE (\%)$\uparrow$
        & $Q\downarrow$
        & DRE (\%)$\uparrow$
        & NRE (\%)$\uparrow$ \\
        \midrule

        \multicolumn{8}{l}{\textit{Distractor-heavy}} \\

        LaSOT
        & 22
        & 0.443
        & 11.02
        & 14.44
        & 0.010
        & 8.47
        & 71.95 \\

        UTB180
        & 22
        & 0.385
        & 18.37
        & 10.01
        & 0.156
        & 12.18
        & 47.84 \\

        \multicolumn{8}{l}{\textit{Generic}} \\

        GOT-10k
        & 31
        & 0.470
        & 4.52
        & 14.96
        & 0.049
        & 9.76
        & 70.93 \\

        LaSOT-ext
        & 29
        & 0.377
        & 11.70
        & 22.01
        & 0.095
        & 7.18
        & 69.04 \\

        \midrule
        \textbf{Total}
        & \textbf{104}
        & \textbf{0.420}
        & \textbf{10.83}
        & \textbf{15.77}
        & \textbf{0.076}
        & \textbf{9.28}
        & \textbf{65.73} \\

        \bottomrule
    \end{tabular}%
    }
\end{table}

To complement the aggregate VOT evaluation in Tab.~\ref{tab:attack_results}, we report results separately on
LaSOT, UTB180, GOT-10k, and LaSOT-ext (See Tab.~\ref{tab:vot_dataset_results}). All datasets use the same fixed windows, initialization protocol, and attack settings as the main
evaluation. Results are macro-averaged within each dataset with equal weight per
sequence, while the combined row covers all sequences. This breakdown
examines consistency across datasets with some variation shown.  GOT-10k and LaSOT-ext cover broader
generic tracking conditions and are not restricted to scenes with explicit
competing objects, yet exhibit comparable degradation. Thus, the attack does
not require a literal distractor in every frame: CDS exploits model-supported
off-target regions when available, while foreground suppression remains
effective otherwise.

\subsection{VOS Transferability}
\label{supp:evaluation:transfer}
\begin{figure}[t]
    \centering
    \includegraphics[width=0.75\columnwidth]{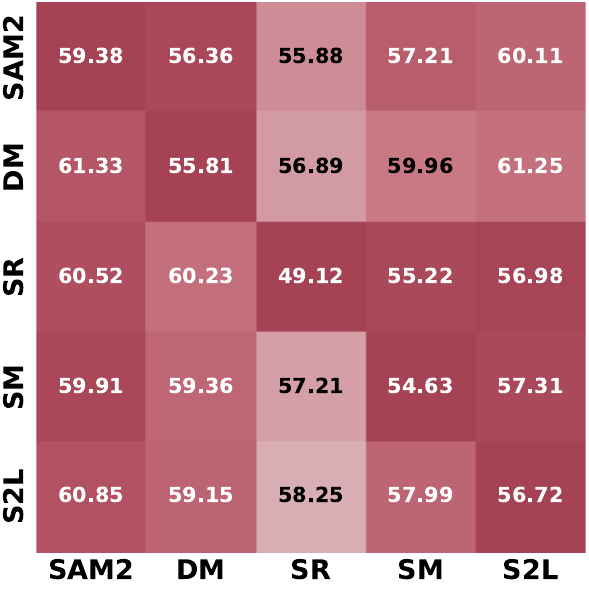}
    \caption{\textbf{Cross-variant VOS transferability under VP.}
    Each row denotes the source tracker used to optimize the perturbation, while
    each column denotes the evaluated target tracker. Lower
    $\mathcal{J}\&\mathcal{F}$ indicates a stronger attack.}
    \label{fig:vos_cross_variant_vp}
\end{figure}

We evaluate VP transfer across SAM2.1, DAM4SAM, SAMURAI, SAMITE, and
SAM2Long on DAVIS 2017.  All variants use the SAM2.1-L checkpoint. The consistently low off-diagonal scores in
\cref{fig:vos_cross_variant_vp} demonstrate strong transfer across tracker
variants. DistractAttack remains effective despite their different tracking
and memory designs, revealing a shared vulnerability across SAM~2-based
trackers.

\subsection{Additional Qualitative Results}
\label{supp:qualitative}

\begin{figure*}[!t]
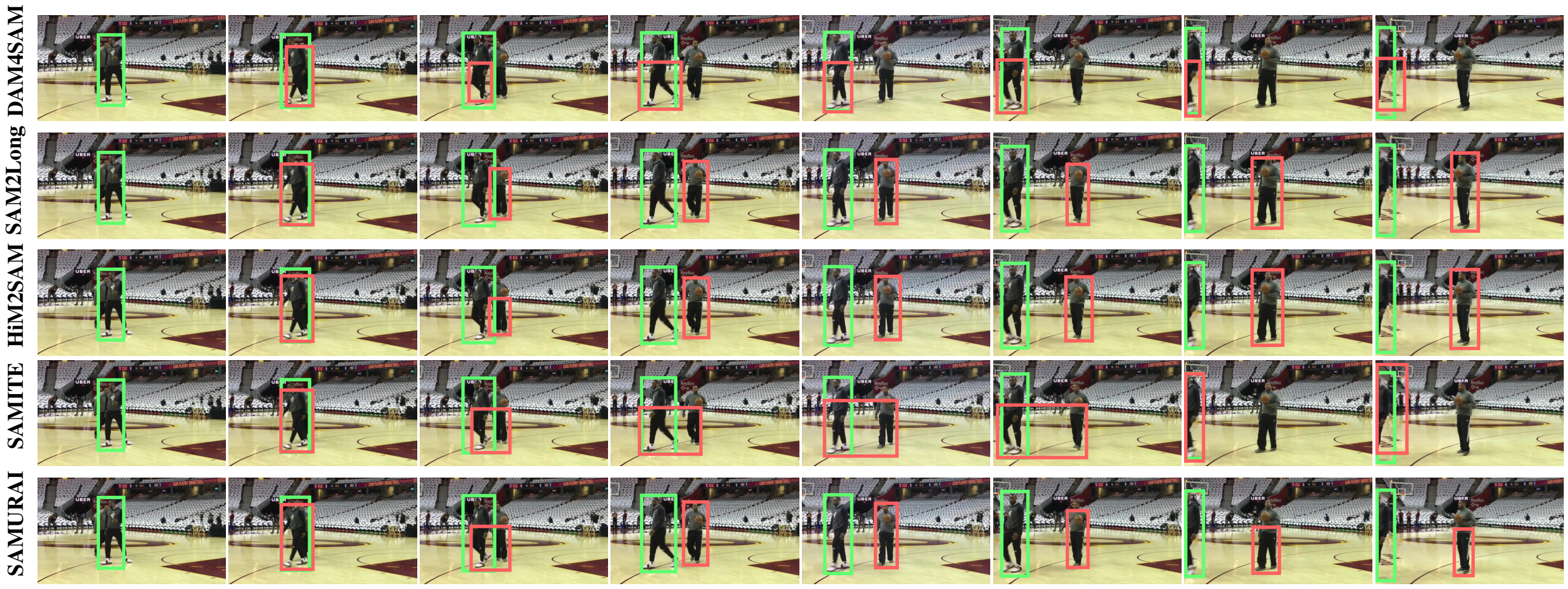

  \centering
  \AcceptedWarmupGrid
  \caption{Qualitative comparison of perturbed tracker variants. Columns represent
  propagation from left to right, where the first column is the prompted frame.
  Green and red denote the clean reference and perturbed predictions, respectively.}
  \label{fig:supp-vot-tracker-variants-eight}
\end{figure*}

\noindent\textbf{Tracker variants.}
As shown in Fig.~\ref{fig:supp-vot-tracker-variants-eight}, we compare identical frames and prompts across DAM4SAM, SAMURAI, HiM2SAM,
SAMITE, and SAM2Long on VOT VP. The examples complement the aggregate results in the main
paper. 

\begin{figure*}[!b]
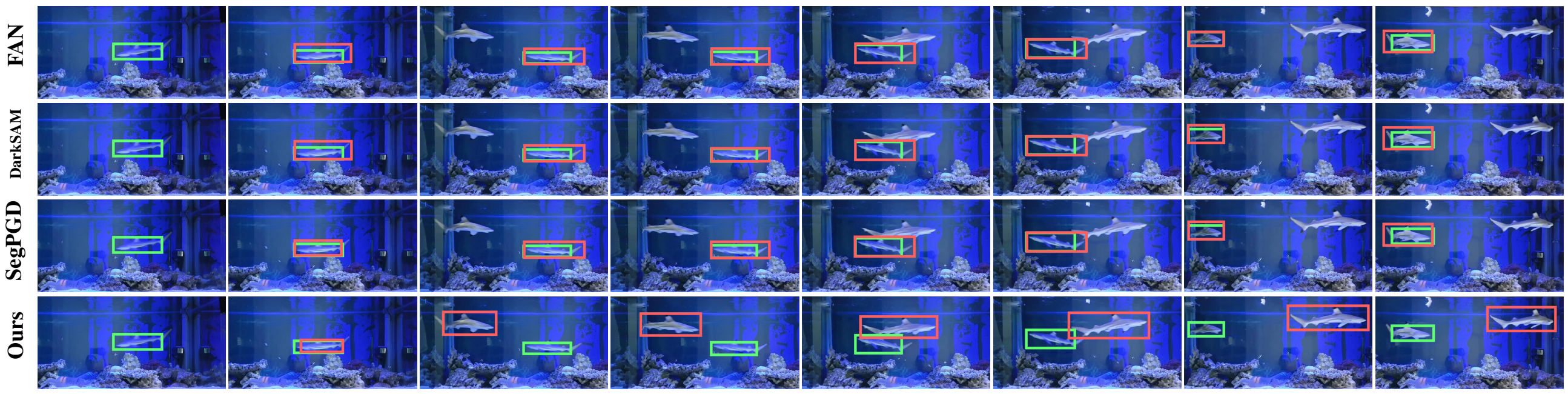

  \centering
  \AcceptedSharkGrid
  \caption{Qualitative comparison of VOT OP methods. Columns represent
  propagation from left to right, where the first column is the prompted frame.
  Green and red denote the clean reference and perturbed predictions.}
  \label{fig:supp-lasot-shark2-eight}
\end{figure*}

\begin{figure*}[t]
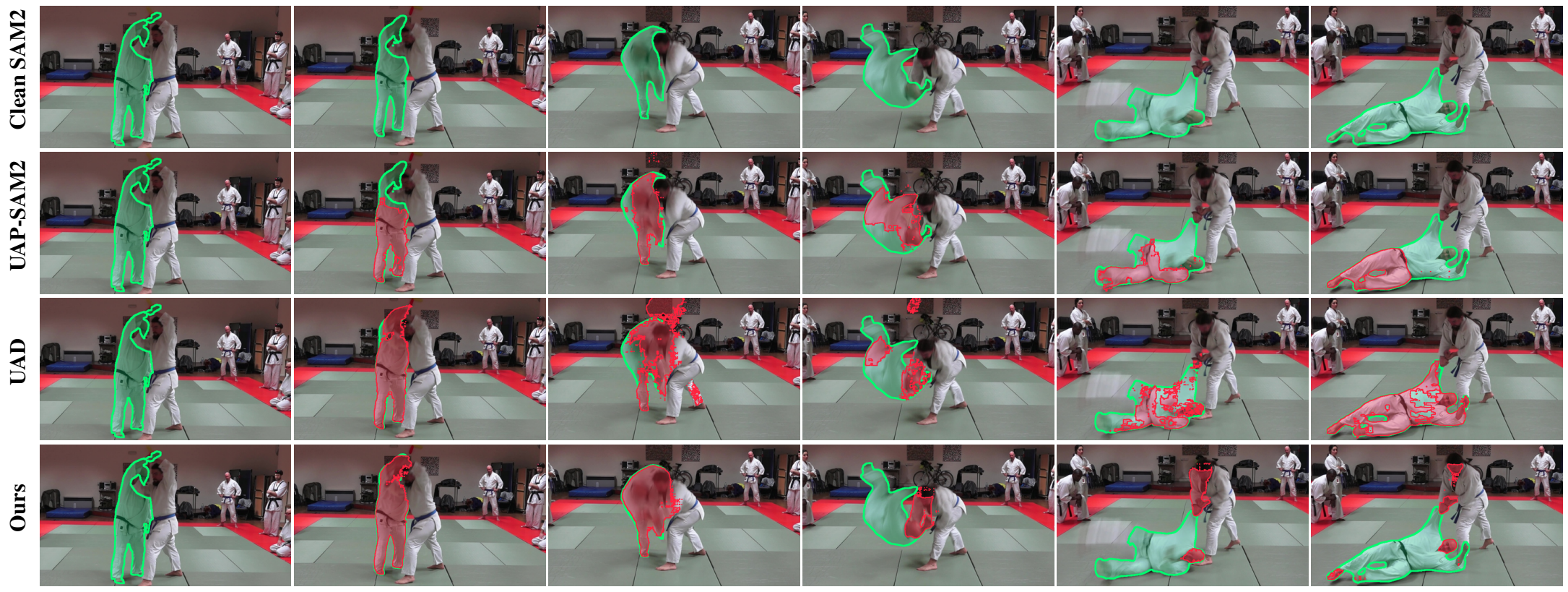

  \centering
  \JudoCleanSix{0.82}
  \caption{%
    Qualitative comparison of VOS VP methods.
    Green denotes the Clean SAM2 reference prediction, while red denotes
    the perturbed prediction. 
  }
  \label{fig:supp-judo-clean-six}
\end{figure*}

\noindent\textbf{Additional method comparisons}
As shown in Fig.~\ref{fig:supp-lasot-shark2-eight} and Fig.~\ref{fig:supp-judo-clean-six}, we extend the qualitative comparison with FAN, SegPGD, DarkSAM, UAD and UAP-SAM2 on VOT OP and VOS VP. 

\begin{figure*}[t]
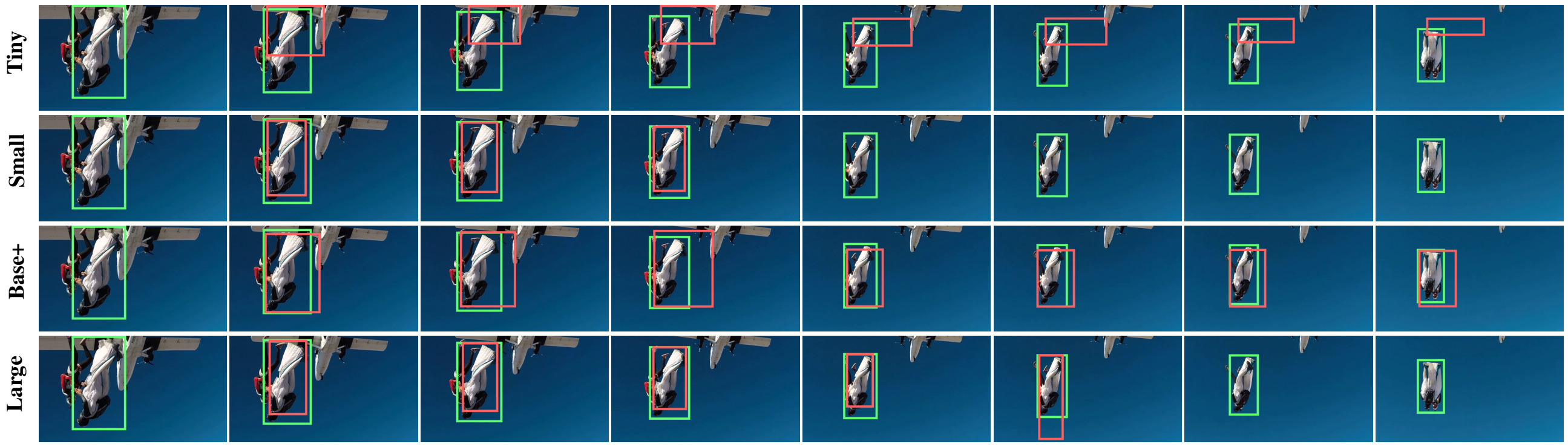

  \centering
  \AcceptedWingsuitGrid
  \caption{Qualitative comparison of perturbed SAM2.1 model-size
  variants on VOT VP.  Columns represent propagation from left to right, where the first column is the prompted frame. Green and
  red denote the clean reference and perturbed predictions.}
  \label{fig:supp-sam21-size-variants-eight}
\end{figure*}

\noindent\textbf{Model size variants.}
As shown in Fig.~\ref{fig:supp-sam21-size-variants-eight}, we next compare the Tiny, Small, Base+, and Large sizes of SAM~2.1 on the same sequences. 

Additional and more clear qualitative comparisons and temporal visualizations are provided
in the accompanying supplementary video.

\subsection{Additional Ablations}
\label{supp:evaluation:ablations}

\textbf{Step Size.}
As in Tab.~\ref{tab:step-size}, we fix the perturbation bound and optimization passes and vary only the
signed-gradient step size. Under VP, $2/255$ gives the strongest overall
degradation ($Q=0.420$, DRE$=10.83\%$, VOS Avg.$=50.97$), while under OP,
$1/255$ performs best overall ($Q=0.076$, VOS Avg.$=19.56$). These settings
are therefore used as the default step sizes.

\begin{table}[t]
    \centering

    \begin{minipage}[t]{0.49\columnwidth}
        \vspace{0pt}
        \centering
        \captionof{table}{Effect of the signed-gradient step size.
        Arrows indicate a stronger attack.}
        \label{tab:step-size}

        \scriptsize
        \setlength{\tabcolsep}{1.5pt}
        \renewcommand{\arraystretch}{0.95}

        \resizebox{\linewidth}{!}{%
        \begin{tabular}{@{}llccc@{}}
            \toprule
            Setting & Step
            & $Q\downarrow$
            & DRE\%$\uparrow$
            & VOS Avg.$\downarrow$ \\
            \midrule
            \multirow{3}{*}{VP}
                & $1/255$ & 0.453 & 6.87 & 66.51 \\
                & $2/255$ & \textbf{0.420}
                          & \textbf{10.83}
                          & \textbf{50.97} \\
                & $4/255$ & 0.445 & 5.67 & 53.09 \\
            \midrule
            \multirow{2}{*}{OP}
                & $1/255$ & \textbf{0.076}
                          & 9.28
                          & \textbf{19.56} \\
                & $2/255$ & 0.122
                          & \textbf{10.23}
                          & 21.67 \\
            \bottomrule
        \end{tabular}%
        }
    \end{minipage}
    \hfill
    \begin{minipage}[t]{0.49\columnwidth}
        \vspace{0pt}
        \centering
        \captionof{table}{Ablation of uniform and distractor-weighted
        $\mathcal{L}_{\mathrm{disp}}$. Arrows indicate a stronger attack.}
        \label{tab:disp-weighting}

        \scriptsize
        \setlength{\tabcolsep}{1.5pt}
        \renewcommand{\arraystretch}{0.95}

        \resizebox{\linewidth}{!}{%
        \begin{tabular}{@{}llccc@{}}
            \toprule
            Setting & Displacement
            & $Q\downarrow$
            & DRE\%$\uparrow$
            & VOS Avg.$\downarrow$ \\
            \midrule
            \multirow{2}{*}{VP}
                & Unweighted
                & 0.450 & 2.51 & 54.11 \\
                & Bank-weighted
                & \textbf{0.420}
                & \textbf{10.83}
                & \textbf{50.97} \\
            \midrule
            \multirow{2}{*}{OP}
                & Unweighted
                & 0.145 & 1.43 & 24.13 \\
                & Bank-weighted
                & \textbf{0.076}
                & \textbf{9.28}
                & \textbf{19.56} \\
            \bottomrule
        \end{tabular}%
        }
    \end{minipage}

\end{table}
\noindent\textbf{Unweighted versus weighted displacement.}
As in Tab.~\ref{tab:disp-weighting}, we isolate the weighting in $\mathcal{L}_{\mathrm{disp}}$ while retaining the
same proposal set, effective-distractor bank, $\mathcal{L}_{\mathrm{mask}}$,
and $\mathcal{L}_{\mathrm{clut}}$. The unweighted variant assigns
$1/|\mathcal{P}_t|$ to every active proposal as in
\cref{eq:uniform-disp-loss}. The weighted variant uses the bank-supported
$w_t^j$ in \cref{eq:proposal-routing,eq:disp-loss}. Thus, the comparison changes
only how foreground displacement is routed.

\begin{table}[t]
    \centering
    \caption{Defense evaluation on SAM~2.1-L under VP and OP
    DistractAttack. The attack is recomputed against each defense.}
    \label{tab:defense_results}
    \scriptsize
    \setlength{\tabcolsep}{2.8pt}
    \renewcommand{\arraystretch}{1.02}
    \resizebox{\columnwidth}{!}{%
    \begin{tabular}{@{}lccc|ccc@{}}
        \toprule
        & \multicolumn{3}{c|}{VP}
        & \multicolumn{3}{c}{OP} \\
        \cmidrule(lr){2-4}\cmidrule(lr){5-7}
        Method
        & $Q$ & DRE\% & NRE\%
        & $Q$ & DRE\% & NRE\% \\
        \midrule

        \textcolor{gray}{Clean reference}
        & \textcolor{gray}{0.836}
        & \textcolor{gray}{0.68}
        & \textcolor{gray}{0.42}
        & \textcolor{gray}{0.836}
        & \textcolor{gray}{0.68}
        & \textcolor{gray}{0.42} \\

        DistractAttack
        & \textcolor{gray}{0.420}
        & \textcolor{gray}{10.83}
        & \textcolor{gray}{15.77}
        & \textcolor{gray}{0.076}
        & \textcolor{gray}{9.28}
        & \textcolor{gray}{65.73} \\

        Memory modulation~\cite{sam2long}
        & 0.476 & 9.52 & 10.33
        & 0.101 & 10.55 & 60.21 \\

        Robust-SAM~\cite{long2025robust}
        & 0.515 & 9.01 & 10.12
        & 0.129 & 3.45 & 54.45 \\

        Both~\cite{long2025robust,sam2long}
        & 0.532 & 8.45 & 10.42
        & 0.133 & 4.51 & 55.45 \\

        \bottomrule
    \end{tabular}%
    }
\end{table}

\section{Defense Study}
\label{sec:supp-defense}

The main paper tests a feature-level defense and a memory-level defense. We
describe both adaptations here. The attack remains white box and is recomputed
against each defended model.

\subsection{Robust-SAM}

Robust-SAM~\cite{long2025robust} adapts the singular values of selected
convolutional weights. We apply this parameterization to the convolutional
weights in the final image-encoder block of SAM~2.1-L. Each selected weight is
decomposed as $W=UPV^\top$, where $U$ and $V$ remain frozen and only the
diagonal matrix $P$ is optimized. All other SAM~2 parameters remain fixed.

We train a separate defended model for each perturbation setting using CPA
adversarial examples generated from 5,000 videos in the SA-V training
split~\cite{sam2}. The training bound matches the corresponding evaluation
setting, with $\epsilon=8/255$ for VP and $\epsilon=4/255$ for OP. Each model is
adapted for 15 epochs using a batch size of 64 and Adam with a learning rate of
$10^{-3}$. Thus, the VP and OP results use independently adapted checkpoints
rather than a single jointly trained defense.

\subsection{Training-Free Attention Modulation}

SAM2Long~\cite{sam2long} improves memory cross-attention by weighting historical
memories according to their object-confidence scores. We isolate this mechanism
from the original memory-tree formulation and apply it directly to SAM~2.1-L,
without additional training or learnable parameters. For the $N+1$ memory
entries, we first construct linearly spaced weights between $w_{\mathrm{low}}$
and $w_{\mathrm{high}}$:
\begin{equation}
w_i^{\mathrm{std}}
=
w_{\mathrm{low}}
+
\frac{i-1}{N}
\left(w_{\mathrm{high}}-w_{\mathrm{low}}\right),
\qquad i=1,\ldots,N+1.
\end{equation}

We then sort the associated confidence scores in ascending order,
$o_{I_1}\leq\cdots\leq o_{I_{N+1}}$, and assign larger weights to memories with
higher scores. The corresponding memory keys are modulated before
cross-attention as
\begin{equation}
w_{I_i}=w_i^{\mathrm{std}},
\qquad
\widetilde{\mathbf M}_{I_i}
=
w_{I_i}\mathbf M_{I_i},
\qquad i=1,\ldots,N+1.
\end{equation}
Unlike SAM2Long, we adopt a wider weighting interval of $[0.75, 1.25]$ to increase the modulation strength, thereby amplifying the distinction between low- and high-confidence memories. Thus, reliable high-confidence memories contribute more strongly to the memory
attention, while low-confidence entries are naturally suppressed.

As shown in Tab.~\ref{tab:defense_results}, both defenses recover part of the lost tracking quality. Robust-SAM gives the
larger individual gain. Their combination reaches the strongest defended
result, yet it remains well below clean tracking. Existing feature and memory
controls therefore reduce the failure without removing it. A stronger defense
may need to estimate identity reliability before each memory is read.

\end{document}